%% file: main.tex
\documentclass[10pt,twocolumn,letterpaper]{article}
\usepackage{comment}
\usepackage[pagenumbers]{cvpr} % To force page numbers, e.g., for an arXiv version
\usepackage{algorithm}
\usepackage{algpseudocode}
\usepackage[table]{xcolor}
\usepackage{multirow}


\definecolor{cvprblue}{rgb}{0.21,0.49,0.74}
\usepackage[pagebackref,breaklinks,colorlinks,allcolors=cvprblue]{hyperref}

\def\paperID{30591} % *** Enter the Paper ID here
\def\confName{CVPR}
\def\confYear{2025}

\title{\textbf{ConsensusDrop: Fusing Visual and Cross-Modal \\ Saliency for Efficient Vision Language Models}}

\author{
Dhruv Parikh\textsuperscript{1} \quad
Haoyang Fan\textsuperscript{1} \quad
Rajgopal Kannan\textsuperscript{2} \quad
Viktor Prasanna\textsuperscript{1} \\
\\
\textsuperscript{1}University of Southern California \\
\{dhruvash, haoyangf, prasanna\}@usc.edu \\
\textsuperscript{2}DEVCOM Army Research Office \\
rajgopal.kannan.civ@army.mil \\
}

\begin{document}
\maketitle

\input{sec/0_abstract}

\input{sec/1_intro}

\input{sec/2_related}
\input{sec/3_motivation}
\input{sec/4_method}
\input{sec/5_experiments}

\input{sec/6_conclusion}

{
    \small
    \bibliographystyle{ieeenat_fullname}
    \bibliography{main}
}

\input{sec/X_suppl}

\end{document}

%% file: sec/0_abstract.tex
\begin{abstract}
Vision--Language Models (VLMs) are expensive because the LLM processes hundreds of largely redundant visual tokens. Existing token reduction methods typically exploit \textit{either} vision-encoder saliency (broad but query-agnostic) \textit{or} LLM cross-attention (query-aware but sparse and costly). We show that neither signal alone is sufficient: fusing them consistently improves performance compared to unimodal visual token selection (ranking). However, making such fusion practical is non-trivial: cross-modal saliency is usually only available \emph{inside} the LLM (too late for efficient pre-LLM pruning), and the two signals are inherently asymmetric, so naive fusion underutilizes their complementary strengths. We propose \textbf{ConsensusDrop}, a training-free framework that derives a \emph{consensus} ranking by reconciling vision encoder saliency with query-aware cross-attention, retaining the most informative tokens while compressing the remainder via encoder-guided token merging. Across LLaVA-1.5/NeXT, Video-LLaVA, and other open-source VLMs, ConsensusDrop consistently outperforms prior pruning methods under identical token budgets and delivers a stronger accuracy--efficiency Pareto frontier---preserving near-baseline accuracy even at aggressive token reductions while reducing TTFT and KV cache footprint. Our code will be open-sourced.
\end{abstract}

%% file: sec/1_intro.tex
\section{Introduction}
\label{sec:intro}

The recent advancement in Large Language Models (LLMs), encompassing both closed- \cite{gpt5, claude-4.5} and open-source paradigms \cite{kimi-k2, gpt-oss, qwen-3-next, glm-4.5}, has fueled a new generation of Vision-Language Models (VLMs). These models achieve complex multimodal reasoning—either natively \cite{llama-4, gpt5} or through architectural extensions \cite{llava, qwen-3-omni}—and have extended capabilities to specialized domains such as document understanding, chart interpretation, and long-video comprehension \cite{deepseek_ocr_2025, qwen25_vl_2025, llava_onevision_2024, llava_onevision15_2025, chartmoe_iclr2025, mmmu_2023, internvl25_2024}. Standard VLMs typically employ a pre-trained vision encoder \cite{clip, siglip} to map images into patch tokens, followed by a multimodal projector \cite{llava, llavanext, improvedllava} that aligns these embeddings with the LLM. However, this paradigm incurs significant computational overhead. A single image is typically encoded into hundreds or thousands of tokens (e.g., 576 in LLaVA-1.5 \cite{llava}, 2880 in LLaVA-Next \cite{llavanext}), vastly exceeding the textual input containing fewer than a hundred tokens. This redundancy leads to prohibitive prefill latency and excessive KV-cache consumption during decoding, hindering scalable deployment in latency-sensitive applications such as robotics and autonomous driving \cite{openvla, edge-llm-survey, unified-vlm}.

\begin{figure*}
    \centering
    \includegraphics[width=0.8\textwidth]{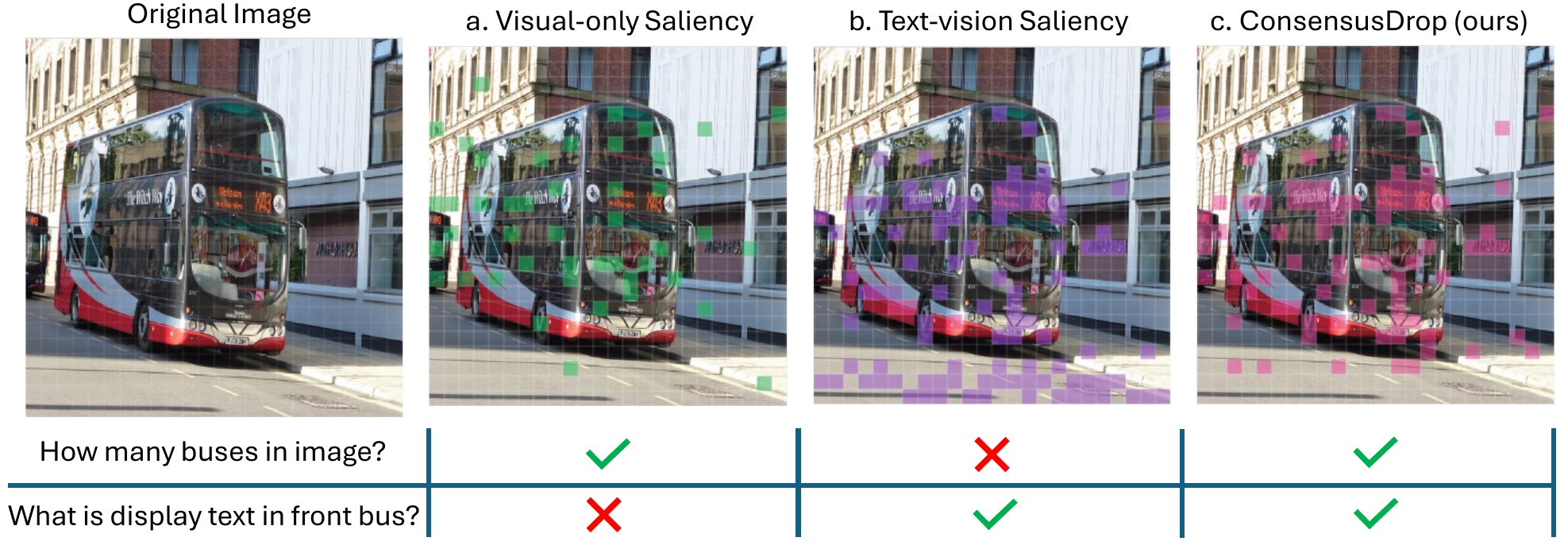}
    \caption{
    \textbf{Motivation: Complementary failure modes of visual token sparsification.}
    Given a single image and two questions requiring distinct visual evidence
    (global structure vs.\ localized fine-grained text), different token selection criteria
    exhibit systematic limitations.
    \emph{Vision-only saliency} preserves broad spatial coverage and visually salient regions, but may discard small, text-critical regions.
    \emph{Text--vision saliency} captures query-aligned local details,
    but is often sparse and biased towards late-index patches.
    \textbf{ConsensusDrop (ours)} retains complementary evidence by fusing visual and cross-modal saliency signals, preserving information required across diverse questions.
    }
    \label{fig:motivation}
\end{figure*}

Motivated by the substantial redundancy in visual tokens, a large body of work improves VLM efficiency by ranking and discarding less informative image patches. These works generally follow two paradigms: \textbf{Cross-modal-only pruning}~\cite{aim-prune-cross, fastv-prune-cross, zipvl-prune-cross, meteor-prune-cross, feather-prune-cross, twigvlm-prune-cross, atpllava-prune-cross, fitprune-prune-cross, sparsevlm-prune-cross, skipvision-prune-cross, pyramiddrop-prune-cross} ranks patches using text--vision attention  between the input prompt and image tokens inside the LLM, while \textbf{Vision-only pruning}~\cite{pact-prune-viz, clstoken-prune-viz, vflowopt-prune-viz, visionzip-prune-viz, stitchintime-prune-viz, dynamictokeneff-prune-viz, vispruner-prune-viz, llavaprumerge-prune-train-viz} filters tokens based on vision encoder-side saliency. While both families have demonstrated meaningful speedups and competitive accuracy under varying retention budgets~\cite{pact-prune-viz, clstoken-prune-viz, vflowopt-prune-viz, visionzip-prune-viz, stitchintime-prune-viz, dynamictokeneff-prune-viz, vispruner-prune-viz, llavaprumerge-prune-train-viz}, they suffer from non-trivial limitations that hinder their effectiveness. Cross-modal-only pruning suffers from: (i) strong \emph{positional bias} from rotary embeddings that over-weights late-index patches, (ii) \emph{sparse and unstable} cross-scores that can miss visually salient regions, and (iii) practical inefficiency—extracting layer attentions disables FlashAttention \cite{flashattention} and induces \emph{non-uniform KV caches} with fragmented memory. 
% Vision–only pruning, while efficient, is \emph{query-agnostic}: importance is decided without the user prompt, which can drop regions crucial to the question. This effect is exacerbated in document or OCR-style benchmarks, where small, text-conditioned visual regions (e.g., embedded text snippets or chart elements) may be pruned despite their high relevance to the query. Moreover, vision encoders exhibit attention-sink behavior, where background tokens accumulate global context and receive inflated scores, while fine-grained, query-relevant regions are often overlooked. 
Conversely, Vision-only pruning remains \emph{query-agnostic}, dropping regions crucial to the question, especially the small, text-conditioned regions (e.g., embedded text snippets or chart elements) in document or OCR-style tasks. Moreover, vision encoders exhibit attention-sink behavior, which inflate background scores at the expense of fine-grained, query-relevant details.
Additionally, several methods in both families require additional VLM training \cite{llamavid-prune-train, llavolta-prune-train, llavamini-prune-train, matryoshka-prune-train, matryoshka2-prune-train, tokenpacker-prune-train, llavaprumerge-prune-train-viz}, which imposes additional training overhead.

We hypothesize that visual and cross-modal saliency are complementary and that \emph{both} are essential for robust visual token selection (Fig.~\ref{fig:motivation}). Our controlled experiments confirm this hypothesis: enforcing a \textbf{multimodal consensus} yields more reliable importance estimates and consistently superior performance compared to unimodal pruning. However, exploiting such consensus introduces two key challenges. First, cross-modal saliency is typically computed \emph{inside} the language model, forcing pruning to occur late in the inference pipeline, which incurs additional overhead and can lead to inefficient execution and KV cache fragmentation~\cite{visionzip-prune-viz}. Second, visual and cross-modal signals are inherently \emph{asymmetric}—one provides broad visual coverage while the other is sparse but query-specific—making naive fusion ineffective. This raises a fundamental question: \emph{how can we reconcile these complementary yet asymmetric signals to derive a principled consensus for token selection?}

Building upon these findings and addressing the limitations, we propose \textbf{ConsensusDrop}, a training-free framework for efficient visual token reduction. Specifically, we first introduce the \textbf{Static Cross-Attention Probe (SCAP)}, a lightweight and training-free module that extracts text--vision attention scores \emph{before} they enter the language model, thereby avoiding the need to access internal LLM attention maps. Using SCAP and the vision encoder’s attention scores, we then design a novel \textbf{Fuser Module} that performs ``multimodal" score fusion, identifying the top-$K$ most informative visual tokens to retain. Subsequently, the remaining tokens are compressed into a dense subset of visual tokens via a novel \textbf{Encoder-Guided Token Merge (EGTM)} module. Empirically, \textbf{ConsensusDrop} achieves state-of-the-art accuracy-efficiency tradeoffs across diverse multimodal benchmarks. 

By pruning before LLM, it retains FlashAttention \cite{flashattention} and lowers KV-cache cost. The training-free design naturally supports generalization to high-resolution and video-based models (e.g., LLaVA-Next and Video-LLaVA). These results highlight that multimodal consensus is an effective method for visual token selection. Our key contributions are as follows:

\begin{itemize}
    \item We show that \textbf{multimodal consensus}—combining visual saliency with cross-modal cues—consistently outperforms unimodal visual token pruning.
    \item We introduce a \textbf{Static Cross-Attention Probe (SCAP)}, a lightweight pre-LLM module for efficiently extracting query-aware cross-modal saliency.
    \item To reconcile the asymmetry between visual and cross-modal cues, we propose a \textbf{Fuser Module} that yields a unified and robust visual token-importance ranking.
    \item We propose \textbf{Encoder-Guided Token Merging (EGTM)} to compactly merge discarded tokens in the projector space under vision encoder guidance.
    \item \textbf{ConsensusDrop} is training-free and plug-and-play, and achieves state-of-the-art accuracy--efficiency trade-offs across multiple open-source VLMs.
\end{itemize}

%% file: sec/2_related.tex
\section{Related Works}

\noindent \textbf{Vision-only Pruning.} This paradigm scores tokens using vision encoder-side saliency signals to remove redundancy before the LLM. 
Representative methods include \cite{clstoken-prune-viz, llavaprumerge-prune-train-viz}, ~\cite{pact-prune-viz, stitchintime-prune-viz, vflowopt-prune-viz, dynamictokeneff-prune-viz}. VisionZip~\cite{visionzip-prune-viz} and VisPruner~\cite{vispruner-prune-viz} achieve state-of-the-art performance by scoring visual patches via encoder attention and compressing discarded tokens through contextual merging or similarity-driven de-duplication.
While effective and training-free, these methods remain inherently query-agnostic because importance is determined solely in the visual modality.

% % \noindent \textbf{Cross-modal Token Pruning.}
% \noindent \textbf{Cross-modal-only Pruning.}
% Another line of work prunes visual tokens using \emph{text-vision cross-attention} from within the LLM. FastV~\cite{fastv-prune-cross} ranks image tokens using early-layer cross-attention to textual queries, while SparseVLM~\cite{sparsevlm-prune-cross} extends this idea with sparsified attention patterns. Subsequent works explore deeper or multi-stage cross-attention signals, including pyramid-based reduction~\cite{pyramiddrop-prune-cross}, training-free pruning via attention aggregation~\cite{fitprune-prune-cross, zipvl-prune-cross}, and multi-encoder collaborative pruning~\cite{meteor-prune-cross}. Other approaches adapt attention-based importance for dynamic skipping or progressive removal in LLM layers~\cite{feather-prune-cross, atpllava-prune-cross, twigvlm-prune-cross, aim-prune-cross, skipvision-prune-cross, topv-prune-cross}. Although effective, these methods extract cross-attention inside the LLM, disabling FlashAttention, fragmenting the KV cache, and yielding \emph{only} prompt-dependent sparsity that frequently overlooks salient visual regions.

\noindent \textbf{Cross-modal-only Pruning.}
This line of work leverages \emph{text--vision cross-attention} extracted from within the LLM to rank and prune patches. 
Approaches range from early-layer ranking~\cite{fastv-prune-cross, sparsevlm-prune-cross} and multi-stage or collaborative aggregation~\cite{pyramiddrop-prune-cross, fitprune-prune-cross, zipvl-prune-cross, meteor-prune-cross} to progressive removal and dynamic skipping across LLM layers~\cite{feather-prune-cross, atpllava-prune-cross, twigvlm-prune-cross, aim-prune-cross, skipvision-prune-cross, topv-prune-cross}. 
While query-aware, extracting signals \emph{inside} the LLM disables FlashAttention optimizations, fragments the KV cache, and yields sparse, prompt-biased scores that often overlook visually salient regions.

%\noindent \textbf{Training-based Token Pruning.}
%A separate line of work integrates token pruning into VLM training, where pruning 
%modules or compression adapters are optimized jointly with the model~\cite{matryoshka-prune-train, matryoshka2-prune-train, llavolta-prune-train, tokenpacker-prune-train, llamavid-prune-train, llavamini-prune-train}.  
%These can maintain robustness under aggressive compression, but require additional training and are not plug-and-play. 
% Existing visual token pruning methods rely on either vision-only saliency, which is efficient but query-agnostic and prone to attention-sink effects, or text–vision saliency extracted inside the LLM, which incurs system-level inefficiencies, sparse and position-biased scores, and fragmented KV caches. 

\textbf{ConsensusDrop}  overcomes these limitations by fusing query-aware cross-modal signals \emph{before} the LLM with vision-side saliency, enabling training-free, plug-and-play token reduction while retaining inference efficiency.

%% file: sec/3_motivation.tex
\section{Motivation}
\label{sec:motiv}

\subsection{Do We Need Both Vision and Cross-modal Cues?}
% We hypothesize that neither the vision encoder nor the language model alone provides a reliable estimate of visual token importance. To study this, we design a controlled setting where both modalities independently score the same $N$ image patches.

% We evaluate the overlapping of top-K tokens with highest score from Vision side and Cross-modal side.
% To examine whether one modality can correct the other, we perform a simple recovery procedure, which discards the one modality's lowest-ranked tokens and replaces the other with the teacher's highest-ranked candidates.

We hypothesize that neither the vision encoder nor the language model alone provides a reliable estimate of visual token importance. To study this, we design a controlled setting where both modalities independently score the same $N$ image patches. We first evaluate the \emph{agreement}, which represents overlap between the top-$K$ tokens derived from vision and cross-modal scores. To examine whether one modality can correct the other, we perform a simple \emph{recovery} procedure: we discard the lowest-ranked tokens from one modality (the student) and replace them with the highest-ranked candidates from the other (the teacher). This controlled setup isolates the contribution of visual versus cross-modal saliency and reveals how often each modality captures information that the other misses, motivating our approach.

% \noindent\textbf{Complementary \& Asymmetric Signals.}
The controlled setting reveals that (Sec. \ref{motiv:analysis}) (i) vision- and cross-modal saliency produce non-trivial ranking disagreements, and (ii) the signals are \emph{asymmetric}: vision saliency provides a stronger base ranking, while cross-modal cues primarily contribute \emph{targeted corrections}. This motivates multimodal fusion with an explicit \emph{vision bias} as our default design (Sec.~\ref{sec:fuser}).

\subsection{Problem Formulation}
\label{sec:setup}

\noindent\textbf{VLM Formulation.} 
Formally, a VLM consists of a frozen vision encoder $\mathcal{E}_v$, a projector $\mathcal{P}$, and a frozen LLM $\mathcal{L}$. Given an image $\mathcal{I}$ and query $\mathbf{T} \in \mathbb{R}^{M \times d}$, the image is projected into visual tokens $\tilde{\mathbf{V}} \in \mathbb{R}^{N \times d}$. The model processes the concatenated sequence $\mathbf{X} = [\mathbf{X}^{\text{sys}}, \tilde{\mathbf{V}}, \mathbf{T}]$. Since $N \gg M$, our target is to compress $\tilde{\mathbf{V}}$ into a compact subset \emph{before} LLM inference.

\noindent\textbf{Scores and Selection.} Vision-side scores $t_v \in \mathbb{R}^N$ are obtained from the penultimate self-attention layer of the vision encoder using CLS-based attention. Cross-modal scores $t_c \in \mathbb{R}^N$ are extracted from a designated transformer layer of the LLM by aggregating attention from all query tokens to each visual token. This follows the setting of visual token pruning as in FastV \cite{fastv-prune-cross}. For this controlled study, to ensure a fixed budget, we define a \emph{Retention Ratio $\rho$} and set the selection count $K = \rho N $.  For each score vector, we take the top-$K$ patch indices, denoted $I_v(K)$ and $I_c(K)$. 

% \noindent\textbf{Agreement and Disagreement:} The overlap fraction between $I_v(K)$ and $I_c(K)$,  $\mathrm{agreement}(K) = |I_v(K) \cap I_c(K)| / K$, 
% % \begin{equation}
% % \mathrm{agreement}(K) = |I_v(K) \cap I_c(K)| / K,
% % \end{equation}
% measures how consistently the two modalities identify important regions; the disagreement ratio is $1 - \mathrm{agreement}(K)$.
\noindent\textbf{Agreement and Disagreement:} We define \emph{agreement} as the overlap $\mathrm{agreement}(K) = |I_v \cap I_c| / K$, measuring how consistently the two modalities identify important regions; the disagreement ratio is $1 - \mathrm{agreement}(K)$.
    
\noindent\textbf{Recovery and Correction:} 
We perform the recovery (Sec. 3.1) by treating one set from $I_v(K)$ and $I_c(K)$ as the student $I_s$ and the other as the teacher $I_t$. For a recovery rate $r$, we drop the lowest $rK$ tokens from $I_s(K)$ and replace them with the highest-ranked $rK$ tokens from $I_t(K)$ that do not already appear in the retained student tokens. Let $\ell$ be the deepest teacher rank we must inspect to find $rK$ such non-overlapping tokens; we define a correction rate as $(K-\ell)/(K-rK)$, which lies in $[0,1]$ and reflects how strongly the teacher disagrees with the student's ranking.

\subsection{Analysis}
\label{motiv:analysis}

% \noindent \textbf{Accuracy Gains with Multimodal Fusion.}
% % Across six established benchmarks, we evaluate 
% % define the metric is the end to end accuracy
% the end-to-end accuracy of 
% LLaVA-1.5-7B under cross-only, vision-only, and multimodal recovery token selection (as described above). Fig.~\ref{fig:accuracy-recovery} shows that recovery \emph{consistently boosts accuracy} at all retention ratios, confirming that joint visual and Cross-modal cues provide a more faithful estimate of token importance than either signal alone. Interestingly, Cross-modal cues perform the worst -- this suggests that while \emph{both} visual \emph{and} Cross-modal cues outperform either modality individually, visual cues \emph{help} the Cross-modal cues more than vice-versa. This aligns with the conclusion drawn by several prior Vision-only pruning methods \cite{visionzip-prune-viz, vispruner-prune-viz, dynamictokeneff-prune-viz}. 

% \noindent\emph{Takeaway.}
% Vision-side saliency provides a strong base importance ranking for visual tokens,
% while Cross-modal cues primarily correct a small but important subset.
% This suggests combining two signals via a \emph{soft consensus}
% mechanism that is biased toward visual saliency (Sec.~\ref{sec:fuser}).

\begin{figure}
    \centering
    \includegraphics[width=\columnwidth]{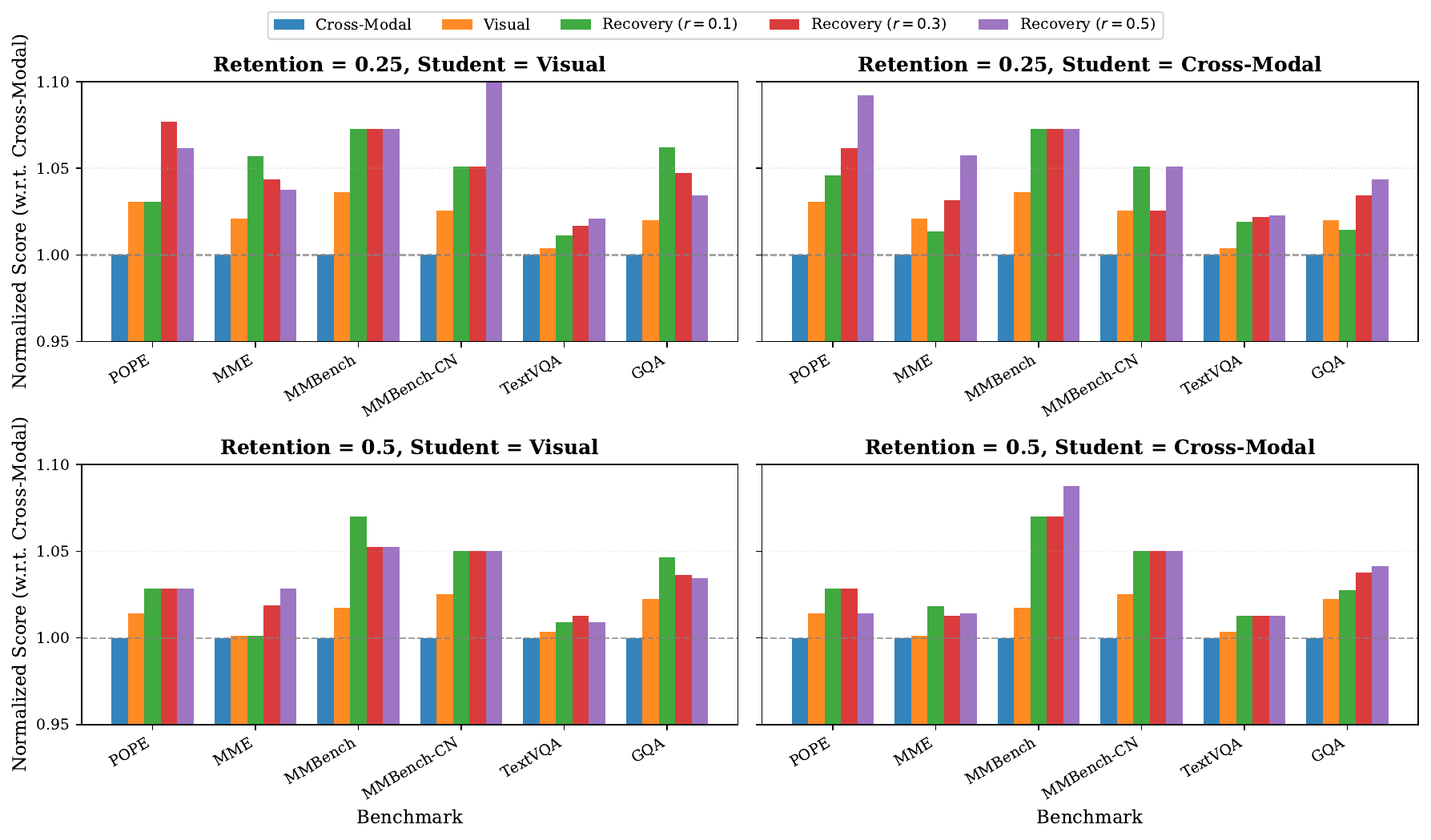}
    \caption{
    \textbf{Multimodal recovery consistently improves accuracy across benchmarks.}
    Using LLaVA-1.5-7B, we compare token selection based on (i) cross-modal scores, (ii) vision scores, and (iii) our recovery mechanism that fuses the two.  
    Across POPE, MME, MMBench, MMBench-CN, TextVQA, and GQA, recovery achieves higher accuracy at all retention ratios (0.25 and 0.50) and for both student configurations (visual and cross-modal).  
    These validate that \emph{neither unimodal signal is sufficient}, and that combining them provides a more reliable estimate of token importance.
    }
    \label{fig:accuracy-recovery}
\end{figure}

\noindent \textbf{Accuracy Gains with Multimodal Fusion.}
We evaluate LLaVA-1.5-7B accuracy across six benchmarks (Fig.~\ref{fig:accuracy-recovery}). Recovery \emph{consistently boosts accuracy} at all retention ratios, confirming that joint cues estimate token importance better than either signal alone. 
Notably, vision scores generally outperform cross-modal scores alone. This implies an \emph{asymmetric synergy}: visual cues provide the foundation, helping cross-modal cues more than vice-versa, aligning with prior vision-only pruning findings \cite{visionzip-prune-viz, vispruner-prune-viz, dynamictokeneff-prune-viz}.

\noindent\emph{Takeaway.}
Vision saliency provides a strong base ranking, while cross-modal cues primarily correct a small, critical subset. This motivates a \emph{vision-biased} consensus mechanism (Sec.~\ref{sec:fuser}).

% \noindent \textbf{Disagreement-Correction-Accuracy Mechanism.}
% To understand \emph{why} consensus-based recovery improves accuracy, we analyze the joint relationship between modal disagreement, correction rate, and accuracy gain (Fig.~\ref{fig:three-way}). A clear pattern emerges. First, higher disagreement between visual and Cross-modal saliency creates more opportunities for meaningful corrections. Second, these corrections directly translate into non-trivial accuracy improvements. Finally, aggressive pruning (low retention) amplifies both disagreement and correction, yielding the largest gains. Together, this reveals a simple trend: disagreement due to aggressive pruning enables more opportunities for Cross-modal correction, which improves VLM accuracy on downstream tasks. This explains why multimodal fusion consistently outperforms either modality in isolation.

\begin{figure}
    \centering
    \includegraphics[width=\columnwidth]{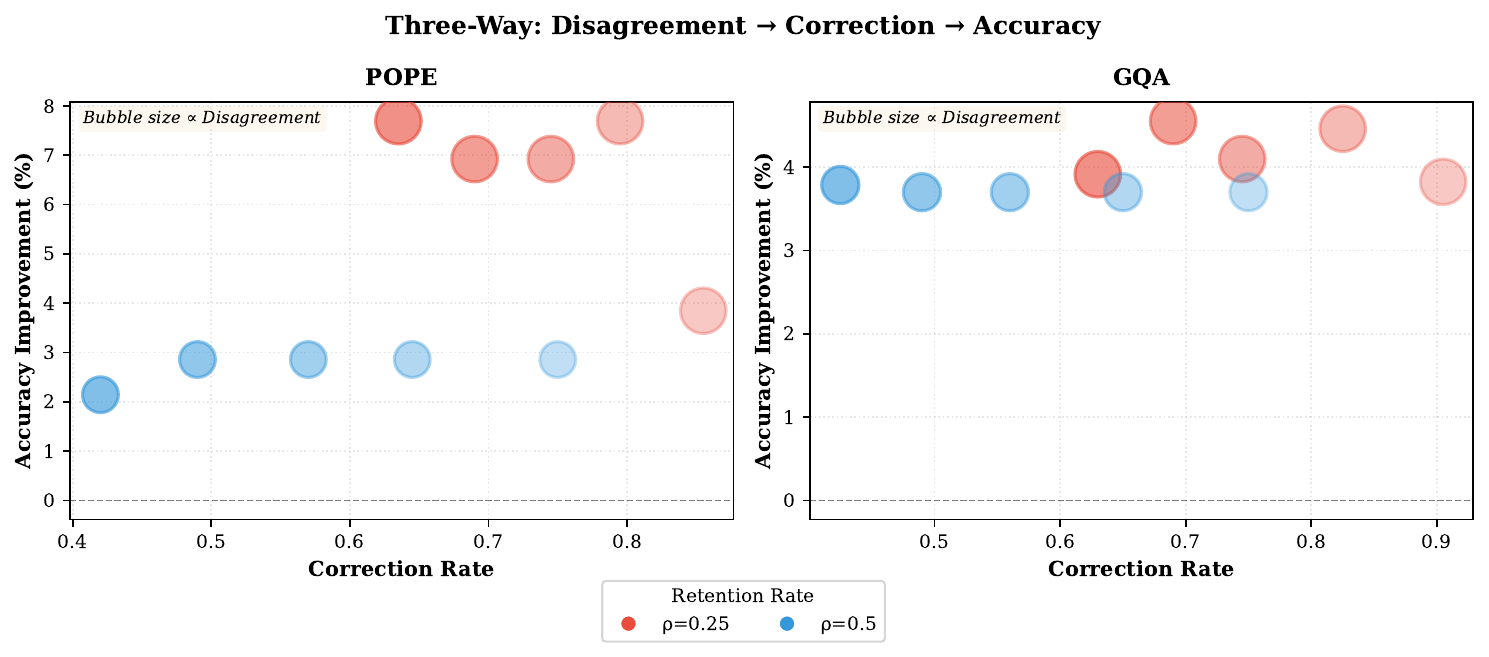}
    \vspace{-2mm}
    \caption{
        \textbf{Three-way relationship between disagreement, correction rate, and accuracy improvement.}
        Bubble size denotes disagreement between visual and cross-modal attention;
        color distinguishes retention ratio ($\rho=0.25$ vs.\ $\rho=0.5$).
        Higher disagreement consistently leads to higher correction, which yields greater accuracy gains. Low-retention (red) exhibit
        the strongest effect, confirming that aggressive pruning amplifies
        productive modal disagreements that consensus recovery can leverage.
    }
    \label{fig:three-way}
    \vspace{-2mm}
\end{figure}

% \noindent\emph{Takeaway.}
% Accuracy gains arise from correcting a small subset of visual tokens where
% vision and Cross-modal saliency disagree. These targeted corrections account
% for most of the improvement, indicating that Cross-modal cues are most
% valuable when applied selectively. 
% %rather than uniformly.
% (Sec. \ref{sec:fuser}).
\noindent \textbf{Disagreement-Correction-Accuracy Mechanism.}
Fig.~\ref{fig:three-way} reveals a clear causal chain: aggressive pruning (low retention) amplifies modal disagreement, creating opportunities for valid cross-modal corrections that directly translate into accuracy gains. 
Essentially, disagreement acts as a signal, highlighting the specific "blind spots" where cross-modal intervention is most effective.

\noindent\emph{Takeaway.}
Since improvements stem from correcting specific disagreements, cross-modal cues are most valuable when applied \emph{selectively} to these conflict regions rather than uniformly (Sec. \ref{sec:fuser}).

% \noindent \textbf{Correction-Rate Decay.}
% We analyze how the correction rate changes with the recovery rate $r$ (Fig.~\ref{fig:correction-decay}). At small $r$ (0.1), correction rates are very high (80--90\%), showing that early recoveries target genuinely disputed tokens. As $r$ increases, correction rates drop sharply (to 40--50\%), revealing a \emph{consensus-saturation} effect: only a few dropped tokens meaningfully disagree between modalities, and recovering beyond this set reintroduces largely redundant tokens. This highlights that consensus recovery selectively captures the truly informative disagreements between modalities, and additional recovery quickly saturates with low-value tokens.

% \noindent\emph{Takeaway.}
% Beyond this small set of disputed tokens, additional recoveries yield
% diminishing returns and reintroduce largely redundant information. This
% suggests that remaining low-saliency tokens should be compactly represented
% rather than retained individually, motivating token merging as an efficient
% alternative to further selection (Sec. \ref{sec:egtm}).

\begin{figure}
    \centering
    \includegraphics[width=\columnwidth]{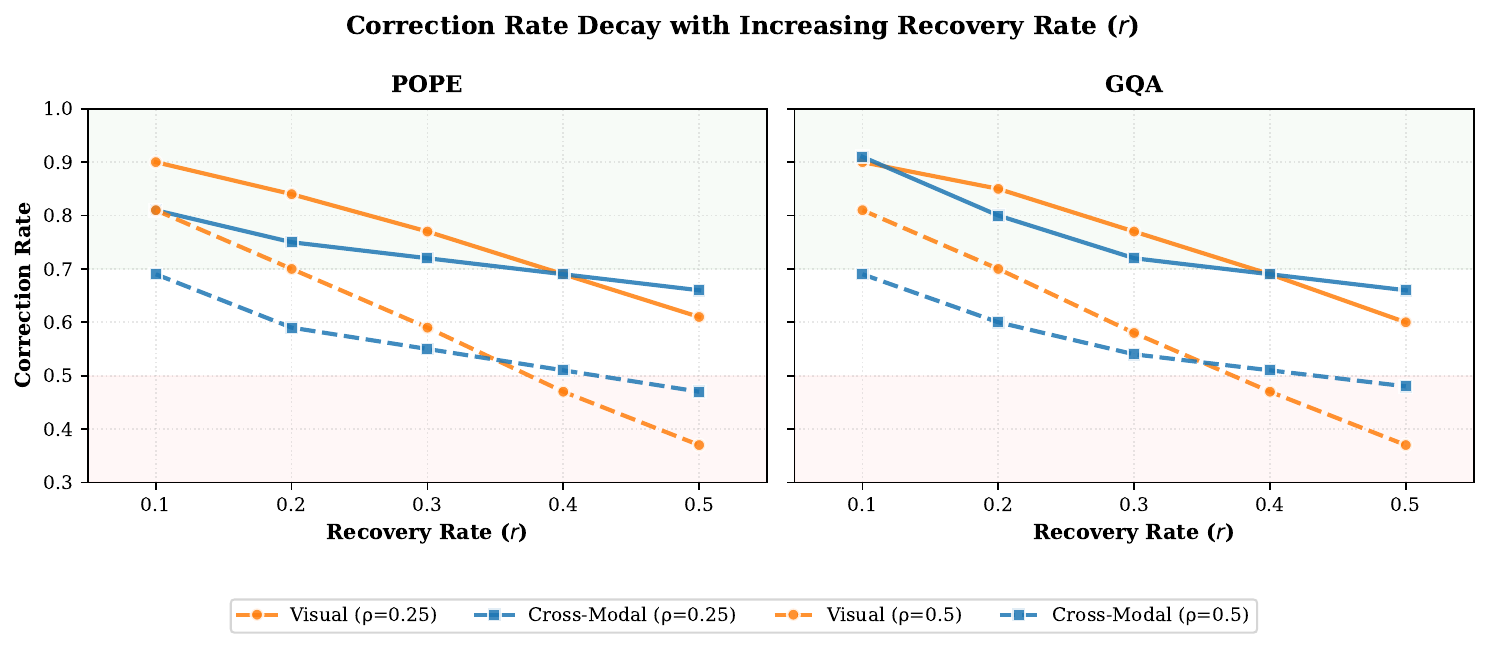}
    \vspace{-6pt}
    \caption{\textbf{Correction-rate decay} as the recovery rate $r$ increases
    for POPE and GQA.  
    Correction rates are highest at small $r$, where nearly all recovered
    tokens correspond to true multimodal disagreements. As $r$ increases, the
    benefit of recovery saturates: additional tokens reintroduced are
    increasingly ones that both modalities already agreed upon. This reveals
    a strong quality--quantity tradeoff and shows that consensus-based
    recovery naturally prioritizes a small set of high-value corrections.}
    \label{fig:correction-decay}
    \vspace{-6pt}
\end{figure}

\noindent \textbf{Correction-Rate Decay.}
In Fig.~\ref{fig:correction-decay}, correction rates start high (80--90\% at $r=0.1$), showing early recoveries target genuine disputed tokens. Rates drop sharply (to 40--50\%) as $r$ increases, revealing a \emph{consensus-saturation} effect: only a few dropped tokens meaningfully disagree between modalities, and recovering beyond this set reintroduces largely redundant tokens.

\noindent\emph{Takeaway.}
Since further recovery yields diminishing returns, the remaining low-saliency tokens should be \emph{merged} rather than retained individually, motivating Encoder-Guided Token Merge (Sec. \ref{sec:egtm}).

\noindent\textbf{What Consensus Means.}
Throughout this paper, we use \emph{consensus} to denote agreement between vision-side and cross-modal saliency signals after reconciling their complementary roles: vision provides a strong base ranking, while cross-modal cues introduce targeted, query-dependent corrections.
% Consensus does not imply equal weighting, but a principled mechanism for allowing both signals to contribute to the final set of important visual tokens.
 Rather than equal weighting, it implies a principled integration of both signals.

% While this analysis confirms the potential of consensus, the core challenge is to estimate query-conditioned importance \emph{without} executing a full multimodal forward pass, ensuring the process remains lightweight and training-free
% This motivates \textbf{ConsensusDrop} (Sec.~\ref{sec:met}), which approximates this consensus using a lightweight, training-free probe \emph{before} the LLM, enabling efficient inference.

%% file: sec/4_method.tex
\section{Method}
\label{sec:met}

%%%% COMMENT: Update title

\begin{figure*}
    \centering
    \includegraphics[width=0.8\linewidth]{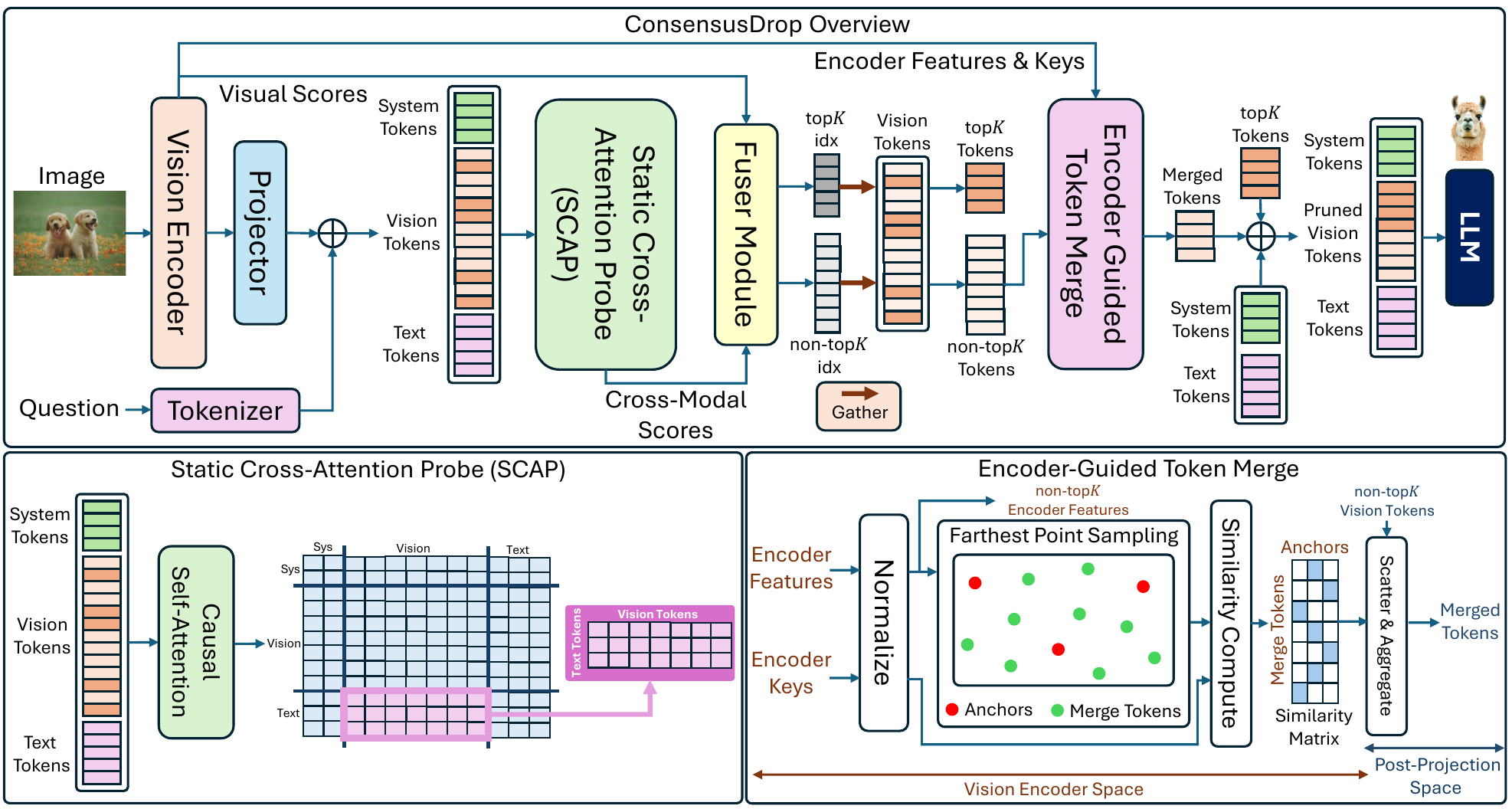}
    \caption{\textbf{Illustration of ConsensusDrop.} Given an image--question pair,
    a frozen vision encoder and projector produce visual tokens, and SCAP
    provides query-aware cross-attention scores. A lightweight fuser combines
    vision and cross-modal saliency to select top-$K$ tokens, while EGTM merges
    the remaining tokens into compact representations, yielding a compressed
    visual sequence for the LLM.}
    \label{fig:SystemOverview}
\end{figure*}

% Building on our empirical findings in Sec.~\ref{sec:motiv}, we now present
% \emph{ConsensusDrop}, a training-free visual token compression framework
% that fuses vision and cross-modal saliency. Our method has three main
% components: a Static Cross-Attention Probe (SCAP), a multimodal fuser, and
% an Encoder-Guided Token Merge (EGTM) module.

Building on Sec.~\ref{sec:motiv}, we present \emph{ConsensusDrop},  a training-free visual token compression framework fusing vision and cross-modal saliency via three training-free components: the Static Cross-Attention Probe (SCAP), a Multimodal Fuser, and Encoder-Guided Token Merge (EGTM). In Fig.~\ref{fig:SystemOverview}, SCAP first extracts cross-modal attention \emph{before} the LLM. The Fuser then combines these with vision-side saliency to select the top-$K$ tokens, while EGTM uses the encoder's feature geometry to compactly merge the remaining low-saliency tokens in the projector space, yielding a simplified yet information-preserving sequence (Algorithm~\ref{alg:consensusdrop}).

% \subsection{Method Overview}

% ConsensusDrop is a training-free framework for reducing visual tokens
% before they enter the LLM. As shown in Fig.~\ref{fig:SystemOverview}, we
% first compute two complementary saliency signals: (i) vision-side attention
% from the frozen vision encoder, and (ii) cross-modal attention from a Static
% Cross-Attention Probe (SCAP). A lightweight \textbf{fuser} combines these
% scores to select the top-$K$ most important tokens.

% The remaining low-saliency tokens are not discarded. An
% \textbf{Encoder-Guided Token Merge (EGTM)} module uses the encoder’s feature
% geometry to cluster and merge them into $M$ compact representatives in the
% projector space. The final visual sequence is the union of the preserved
% top-$K$ tokens and the merged tokens, yielding a shorter yet
% information-preserving input to the LLM. Algorithm~\ref{alg:consensusdrop} summarizes the full ConsensusDrop pipeline.

\subsection{Static Cross-Attention Probe (SCAP)}
\label{sec:scap}

SCAP is a lightweight, training-free probe that estimates query-conditioned
saliency for each visual token \emph{before} the main LLM forward pass.
Concretely, SCAP replicates the input LayerNorm and multi-head self-attention
module from the first decoder layer of the LLM, and applies it once to the
full multimodal input sequence
$\mathbf{X}=[\mathbf{X}^{\text{sys}};\tilde{\mathbf{V}};\mathbf{T}]$.
Let $S = |\mathbf{X}^{\text{sys}}| + N + L$ denote the total sequence length.
The resulting head-averaged attention matrix
$\tilde{\mathbf{A}}\in\mathbb{R}^{S\times S}$ is
\begin{equation}
\tilde{\mathbf{A}} = \frac{1}{H}\sum_{h=1}^{H}
\mathrm{Softmax}\!\left(\frac{\mathbf{Q}^{(h)}(\mathbf{K}^{(h)})^\top}{\sqrt{d_h}}\right),
\end{equation}
where Softmax is applied row-wise and the same causal mask as the LLM is used.

% We extract the text-to-vision block
% $\mathbf{A}=\tilde{\mathbf{A}}[\mathcal{I}_{\text{text}},\mathcal{I}_{\text{vis}}]
% \in\mathbb{R}^{L\times N}$,
% where $\mathbf{A}_{i,j}$ denotes the attention from text token $t_i$ to visual
% token $\tilde{v}_j$. We aggregate $\mathbf{A}$ into cross-modal saliency scores
% $\mathbf{s}^{(c)}\in\mathbb{R}^{N}$ using default strategies, \emph{last-token:} $\mathbf{s}^{(c)}=\mathrm{Norm}(\mathbf{A}_{L,:})$;
% (see \textbf{Appendix~\ref{app:scap_strategies}} for other strategies).
%(ii) \emph{all-token:} row-normalize each text token’s attention over visual
% tokens and then average across text tokens; and (iii) \emph{max-token:}
% $\mathbf{s}^{(c)}=\mathrm{Norm}(\max_{i}\mathbf{A}_{i,:})$.
% Here $\mathrm{Norm}(\mathbf{z})=\mathbf{z}/(\sum_j z_j+\epsilon)$ ensures a
% valid distribution over visual tokens. 

We extract the text-to-vision block $\mathbf{A}=\tilde{\mathbf{A}}[\mathcal{I}_{\text{text}},\mathcal{I}_{\text{vis}}] \in\mathbb{R}^{L\times N}$, where $\mathbf{A}_{i,j}$ denotes the attention from text token $t_i$ to visual token $\tilde{v}_j$. 
We aggregate $\mathbf{A}$ into cross-modal saliency scores $\mathbf{s}^{(c)}\in\mathbb{R}^{N}$ using the \emph{all-token} strategy by default: 
$\mathbf{s}^{(c)} = \frac{1}{L} \sum_{i=1}^{L} \mathrm{Norm}(\mathbf{A}_{i,:})$ 
(see \textbf{Appendix~\ref{app:scap_strategies}} for other variants).

\subsection{Fuser Module}
\label{sec:fuser}

Given an input image, we obtain two complementary saliency vectors over the $N$ visual tokens: a vision-side score and a cross-modal score. From the frozen vision encoder, we extract vision-side saliency
$\mathbf{s}^{(v)} \in \mathbb{R}^N$ using \emph{either} (i) CLS-to-patch attention or (ii) the aggregate attention \emph{received} by each patch from all other patches, both computed from the penultimate self-attention layer of the vision encoder (with attention scores aggregated across heads). SCAP (Sec.~\ref{sec:scap}) produces the query-conditioned cross-modal saliency $\mathbf{s}^{(c)} \in \mathbb{R}^N$.

\noindent \textbf{Temperature Scaling.}
Before fusion, both saliency vectors are converted into normalized importance distributions using temperature scaling. For modality $m \in \{v,c\}$, we define
\begin{equation}
\hat{\mathbf{s}}^{(m)}_j =
\frac{\left(\mathbf{s}^{(m)}_j\right)^{1/\tau_m}}
     {\sum_{k=1}^{N} \left(\mathbf{s}^{(m)}_k\right)^{1/\tau_m}},
\qquad j = 1,\dots,N,
\end{equation}
where $\tau_m > 0$ controls distribution sharpness.
Larger temperatures yield flatter distributions, while smaller temperatures emphasize high-scoring tokens (sharper distribution).

\noindent \textbf{Convex Fuser.}
Our default Fuser combines the two normalized distributions via a convex
combination:
\begin{equation}
\hat{\mathbf{s}} = \alpha\,\hat{\mathbf{s}}^{(v)} + (1-\alpha)\,\hat{\mathbf{s}}^{(c)},
\end{equation}
where $\alpha \in [0,1]$ controls the vision--cross-modal balance.
Motivated by the asymmetric behavior observed in Sec.~\ref{sec:motiv},
we bias fusion toward vision saliency and use $\alpha > 0.5$
(default $\alpha=0.7$) unless otherwise specified.
Tokens are ranked according to $\hat{\mathbf{s}}$; the top-$K$ indices
$\mathcal{I}_{\text{top}}$ are preserved, and the remaining
$\mathcal{I}_{\text{non}} = [N] \setminus \mathcal{I}_{\text{top}}$
are passed to EGTM for compression.
The fused distribution $\hat{\mathbf{s}}$ is only defined for this convex fuser.
(See \textbf{Appendix~\ref{app:fuser_variants}} for alternative fusion mechanisms).

% \noindent \textbf{Recovery Fuser.}
% We additionally consider a recovery-based fuser inspired by the controlled
% analysis in Sec.~\ref{sec:motiv}.
% One modality is treated as a \emph{student} and the other as a \emph{teacher}.
% Starting from the student’s top-$K$ tokens, we drop the lowest $rK$ tokens
% and replace them with the highest-ranked, non-overlapping tokens from the
% teacher, where $r \in [0,1]$ is a recovery rate.
% This mechanism operates directly on ranked indices rather than fused
% distributions.
% Unless stated otherwise, we use the convex fuser as the default and employ
% recovery fusion primarily to validate the analysis in Sec.~\ref{sec:motiv}.

% \begin{figure*}
%     \centering
%     \includegraphics[width=\linewidth]{Figure/fusion-full.pdf}
%     \caption{\textbf{Fuser module designs.}
%     The Fuser module has two variants: \textbf{Convex} and \textbf{Recovery}.
%     \emph{Left:} the Convex Fuser temperature-normalizes vision and cross-modal
%     scores and forms a convex combination with weight $\alpha$ to obtain a
%     unified importance distribution.
%     \emph{Right:} the Recovery Fuser treats cross-modal scores as the student
%     and vision scores as the teacher, selecting top-$K$ tokens from the student
%     and recovering a small set of high-ranked teacher tokens based on a
%     recovery rate.}
%     \label{fig:fuser}
% \end{figure*}

\subsection{Encoder-Guided Token Merging (EGTM)}
\label{sec:egtm}

After the Fuser selects the top-$K$ indices $\mathcal{I}_{\text{top}}$,
we compress the remaining $R = N - K$ visual tokens
$\mathcal{I}_{\text{non}} = [N] \setminus \mathcal{I}_{\text{top}}$
into $M$ merged tokens, producing a final visual sequence of length $K+M$.
EGTM is \emph{encoder-guided}: structural decisions (anchor selection and token
assignment) are computed using features from the frozen vision encoder, while
feature aggregation is performed in the post-projector space to maintain
compatibility with the LLM.

\noindent\textbf{Anchor Selection via FPS.}
Let $\mathbf{V} \in \mathbb{R}^{N \times d_v}$ denote the penultimate-layer
vision encoder patch features (pre-projector), and define
$\mathbf{V}_{\text{non}} = \mathbf{V}[\mathcal{I}_{\text{non}}] \in \mathbb{R}^{R \times d_v}$.
We $\ell_2$-normalize $\mathbf{V}_{\text{non}}$ and select $M$ diverse anchor
tokens using farthest point sampling (FPS) \cite{pointnetpp-fps}, yielding a set
of anchor indices $\mathcal{A} \subseteq \mathcal{I}_{\text{non}}$ with
$|\mathcal{A}| = M$. The remaining indices are denoted by
$\mathcal{U} = \mathcal{I}_{\text{non}} \setminus \mathcal{A}$.

\noindent\textbf{Hard Assignment via Encoder Keys.}
Let $\mathbf{K} \in \mathbb{R}^{H_v \times N \times d_k}$ denote the key tensors
from the same penultimate vision encoder block (one per attention head).
We construct a head-averaged, $\ell_2$-normalized key embedding per patch,
resulting in
$\mathbf{G} \in \mathbb{R}^{N \times d_k}$.
Restricting to non-top-$K$ tokens gives
$\mathbf{G}_{\text{non}} = \mathbf{G}[\mathcal{I}_{\text{non}}] \in \mathbb{R}^{R \times d_k}$. Each non-anchor (merge) token $u \in \mathcal{U}$ is assigned to the most similar anchor in this key space via a hard nearest-anchor rule: $a^\star(u) = \arg\max_{a \in \mathcal{A}} \;\langle \mathbf{g}_u, \mathbf{g}_a \rangle$, 
where $\mathbf{g}_u$ and $\mathbf{g}_a$ are rows of $\mathbf{G}_{\text{non}}$.
This dot-product is used solely for \emph{assignment}, yielding a discrete
partition of non-top-$K$ tokens around the selected anchors.

\noindent\textbf{Merging in Projector Space.}
Let $\tilde{\mathbf{V}} \in \mathbb{R}^{N \times d}$ denote the projected visual
tokens fed to the LLM.
For each anchor $a \in \mathcal{A}$, we form one merged token by averaging its
own projected token with all projected tokens assigned to it:
\begin{equation}
\tilde{\mathbf{v}}^{\text{merge}}_{a}
=
\frac{1}{|\mathcal{C}(a)|}
\sum_{j \in \mathcal{C}(a)} \tilde{\mathbf{v}}_j ,
\end{equation}
where $\mathcal{C}(a) = \{a\} \cup \{u \in \mathcal{U} : a^\star(u) = a\}.$
Finally, EGTM outputs the reduced visual sequence
\begin{equation}
\tilde{\mathbf{V}}_{\text{final}}
=
\mathrm{Concat}
\big(
\tilde{\mathbf{V}}[\mathcal{I}_{\text{top}}],
\{\tilde{\mathbf{v}}^{\text{merge}}_{a}\}_{a \in \mathcal{A}}
\big)
\in \mathbb{R}^{(K+M) \times d}.
\end{equation}

\noindent\textbf{Multimodal Design Rationale.}
EGTM explicitly decouples \emph{where} structure is inferred from \emph{where}
compression is applied: the vision encoder provides a semantically meaningful
metric space for selecting anchors and assigning tokens, while the final merging
is performed in the LLM-aligned projector space.
This design enables aggressive visual token compression while preserving the
multimodal representations consumed by the language model.

\begin{table*}[t!]
  \centering
  \resizebox{\linewidth}{!}{%
  \begin{tabular}{lccccccccccc|ccc}
    \toprule
    \textbf{Method} &
    \textbf{$\text{VQA}^\text{V2}$} & \textbf{GQA} &
    \textbf{$\text{SQA}^\text{IMG}$} & \textbf{$\text{VQA}^\text{Text}$} &
    \textbf{POPE} & \textbf{MME} & \textbf{MMB} &
    \textbf{$\text{MMB}^\text{CN}$} & \textbf{MMVet} &
    \textbf{Acc} & \textbf{Rel} &
    \textbf{TTFT (ms)} & \textbf{TPOT (ms)} & \textbf{KV (MB)} \\
    \midrule
    \rowcolor{lightgray!30}
    \multicolumn{15}{c}{\textit{Upper bound: 576 tokens (100\%)}} \\
    LLaVA-1.5-7B &
      78.5 & 62.0 & 66.8 & 58.2 & 85.9 & 1510.7 &
      64.3 & 58.3 & 31.1 & 64.5 & 100\% &
      106.4 & 27.7 & 321 \\
    
    \midrule
    \rowcolor{lightgray!30}
    \multicolumn{15}{c}{\textit{Retain 192 tokens ($\downarrow 66.7\%$)}} \\
    ToMe \cite{bolya2022tome} &
      68.0 & 54.3 & 65.2 & 52.1 & 72.4 & 1148.8 &
      60.5 & 50.2 & 27.8 & 56.4 & 87.4\% &
      --- & --- & --- \\
    FastV \cite{fastv-prune-cross} &
      67.1 & 52.6 & \textbf{69.1} & 52.5 & 64.8 & 1252.1 &
      61.0 & 52.4 & 28.5 & 56.7 & 87.9\% &
      78.0 (1.36$\times$) & 26.0 (1.07$\times$) & 144 ($\downarrow$55\%) \\
    SparseVLM \cite{sparsevlm-prune-cross} &
      \underline{77.0} & 59.5 & 68.7 & \textbf{57.8} & 85.3 & 1395.4 &
      \textbf{64.1} & 52.1 & 30.2 & 62.7 & 97.2\% &
      86.6 (1.23$\times$) & 38.3 (0.72$\times$) & \textbf{119 ($\downarrow$63\%)} \\
    VisionZip \cite{visionzip-prune-viz} &
      76.8 & 59.3 & 68.9 & 57.3 & 85.3 & 1443.6 &
      63.0 & 56.9 & 31.7 & 63.5 & 98.4\% &
      \textbf{69.5 (1.53$\times$)} & \textbf{25.5 (1.09$\times$)} & 129 ($\downarrow$60\%) \\
    VisPruner \cite{vispruner-prune-viz} &
      76.9 & \underline{59.6} & 68.7 & \underline{57.6} & \textbf{86.0} & \underline{1468.4} &
      63.7 & \underline{57.4} & \textbf{33.3} & \underline{64.1} & \underline{99.4\%} &
      \underline{73.0 (1.46$\times$)} & 26.2 (1.06$\times$) & 129 ($\downarrow$60\%) \\
    \textbf{ConsensusDrop} &
      \textbf{77.1} & \textbf{60.4} & \underline{69.0} &
      57.5 & \underline{85.5} & \textbf{1476.2} &
      \underline{63.8} & \textbf{58.0} & \underline{33.1} &
      \textbf{64.3} & \textbf{99.7\%} &
      80.4 (1.32$\times$) & \underline{26.2 (1.06$\times$)} & \underline{129 ($\downarrow$60\%)} \\
    
    \midrule
    \rowcolor{lightgray!30}
    \multicolumn{15}{c}{\textit{Retain 128 tokens ($\downarrow 77.8\%$)}} \\
    ToMe \cite{bolya2022tome} &
      63.0 & 52.4 & 59.6 & 49.1 & 62.8 & 1088.4 &
      53.3 & 48.8 & 27.2 & 52.3 & 81.1\% &
      --- & --- & --- \\
    FastV \cite{fastv-prune-cross} &
      61.8 & 49.6 & 60.2 & 50.6 & 59.6 & 1208.9 &
      56.1 & 51.4 & 28.1 & 53.1 & 82.3\% &
      72.1 (1.48$\times$) & 25.8 (1.07$\times$) & 117 ($\downarrow$64\%) \\
    SparseVLM \cite{sparsevlm-prune-cross} &
      73.8 & 56.0 & 67.1 & 54.9 & 80.5 & 1376.2 &
      60.0 & 51.1 & 30.0 & 60.2 & 93.3\% &
      83.0 (1.28$\times$) & 37.7 (0.74$\times$) & \textbf{87 ($\downarrow$73\%)} \\
    PruMerge+ \cite{shang2024llavaprumerge} &
      74.7 & 57.8 & 67.6 & 54.3 & 81.5 & 1420.5 &
      61.3 & 54.7 & 28.7 & 61.3 & 95.0\% &
      --- & --- & --- \\
    VisionZip \cite{visionzip-prune-viz} &
      75.6 & 57.6 & 68.9 & 56.8 & 83.2 & 1432.4 &
      62.0 & 56.7 & 32.6 & 62.8 & 97.4\% &
      \textbf{60.2 (1.77$\times$)} & \textbf{25.3 (1.10$\times$)} & 97 ($\downarrow$70\%) \\
    VisPruner \cite{vispruner-prune-viz} &
      \underline{75.8} & \underline{58.2} & \underline{69.1} & \underline{57.0} & \underline{84.6} & \underline{1461.4} &
      \underline{62.7} & \underline{57.3} & \underline{33.7} & \underline{63.5} & \underline{98.4\%} &
      72.5 (1.47$\times$) & 26.2 (1.06$\times$) & 97 ($\downarrow$70\%) \\
    \textbf{ConsensusDrop} &
      \textbf{76.1} & \textbf{60.2} & \textbf{69.3} &
      \textbf{57.1} & \textbf{85.1} & \textbf{1467.7} &
      \textbf{63.2} & \textbf{58.1} & \textbf{33.8} &
      \textbf{64.0} & \textbf{99.2\%} &
      \underline{70.6 (1.51$\times$)} & \underline{25.8 (1.07$\times$)} & \underline{97 ($\downarrow$70\%)} \\
    
    \midrule
    \rowcolor{lightgray!30}
    \multicolumn{15}{c}{\textit{Retain 64 tokens ($\downarrow 88.9\%$)}} \\
    ToMe \cite{bolya2022tome} &
      57.1 & 48.6 & 50.0 & 45.3 & 52.5 & 922.3 &
      43.7 & 38.9 & 24.1 & 45.1 & 69.9\% &
      --- & --- & --- \\
    FastV \cite{fastv-prune-cross} &
      55.0 & 46.1 & 51.1 & 47.8 & 48.0 & 1019.6 &
      48.0 & 42.7 & 25.8 & 46.2 & 71.6\% &
      71.4 (1.50$\times$) & 25.9 (1.07$\times$) & 96 ($\downarrow$70\%) \\
    SparseVLM \cite{sparsevlm-prune-cross} &
      68.2 & 52.7 & 62.2 & 51.8 & 75.1 & 1221.1 &
      56.2 & 46.1 & 23.3 & 55.2 & 85.6\% &
      84.0 (1.27$\times$) & 38.1 (0.73$\times$) & \textbf{55 ($\downarrow$83\%)} \\
    PruMerge+ \cite{shang2024llavaprumerge} &
      67.4 & 54.9 & 68.6 & 53.0 & 77.4 & 1198.2 &
      59.3 & 51.0 & 25.9 & 57.5 & 89.1\% &
      --- & --- & --- \\
    VisionZip \cite{visionzip-prune-viz} &
      72.4 & 55.1 & 69.0 & 55.5 & 77.0 & 1365.6 &
      60.1 & \underline{55.4} & \underline{31.7} & 60.5 & 93.8\% &
      \textbf{58.4 (1.82$\times$)} & \underline{25.5 (1.09$\times$)} & 65 ($\downarrow$80\%) \\
    VisPruner \cite{vispruner-prune-viz} &
      \underline{72.7} & \underline{55.4} & \underline{69.1} & \underline{55.8} & \textbf{80.4} & \underline{1369.9} &
      \underline{61.3} & 55.1 & \textbf{32.3} & \underline{61.2} & \underline{94.9\%} &
      72.0 (1.48$\times$) & 25.7 (1.08$\times$) & 65 ($\downarrow$80\%) \\
    \textbf{ConsensusDrop} &
      \textbf{73.1} & \textbf{56.2} & \textbf{70.2} &
      \textbf{56.4} & \underline{79.8} & \textbf{1372.9} &
      \textbf{61.7} & \textbf{55.8} & 31.2 &
      \textbf{61.4} & \textbf{95.2\%} &
      \underline{64.7 (1.64$\times$)} & \textbf{25.2 (1.10$\times$)} & \underline{65 ($\downarrow$80\%)} \\
    \midrule
    \rowcolor{lightgray!30}
    \multicolumn{15}{c}{\textit{Retain 32 tokens ($\downarrow 94.4\%$)}} \\
    ToMe \cite{bolya2022tome} &
      46.8 & 43.6 & 41.4 & 38.3 & 39.0 & 828.4 &
      31.6 & 28.1 & 17.3 & 36.4 & 56.4\% &
      --- & --- & --- \\
    FastV \cite{fastv-prune-cross} &
      43.4 & 41.5 & 42.6 & 42.5 & 32.5 & 884.6 &
      37.8 & 33.2 & 20.7 & 37.6 & 58.3\% &
      66 (1.61$\times$) & 26.3 (1.05$\times$) & 81 ($\downarrow$75\%) \\
    SparseVLM \cite{sparsevlm-prune-cross} &
      58.6 & 48.3 & 57.3 & 46.1 & 67.9 & 1046.7 &
      51.4 & 40.6 & 18.6 & 49.0 & 76.0\% &
      77.8 (1.37$\times$) & 36.7 (0.75$\times$) & \textbf{48 ($\downarrow$85\%)} \\
    PruMerge+ \cite{shang2024llavaprumerge} &
      54.9 & 51.1 & 68.5 & 50.6 & 70.9 & 940.8 &
      56.8 & 47.0 & 21.4 & 52.0 & 80.6\% &
      --- & --- & --- \\
    VisionZip \cite{visionzip-prune-viz} &
      67.1 & 51.8 & 68.8 & 53.1 & 68.7 & 1247.4 &
      57.7 & 50.3 & 25.5 & 56.2 & 87.1\% &
      \textbf{54.3 (1.96$\times$)} & \underline{26.1 (1.06$\times$)} & 49 ($\downarrow$85\%) \\
    VisPruner \cite{vispruner-prune-viz} &
      \underline{67.7} & \underline{52.2} & \underline{69.2} & \underline{53.9} & \underline{72.7} & \underline{1271.0} &
      \underline{58.4} & \underline{52.7} & \underline{28.8} & \underline{57.7} & \underline{89.5\%} &
      71.5 (1.49$\times$) & 26.4 (1.05$\times$) & 49 ($\downarrow$85\%) \\
    \textbf{ConsensusDrop} &
      \textbf{67.8} & \textbf{53.2} & \textbf{70.1} &
      \textbf{54.7} & \textbf{73.1} & \textbf{1275.8} &
      \textbf{59.0} & \textbf{54.1} & \textbf{30.1} &
      \textbf{58.4} & \textbf{90.5\%} &
      \underline{60.0 (1.77$\times$)} & \textbf{25.1 (1.10$\times$)} & \underline{49 ($\downarrow$85\%)} \\
    \bottomrule
  \end{tabular}}
  \caption{Performance comparison on LLaVA-1.5-7B across several retention settings. Acc is the average accuracy; Rel is relative performance vs.\ baseline. TTFT = Time To First Token with speedup in parentheses; TPOT = Time Per Output Token with speedup; KV = KV cache memory with reduction percentage. \textbf{Bold} indicates best; \underline{underline} indicates second-best.}
  \label{tab:llava-1.5-singlecol}
\end{table*}

%% file: sec/5_experiments.tex
\section{Experiments}
\label{sec:exp}

\subsection{Experimental setup}

% \noindent \textbf{Datasets.} We evaluate our method on 9 widely used image-based multi-modal benchmarks. For general visual question answering, we consider VQAv2~\cite{goyal2017vqav2}, GQA~\cite{hudson2019gqa}, ScienceQA-IMG~\cite{lu2022sqa}, and TextVQA~\cite{singh2019textvqa}. To further assess broader multi-modal understanding, we additionally adopt POPE~\cite{li2023pope}, MME~\cite{fu2024mme}, MMBench~\cite{liu2025mmbench}, MMBench-CN~\cite{liu2025mmbench}, and MM-Vet~\cite{yu2023mmvet}. We additionally evaluate on the following widely used video question answering benchmarks: TGIF-QA \cite{tgif-qa}, MSVD-QA \cite{msvd-qa}, and MSRVTT-QA \cite{msvd-qa}. All experiments follow the official evaluation protocols.
% % of the corresponding benchmarks. 

\noindent \textbf{Datasets.} We evaluate ConsensusDrop on 12 benchmarks across three distinct categories:
(1) \textbf{General VQA}: VQAv2~\cite{goyal2017vqav2}, GQA~\cite{hudson2019gqa}, ScienceQA-IMG~\cite{lu2022sqa}, and TextVQA~\cite{singh2019textvqa}; 
(2) \textbf{Multimodal Understanding}: POPE~\cite{li2023pope}, MME~\cite{fu2024mme}, MMBench/CN~\cite{liu2025mmbench}, and MM-Vet~\cite{yu2023mmvet}; 
and (3) \textbf{Video QA}: TGIF-QA~\cite{tgif-qa}, MSVD-QA~\cite{msvd-qa}, and MSRVTT-QA~\cite{msvd-qa}. 
All experiments follow official evaluation protocols.

% \noindent \textbf{Model Architectures.} We apply ConsensusDrop to several representative vision-language models (VLMs). Our main experiments are based on the LLaVA family, including LLaVA-1.5~\cite{liu2024llava1.5}, LLaVA-NeXT~\cite{liu2024llavanext} (high-resolution inputs), and Video-LLaVA \cite{lin2023videollava} (video understanding). We further evaluate on other widely used open-source VLMs, including Qwen-VL~\cite{bai2023qwenvl}, InternVL~\cite{chen2024internvl}, and CogVLM~\cite{wang2025cogvlm}. 

% \noindent \textbf{Model architectures.} We apply VisPruner to various VLM architectures, including the LLaVA series such as LLaVA-1.5~\cite{liu2024llava1.5}, LLaVA-NeXT~\cite{liu2024llavanext} for high-resolution inputs, and Video-LLaVA~\cite{lin2023videollava} for video understanding, as well as other widely used models like Qwen-VL~\cite{bai2023qwenvl}, InternVL~\cite{chen2024internvl}, and CogVLM~\cite{wang2025cogvlm}. For all models, we follow the same inference settings as the original papers.

\noindent \textbf{Model Architectures.} We apply ConsensusDrop to representative VLMs: 
(1) the \textbf{LLaVA Family} (main experiments): LLaVA-1.5~\cite{liu2024llava1.5}, LLaVA-NeXT~\cite{liu2024llavanext} (high-resolution), and Video-LLaVA~\cite{lin2023videollava}(video understanding); 
and (2) \textbf{Other Open-source VLMs}: Qwen-VL~\cite{bai2023qwenvl}, InternVL~\cite{chen2024internvl}, and CogVLM~\cite{wang2025cogvlm}.

\noindent \textbf{Evaluation Setting.} We follow the official inference hyperparameters and decoding settings for each model. For ConsensusDrop, we use the Convex Fuser by default, with $\tau_c = \tau_m = 1$ and $\alpha=0.7$, with EGTM enabled.

\noindent \textbf{Comparison Methods.} As pruning baselines, we consider ToMe~\cite{bolya2022tome}, FastV~\cite{fastv-prune-cross}, SparseVLM~\cite{sparsevlm-prune-cross}, LLaVA-PruMerge~\cite{shang2024llavaprumerge}, VisionZip~\cite{visionzip-prune-viz}, and VisPruner~\cite{vispruner-prune-viz}. These methods cover a variety of pruning strategies applied at different stages of VLMs, 
providing a comprehensive and fair comparison.

% \noindent \textbf{Comparison Methods.} We benchmark against representative pruning baselines covering diverse strategies across VLM stages: ToMe~\cite{bolya2022tome}, FastV~\cite{fastv-prune-cross}, SparseVLM~\cite{sparsevlm-prune-cross}, LLaVA-PruMerge~\cite{shang2024llavaprumerge}, VisionZip~\cite{visionzip-prune-viz}, and VisPruner~\cite{vispruner-prune-viz}.

\subsection{Main Results}
\label{sec:main_results}

% We apply \textbf{ConsensusDrop} to \textbf{LLaVA-1.5-7B}
% and compare against vision-only pruning and merging methods (VisionZip \cite{visionzip-prune-viz}, VisPruner \cite{vispruner-prune-viz}, ToMe \cite{bolya2022tome}, PruMerge+ \cite{shang2024llavaprumerge}) as well as cross-modal-only pruning methods (FastV \cite{fastv-prune-cross}, SparseVLM \cite{sparsevlm-prune-cross}).
Table~\ref{tab:llava-1.5-singlecol} presents the performance of \textbf{ConsensusDrop} on LLaVA-1.5-7B
% Table~\ref{tab:llava-1.5-singlecol} reports performance
across four retention budgets (192, 128, 64, and 32 visual tokens out of 576). ConsensusDrop consistently achieves the strongest accuracy--efficiency trade-off across all budgets, preserving \textbf{99.7\%}, \textbf{99.2\%}, \textbf{95.2\%}, and \textbf{90.5\%} of baseline performance, respectively. As compression becomes more aggressive, the advantage of ConsensusDrop becomes decisive. At \textbf{32 tokens} ($\downarrow$94.4\%), it achieves the best performance across \emph{all} benchmarks, substantially outperforming both vision-only and cross-modal-only methods while retaining \textbf{90.5\%} of baseline accuracy. This result highlights the importance of multimodal consensus under extreme token reduction, where selecting the \emph{right} visual tokens becomes critical. From an efficiency perspective, ConsensusDrop consistently improves over cross-modal-only methods such as FastV \cite{fastv-prune-cross} and SparseVLM \cite{sparsevlm-prune-cross} in both TTFT and KV cache footprint, while remaining competitive with vision-only methods. VisionZip \cite{visionzip-prune-viz} achieves the lowest latency across settings; however, ConsensusDrop delivers substantially higher accuracy at comparable token budgets, yielding a more favorable accuracy--latency trade-off overall. Finally, while SparseVLM \cite{sparsevlm-prune-cross} has the smallest analytical KV cache size, its LLM-internal stage-wise pruning leads to non-uniform KV cache sizes across layers, which can result in memory fragmentation not reflected in Table \ref{tab:llava-1.5-singlecol}. In contrast, ConsensusDrop performs pruning prior to LLM inference, leading to a compact and uniform KV cache.

\subsection{ConsensusDrop with Higher Resolution Inputs}
\label{subsec:consens-with-high-in-res}

Across high-resolution inputs in LLaVA-NeXT-7B~\cite{llavanext}, where the visual sequence expands to \textbf{2880} tokens, ConsensusDrop consistently matches or outperforms the strongest pruning baselines under identical token budgets (Tab.~\ref{tab:llava-next}). At \textbf{640 tokens} ($\downarrow 77.8\%$), it preserves \textbf{99.0\%} of full-model performance, slightly exceeding VisPruner and maintaining strong accuracy across GQA, POPE, and MME. With more aggressive pruning to \textbf{320} ($\downarrow 88.9\%$) and \textbf{160 tokens} ($\downarrow 94.4\%$), ConsensusDrop retains \textbf{93.9\%} and \textbf{87.1\%} relative performance, respectively, outperforming VisionZip, VisPruner, and cross-modal-only methods by clear margins. These results show that multimodal consensus remains effective even under dense tokenization, enabling aggressive compression while preserving LLM-aligned visual evidence.

\subsection{ConsensusDrop with Video Understanding}
\label{subsec:consens-with-video}

Tab.~\ref{tab:video-llava} evaluates ConsensusDrop on Video-LLaVA across three video QA benchmarks, a setting with heavy temporal redundancy from continuous frames (8 frames at 224 resolution, \textbf{2048} tokens). Under identical reduction ratios, ConsensusDrop consistently matches or outperforms prior methods. At \textbf{455 tokens} ($\downarrow 77.8\%$), it achieves the best average accuracy/score (48.6/3.32), matching the full model and slightly exceeding VisPruner and FastV. With more aggressive pruning to \textbf{227} ($\downarrow 88.9\%$) and \textbf{114 tokens} ($\downarrow 94.4\%$), ConsensusDrop maintains the strongest average performance, retaining roughly \textbf{92\%} of full-model accuracy and outperforming competing baselines.

\begin{figure}
  \centering
  \includegraphics[width=0.85\columnwidth]{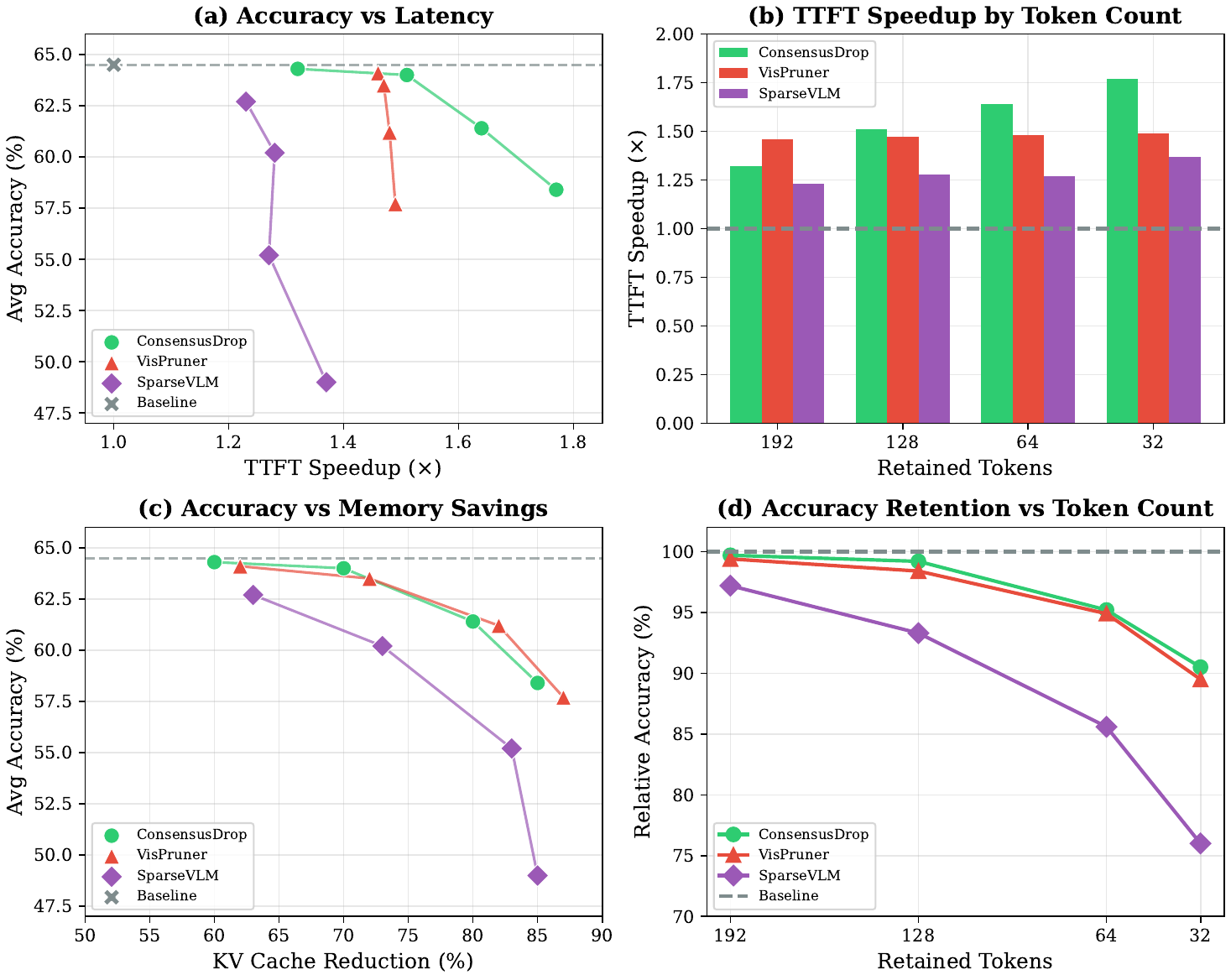}
  \caption{\textbf{Global accuracy--efficiency trade-offs on LLaVA-1.5-7B.}
  ConsensusDrop consistently lies on or above the Pareto frontier compared to
  vision-only (VisPruner) and cross-modal-only (SparseVLM) pruning across token
  budgets $\{192,128,64,32\}$.}
  \label{fig:eff_combined}
\end{figure}

\subsection{ConsensusDrop on other VLM Architectures}
\label{subsec:consensus-other-vlm-arch}
\begin{table}
  \centering
  \footnotesize
  \setlength{\tabcolsep}{2.5pt}
  \renewcommand{\arraystretch}{1.05}
  \resizebox{\columnwidth}{!}{%
    \begin{tabular}{l *{5}{c} cc}
      \toprule
      \textbf{Method} &
      \textbf{$\text{VQA}^\text{V2}$} & \textbf{GQA} &
      \textbf{$\text{VQA}^\text{Text}$} & \textbf{POPE} & \textbf{MME} &
      \textbf{Acc.} & \textbf{Rel.} \\
      \midrule
      \rowcolor{lightgray!80}
      \multicolumn{8}{c}{\textit{Upper bound: all 2880 visual tokens} ($\mathbf{100\%}$)} \\
      LLaVA-NeXT-7B & 81.2 & 62.9 & 59.6 & 86.3 & 1513.8 & 73.1 & 100.0\% \\
      \midrule
      \rowcolor{lightgray!80}
      \multicolumn{8}{c}{\textit{Retain 640 tokens} \textcolor{OliveGreen}{($\downarrow\mathbf{77.8\%}$)}} \\
      FastV~\cite{fastv-prune-cross} &
        78.9 & 60.4 & 58.4 & 83.1 & 1477.3 & 70.9 & 97.0\% \\
      SparseVLM~\cite{sparsevlm-prune-cross} &
        78.2 & 59.1 & 56.2 & 80.9 & 1456.3 & 69.4 & 94.9\% \\
      VisionZip~\cite{visionzip-prune-viz} &
        79.2 & 60.1 & 58.5 & 82.2 & 1468.4 & 70.7 & 96.7\% \\
      VisPruner~\cite{vispruner-prune-viz} &
        79.8 & 61.6 & 59.3 & 85.9 & 1480.7 & 72.1 & 98.6\% \\
      \textbf{ConsensusDrop} &
        79.9 & 62.1 & 59.7 & 86.1 & 1483.9 & 72.4 & 99.0\% \\
      \midrule
      \rowcolor{lightgray!80}
      \multicolumn{8}{c}{\textit{Retain 320 tokens} \textcolor{OliveGreen}{($\downarrow\mathbf{88.9\%}$)}} \\
      FastV~\cite{fastv-prune-cross} &
        71.9 & 55.9 & 55.7 & 71.7 & 1282.9 & 63.9 & 87.4\% \\
      SparseVLM~\cite{sparsevlm-prune-cross} &
        71.4 & 56.5 & 52.4 & 73.5 & 1342.7 & 64.2 & 87.8\% \\
      VisionZip~\cite{visionzip-prune-viz} &
        74.2 & 58.1 & 55.3 & 75.0 & 1348.8 & 66.0 & 90.3\% \\
      VisPruner~\cite{vispruner-prune-viz} &
        75.7 & 58.4 & 57.6 & 80.4 & 1370.1 & 68.1 & 93.2\% \\
      \textbf{ConsensusDrop} &
        76.1 & 59.1 & 58.0 & 81.5 & 1372.8 & 68.7 & 93.9\% \\
      \midrule
      \rowcolor{lightgray!80}
      \multicolumn{8}{c}{\textit{Retain 160 tokens} \textcolor{OliveGreen}{($\downarrow\mathbf{94.4\%}$)}} \\
      FastV~\cite{fastv-prune-cross} &
        61.8 & 49.8 & 51.9 & 51.7 & 1079.5 & 53.8 & 73.6\% \\
      SparseVLM~\cite{sparsevlm-prune-cross} &
        62.2 & 50.2 & 45.1 & 54.6 & 1167.1 & 54.1 & 74.0\% \\
      VisionZip~\cite{visionzip-prune-viz} &
        67.3 & 54.3 & 54.7 & 59.4 & 1239.7 & 59.5 & 81.4\% \\
      VisPruner~\cite{vispruner-prune-viz} &
        70.6 & 54.7 & 56.0 & 72.9 & 1226.0 & 63.1 & 86.3\% \\
      \textbf{ConsensusDrop} &
        71.0 & 55.3 & 57.2 & 73.2 & 1232.1 & 63.7 & 87.1\% \\
      \bottomrule
    \end{tabular}%
  }
  \caption{Performance comparison of pruning methods on LLaVA-NeXT-7B.
  Acc. denotes the aggregated score across benchmarks, and
  Rel. represents the percentage of performance retained at the
  corresponding reduction ratio relative to the full-token LLaVA-NeXT-7B model.}
  \label{tab:llava-next}
\end{table}

\begin{table}
  \centering
  \footnotesize
  \setlength{\tabcolsep}{2.5pt}
  \renewcommand{\arraystretch}{1.05}
  \resizebox{\columnwidth}{!}{%
    \begin{tabular}{lcccccccc}
      \toprule
      \multirow{2}{*}{\textbf{Method}} &
      \multicolumn{2}{c}{\textbf{TGIF-QA}} &
      \multicolumn{2}{c}{\textbf{MSVD-QA}} &
      \multicolumn{2}{c}{\textbf{MSRVTT-QA}} &
      \multicolumn{2}{c}{\textbf{Average}} \\
      & Acc. & Score & Acc. & Score & Acc. & Score & Acc. & Score \\
      \midrule
      \rowcolor{lightgray!80}
      \multicolumn{9}{c}{\textit{Upper bound: all 2048 visual tokens} ($\mathbf{100\%}$)} \\
      Video-LLaVA &
        19.8 & 2.53 & 70.5 & 3.93 & 57.5 & 3.50 & 49.3 & 3.32 \\
      \rowcolor{lightgray!80}
      \multicolumn{9}{c}{\textit{Retain 455 tokens} \textcolor{OliveGreen}{($\downarrow\mathbf{77.8\%}$)}} \\
      FastV~\cite{fastv-prune-cross} &
        19.2 & 2.50 & 69.1 & 3.91 & 54.4 & 3.42 & 47.6 & 3.28 \\
      VisPruner~\cite{vispruner-prune-viz} &
        18.4 & 2.49 & 70.2 & 3.95 & 56.7 & 3.50 & 48.4 & 3.31 \\
      \textbf{ConsensusDrop} &
        18.6 & 2.51 & 70.3 & 3.94 & 56.8 & 3.52 & 48.6 & 3.32 \\
      \rowcolor{lightgray!80}
      \multicolumn{9}{c}{\textit{Retain 227 tokens} \textcolor{OliveGreen}{($\downarrow\mathbf{88.9\%}$)}} \\
      FastV~\cite{fastv-prune-cross} &
        14.3 & 2.42 & 68.9 & 3.90 & 53.0 & 3.40 & 45.4 & 3.24 \\
      VisPruner~\cite{vispruner-prune-viz} &
        15.9 & 2.41 & 69.3 & 3.92 & 55.6 & 3.45 & 46.9 & 3.26 \\
      \textbf{ConsensusDrop} &
        16.1 & 2.43 & 69.5 & 3.94 & 55.8 & 3.43 & 47.1 & 3.27 \\
      \rowcolor{lightgray!80}
      \multicolumn{9}{c}{\textit{Retain 114 tokens} \textcolor{OliveGreen}{($\downarrow\mathbf{94.4\%}$)}} \\
      FastV~\cite{fastv-prune-cross} &
        10.6 & 2.29 & 64.1 & 3.78 & 52.4 & 3.39 & 42.4 & 3.15 \\
      VisPruner~\cite{vispruner-prune-viz} &
        14.1 & 2.35 & 65.4 & 3.79 & 54.1 & 3.41 & 44.5 & 3.18 \\
      \textbf{ConsensusDrop} &
        14.4 & 2.38 & 66.7 & 3.82 & 54.4 & 3.45 & 45.2 & 3.22 \\
      \bottomrule
    \end{tabular}%
  }
  \caption{Video understanding performance with Video-LLaVA across three video question answering benchmarks.
  Performance is evaluated on the first 1{,}000 samples from each benchmark, and \texttt{gpt-3.5-turbo} is used for scoring as in \cite{fastv-prune-cross}.}
  \label{tab:video-llava}
  \vspace{-3mm}
\end{table}

\begin{table}
  \centering
  \footnotesize
  \setlength{\tabcolsep}{2.5pt}
  \renewcommand{\arraystretch}{1.05}
  \resizebox{\columnwidth}{!}{%
    \begin{tabular}{l *{4}{c} cc}
      \toprule
      \textbf{Method} &
      \textbf{$\text{VQA}^\text{V2}$} & \textbf{GQA} &
      \textbf{$\text{SQA}^\text{IMG}$} & \textbf{$\text{VQA}^\text{Text}$} &
      \textbf{Acc.} & \textbf{Rel.} \\
      \midrule
      \rowcolor{lightgray!80}
      \multicolumn{7}{c}{\textit{Upper bound: all 256 visual tokens} ($\mathbf{100\%}$)} \\
      Qwen-VL-7B &
        78.8 & 59.3 & 67.1 & 63.8 & 67.3 & 100.0\% \\
      \rowcolor{lightgray!80}
      \multicolumn{7}{c}{\textit{Retain 128 tokens} \textcolor{OliveGreen}{($\downarrow\mathbf{50\%}$)}} \\
      FastV~\cite{fastv-prune-cross} &
        76.5 & 56.9 & 65.3 & 58.2 & 64.2 & 95.4\% \\
      VisPruner~\cite{vispruner-prune-viz} &
        77.4 & 57.8 & 65.9 & 59.6 & 65.2 & 96.9\% \\
      \textbf{ConsensusDrop} &
        78.1 & 58.9 & 66.4 & 61.1 & 66.1 & 98.2\% \\
      \midrule
      \rowcolor{lightgray!80}
      \multicolumn{7}{c}{\textit{Upper bound: all 576 visual tokens} ($\mathbf{100\%}$)} \\
      InternVL-Chat-13B &
        79.3 & 62.9 & 66.3 & 57.0 & 66.4 & 100.0\% \\
      \rowcolor{lightgray!80}
      \multicolumn{7}{c}{\textit{Retain 144 tokens} \textcolor{OliveGreen}{($\downarrow\mathbf{75\%}$)}} \\
      FastV~\cite{fastv-prune-cross} &
        74.1 & 58.2 & 66.6 & 55.6 & 63.6 & 95.8\% \\
      VisPruner~\cite{vispruner-prune-viz} &
        76.7 & 60.2 & 67.3 & 55.3 & 64.9 & 97.7\% \\
      \textbf{ConsensusDrop} &
        76.9 & 60.5 & 67.9 & 56.1 & 65.4 & 98.5\% \\
      \midrule
      \rowcolor{lightgray!80}
      \multicolumn{7}{c}{\textit{Upper bound: all 1225 visual tokens} ($\mathbf{100\%}$)} \\
      CogVLM-chat-1.1-17B &
        80.9 & 58.2 & 68.4 & 69.2 & 69.2 & 100.0\% \\
      \rowcolor{lightgray!80}
      \multicolumn{7}{c}{\textit{Retain 123 tokens} \textcolor{OliveGreen}{($\downarrow\mathbf{90\%}$)}} \\
      FastV~\cite{fastv-prune-cross} &
        74.2 & 40.3 & 63.5 & 41.9 & 55.0 & 79.5\% \\
      VisPruner~\cite{vispruner-prune-viz} &
        74.6 & 48.4 & 68.2 & 59.6 & 62.7 & 90.6\% \\
      \textbf{ConsensusDrop} &
        75.1 & 49.3 & 69.6 & 59.5 & 63.4 & 91.6\% \\
      \bottomrule
    \end{tabular}%
  }
  \caption{Performance of pruning methods on different VLM architectures.
  Acc. denotes the average accuracy across benchmarks, and
  Rel. represents the percentage of performance retained.}
  \label{tab:qwen-internvl-cogvlm}
  \vspace{-3mm}
\end{table}

We further evaluate ConsensusDrop on three representative VLM backbones with distinct vision--language interfaces: Qwen-VL-7B~\cite{bai2023qwenvl}, InternVL-Chat-13B~\cite{chen2024internvl}, and CogVLM-chat-1.1-17B~\cite{wang2025cogvlm} (Tab.~\ref{tab:qwen-internvl-cogvlm}). Across all token budgets, ConsensusDrop consistently outperforms the cross-modal-only pruning baseline FastV~\cite{fastv-prune-cross} and yields small but reliable gains over VisPruner in average accuracy. For Qwen-VL at 128/256 tokens ($\downarrow 50\%$), ConsensusDrop retains \textbf{98.2\%} of full performance (Acc.\ 66.1 vs.\ 67.3), improving over VisPruner (96.9\%). On InternVL at 144/576 tokens ($\downarrow 75\%$), it again achieves the best retained performance (98.5\%) and the highest Acc.\ (65.4). Under the most extreme compression on CogVLM (123/1225 tokens, $\downarrow 90\%$), ConsensusDrop preserves \textbf{91.6\%} of the original accuracy, exceeding both FastV (79.5\%) and VisPruner (90.6\%).

\begin{figure}
  \centering
  \includegraphics[width=0.85\columnwidth]{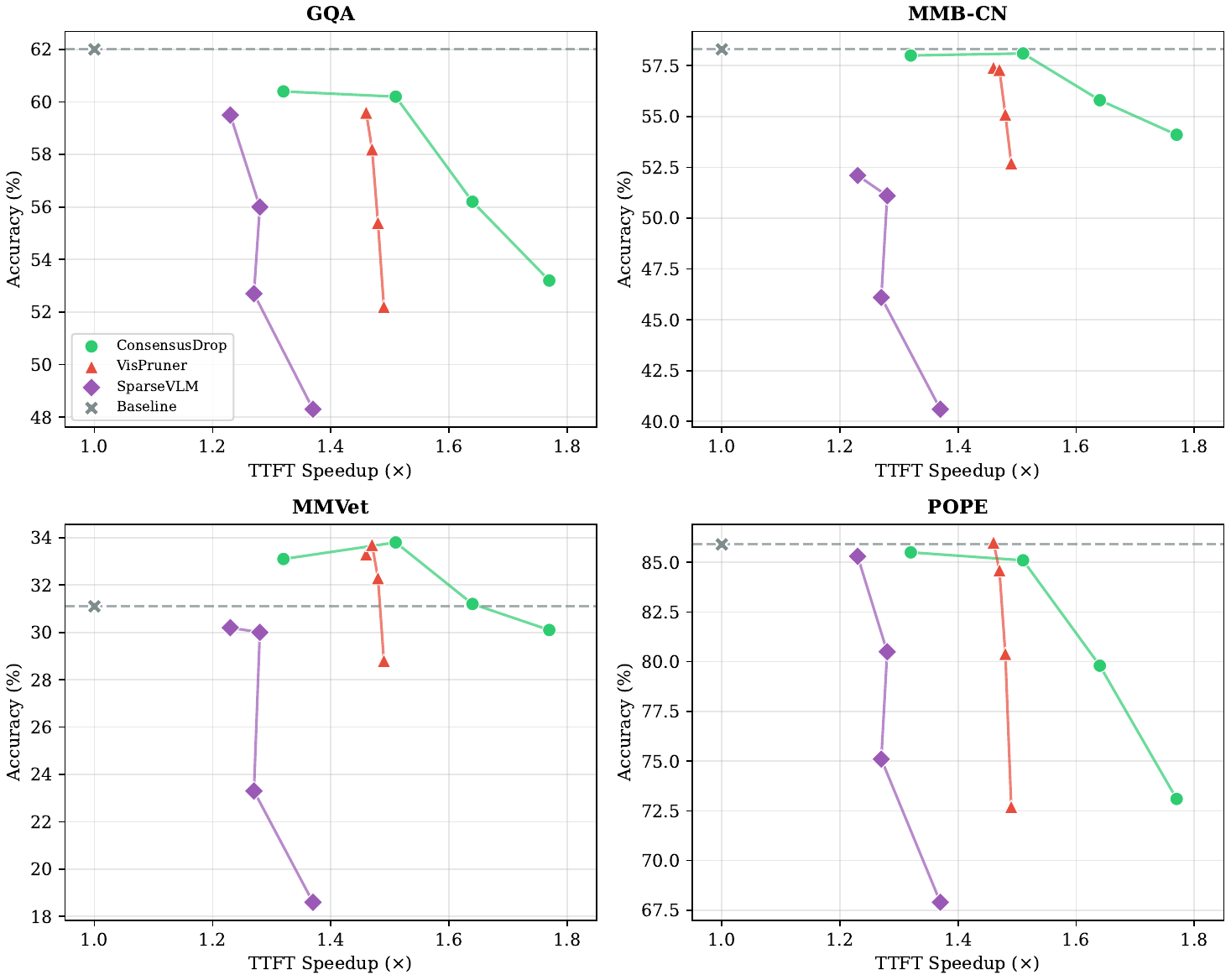}
  \caption{\textbf{Per-dataset accuracy vs.\ TTFT trade-offs.}
  Across datasets, ConsensusDrop preserves
  higher accuracy at comparable or higher TTFT speedups, indicating consistent
  efficiency gains.}
  \label{fig:eff_per_dataset}
\end{figure}

\subsection{Efficiency Analysis} 
\label{sec:efficiency}
We evaluate the accuracy--efficiency trade-off of token pruning on LLaVA-1.5-7B, focusing on \textbf{TTFT} (prefill latency), \textbf{TPOT} (decode latency), and \textbf{KV cache} memory. \textbf{Latency.} As the token budget decreases, ConsensusDrop achieves substantial TTFT speedups while maintaining strong accuracy retention. At aggressive compression (32 tokens, $\downarrow$94.4\%), it retains \textbf{90.5\%} of baseline accuracy with a \textbf{1.77$\times$} TTFT speedup, outperforming both baselines. \textbf{Memory.} By pruning tokens prior to LLM inference, ConsensusDrop yields uniform KV cache layouts and consistent memory savings, reducing KV usage by up to \textbf{85\%} at extreme compression. \textbf{Pareto Optimality.} Fig.~\ref{fig:eff_combined} shows that ConsensusDrop consistently achieves a more favorable accuracy--latency--memory trade-off than VisPruner and SparseVLM, with the advantage most pronounced at low token budgets where token selection quality is critical. \textbf{Across Datasets.} Fig.~\ref{fig:eff_per_dataset} confirms that these gains generalize across multiple benchmarks, with ConsensusDrop remaining closer to the Pareto frontier on all evaluated datasets.

%% file: sec/6_conclusion.tex
\section{Conclusion}
\label{sec:conc}

ConsensusDrop is a training-free, plug-and-play framework that exploits multimodal consensus to retain the most informative visual tokens in VLMs. By fusing both vision and cross-modal saliency, it consistently outperforms prior pruning methods under aggressive token reduction.

%% file: sec/X_suppl.tex
\newpage

\appendix

\section{Related Works}
\label{app:relatedAppendix}

\noindent \textbf{Vision Language Models.}
Large Language Models (LLMs) \cite{gpt5, claude-4.5, grok-4, kimi-k2, llama-4, gpt-oss, qwen-3-next, glm-4.5} have driven rapid progress in multimodal reasoning, motivating the development of Vision Language Models (VLMs) for image and video understanding~\cite{llava, mm-vet, mme, mme-bench}. A standard VLM \cite{llava, llavanext, improvedllava} pairs a pre-trained vision encoder such as CLIP or SigLIP~\cite{clip, siglip} with a lightweight projector~\cite{llava} that maps visual patches into the LLM embedding space. Although modern VLMs achieve strong performance across diverse tasks~\cite{qwen25_vl_2025, internvl25_2024, minicpm_v_2024}, they still struggle on fine-grained perception tasks~\cite{fine-grained-perc} and are prone to visual hallucination~\cite{lost-embedding, blind-vlm, hallu-vlm}. Efforts to improve visual grounding have included higher input resolutions~\cite{high-res, llava-uhd} and multi-encoder designs~\cite{internvl25_2024, internvl_x_2025}. Meanwhile, video understanding and OCR/document tasks require processing many frames or extremely high-resolution inputs~\cite{deepseek_ocr_2025, mplug_docowl2_2025, frag_2025, keyframevid-prune-generic}, further increasing visual token counts. This results in VLMs processing hundreds to thousands of visual tokens per input.
% —e.g., LLaVA-1.5 (576 tokens) and LLaVA-Next (2880 tokens)
% . Because visual tokens dominate prefill computation (which scales quadratically with token count) and inflate KV-cache memory and decode-time FLOPs (which scale linearly), reducing redundant visual tokens has become essential for practical and scalable VLM deployment~\cite{openvla, edge-llm-survey}.
Visual tokens dominate prefill computation, scaling quadratically with token count, while inflating KV-cache memory and decode-time FLOPs, which scale linearly. Consequently, reducing redundant visual tokens has become essential for practical and scalable VLM deployment~\cite{openvla, edge-llm-survey}.

% \noindent \textbf{Visual Saliency Based Token Pruning.}
\noindent \textbf{Vision-only Pruning.}
One line of efficient VLM work prunes visual tokens directly from the vision encoder by leveraging vision encoder-side saliency signals. These approaches typically score patches using self-attention or related visual cues and remove low-importance tokens before they reach the language model. Representative methods include \cite{clstoken-prune-viz, llavaprumerge-prune-train-viz}, ~\cite{pact-prune-viz, stitchintime-prune-viz, vflowopt-prune-viz, dynamictokeneff-prune-viz}. VisionZip~\cite{visionzip-prune-viz} and VisPruner~\cite{vispruner-prune-viz} achieve state-of-the-art performance by scoring visual patches via encoder attention and compressing discarded tokens through contextual merging or similarity-driven de-duplication. While effective and training-free, these methods remain inherently query-agnostic because importance is determined solely in the visual modality.

% \noindent \textbf{Cross-Modal Token Pruning.}
\noindent \textbf{Cross-Modal-only Pruning.}
Another line of work prunes visual tokens using \emph{text-vision cross-attention} from within the LLM. FastV~\cite{fastv-prune-cross} ranks image tokens using early-layer cross-attention to textual queries, while SparseVLM~\cite{sparsevlm-prune-cross} extends this idea with sparsified attention patterns. Subsequent works explore deeper or multi-stage cross-attention signals, including pyramid-based reduction~\cite{pyramiddrop-prune-cross}, training-free pruning via attention aggregation~\cite{fitprune-prune-cross, zipvl-prune-cross}, and multi-encoder collaborative pruning~\cite{meteor-prune-cross}. Other approaches adapt attention-based importance for dynamic skipping or progressive removal in LLM layers~\cite{feather-prune-cross, atpllava-prune-cross, twigvlm-prune-cross, aim-prune-cross, skipvision-prune-cross, topv-prune-cross}. Although effective, these methods extract cross-attention inside the LLM, disabling FlashAttention, fragmenting the KV cache, and yielding \emph{only} prompt-dependent sparsity that frequently overlooks salient visual regions.

\noindent \textbf{Training-based Token Pruning.}
A separate line of work integrates token pruning into VLM training, where pruning modules or compression adapters are optimized jointly with the model~\cite{matryoshka-prune-train, matryoshka2-prune-train, llavolta-prune-train, tokenpacker-prune-train, llamavid-prune-train, llavamini-prune-train}.  
These can maintain robustness under aggressive compression, but require additional training and are not plug-and-play.

% Existing visual token pruning methods rely on either vision-only saliency, which is efficient but query-agnostic and prone to attention-sink effects, or text–vision saliency extracted inside the LLM, which incurs system-level inefficiencies, sparse and position-biased scores, and fragmented KV caches. 
ConsensusDrop overcomes these limitations by extracting query-aware cross-modal signals \emph{before} the LLM and fusing them with vision-side saliency, enabling training-free, plug-and-play token reduction that preserves both global visual context and query-critical patch tokens while retaining inference efficiency.

\section{Hypothesis Evaluation Setting}
\label{app:hypothesis}

This appendix provides implementation details for the controlled study in Section~\ref{sec:motiv}. Our goal is to compute agreement, disagreement, and recovery behavior between vision-side and cross-modal saliency signals. 

\subsection{Experimental Setting}
\label{app:hypothesis:setup}

\noindent \textbf{Setup.}
All hypothesis experiments are run with LLaVA-1.5-7B \cite{llava, improvedllava} and CLIP-ViT-L/14 at $336\times336$, yielding $N=576$ visual tokens (a $24\times24$ grid). Experiments use a single NVIDIA RTX 6000 Ada GPU (48 GB).

\noindent \textbf{Visual Token Pruning via FastV.}
Following FastV~\cite{fastv-prune-cross}, we perform early visual token pruning within the LLM decoder. Let $K$ denote the \emph{1-indexed} layer after which tokens are pruned. Attention scores are extracted from layer $K$, and the pruned token set is forwarded to layer $K{+}1$. In all experiments, we set $K=2$, so that attention is computed at the second decoder layer and pruning takes effect before the third layer.

\noindent \textbf{Retention Ratios \& Benchmarks.}
We test $\rho\in\{0.25,0.50\}$ (retain $K_{\text{tok}}=\lfloor \rho N\rfloor\in\{144,288\}$ tokens) on POPE \cite{li2023pope}, MME \cite{mme}, MMBench \cite{liu2025mmbench}, MMBench-CN \cite{liu2025mmbench}, TextVQA \cite{singh2019textvqa}, and GQA \cite{hudson2019gqa}. We evaluate recovery rates $r \in \{0.1, 0.3, 0.5\}$. For a given retention budget of $K_{\text{tok}}$ visual tokens, we replace the bottom $\lfloor rK_{\text{tok}} \rfloor$ tokens selected by the student modality with tokens drawn from the teacher modality’s top-$K_{\text{tok}}$ set. The swapped tokens are required to be non-overlapping with the retained student tokens, and higher-ranked teacher tokens are given priority during the replacement process.

\subsection{Student--Teacher Recovery Mechanism}
\label{app:hypothesis:recovery}

We formalize the student--teacher recovery mechanism used in our hypothesis evaluation, also matching the recovery fuser in Section \ref{sec:fuser}.

\noindent \textbf{Saliency Signals.}
For each image, we compute two saliency vectors over the $N$ visual tokens: (i) vision-side scores $\mathbf{s}^{(v)}\in\mathbb{R}^N$ extracted from the vision encoder, and (ii) cross-modal scores $\mathbf{s}^{(c)}\in\mathbb{R}^N$ extracted from the LLM decoder. \emph{Vision-Side Scores.} By default, we use CLS-to-patch attention ($t_v^{\text{cls}}$) from the penultimate self-attention layer of the vision encoder. \emph{Cross-Modal Scores.} Cross-modal saliency is extracted from decoder layer $K$ (1-indexed), i.e., the last layer before token pruning is applied. By default, we use
$t_{\text{cross}}^{\text{last}}$, corresponding to the attention from the final text token to the image tokens.

\noindent \textbf{Student/Teacher Assignment.}
We assign one modality as \emph{student} and the other as \emph{teacher}:
$\mathbf{s}^{(S)}\in\{\mathbf{s}^{(c)},\mathbf{s}^{(v)}\}$ and $\mathbf{s}^{(T)}$ is the other.

\noindent \textbf{Recovery via Swap.}
Given retention ratio $\rho$ and swap rate $r$, define
\[
K_{\text{tok}}=\lfloor \rho N\rfloor,\qquad M_{\text{tok}}=\lfloor rK_{\text{tok}}\rfloor.
\]
Recovery keeps the student's top $(K_{\text{tok}}-M_{\text{tok}})$ tokens, and then swaps in $M_{\text{tok}}$ high-ranked teacher tokens that are not already selected by the student. This implements a minimal intervention: the teacher only contributes where it disagrees with the student's retained set.

\begin{algorithm}
\caption{Student--Teacher Recovery}
\label{alg:recovery}
\begin{algorithmic}[1]
\Require Student scores $\mathbf{s}^{(S)}\in\mathbb{R}^N$, teacher scores $\mathbf{s}^{(T)}\in\mathbb{R}^N$, retention ratio $\rho$, swap rate $r$
\State $K_{\text{tok}}\gets \lfloor \rho N \rfloor,\quad M_{\text{tok}}\gets \lfloor rK_{\text{tok}} \rfloor$
\State $\mathcal{I}_S \gets \textsc{ArgsortDesc}(\mathbf{s}^{(S)})[1:K_{\text{tok}}]$ \Comment{student top-$K_{\text{tok}}$}
\State $\mathcal{I}_T \gets \textsc{ArgsortDesc}(\mathbf{s}^{(T)})[1:K_{\text{tok}}]$ \Comment{teacher top-$K_{\text{tok}}$}
\State $\mathcal{I}_1 \gets \mathcal{I}_S[1:K_{\text{tok}}-M_{\text{tok}}]$ \Comment{keep student top}
\State $\mathcal{I}_2 \gets [\,]$
\For{$i=1$ to $K_{\text{tok}}$}
  \If{$|\mathcal{I}_2| = M_{\text{tok}}$} \textbf{break} \EndIf
  \If{$\mathcal{I}_T[i] \notin \mathcal{I}_1$}
    \State append $\mathcal{I}_T[i]$ to $\mathcal{I}_2$
  \EndIf
\EndFor
\State $\mathcal{I}_{\text{final}} \gets \textsc{Concat}(\mathcal{I}_1,\mathcal{I}_2)$ \Comment{$|\mathcal{I}_{\text{final}}|=K_{\text{tok}}$}
\State \Return $\mathcal{I}_{\text{final}}$
\end{algorithmic}
\end{algorithm}

\noindent \textbf{Agreement/Disagreement.}
Let $\mathcal{I}_S$ and $\mathcal{I}_T$ be the student/teacher top-$K_{\text{tok}}$ sets. We compute:
\[
\text{Agreement}=\frac{|\mathcal{I}_S\cap\mathcal{I}_T|}{K_{\text{tok}}},\qquad
\text{Disagreement}=\frac{|\mathcal{I}_S\setminus\mathcal{I}_T|}{K_{\text{tok}}}.
\]

\noindent \textbf{Correction Factor.}
Let $c$ be the smallest prefix length such that the teacher prefix contains $M_{\text{tok}}$ tokens not in $\mathcal{I}_1$, i.e.,
$|\{\mathcal{I}_T[1{:}c]\setminus \mathcal{I}_1\}|=M_{\text{tok}}$.
We define the correction factor as
\begin{equation}
\text{CF}=\frac{K_{\text{tok}}-c}{K_{\text{tok}}-M_{\text{tok}}}.
\end{equation}
Intuitively, when $c$ is small, higher ranked teacher tokens are swapped with the student token set. This implies that tokens ranked important by the teacher were missed by the student -- leading to ``stronger" or ``higher" correction through recovery.

%\section{Fuser Module}
%\label{app:fuser}

%\section{Encoder-Guided Token Merge (EGTM)}
%\label{app:egtm}

\section{ConsensusDrop and EGTM Algorithms}
\label{app:algorithms}

\noindent\textbf{Discussion.}
Algorithms~\ref{alg:consensusdrop} and~\ref{alg:egtm} summarize the full ConsensusDrop inference pipeline and its encoder-guided token merging component. ConsensusDrop first identifies a compact set of salient visual tokens via multimodal consensus, then compresses the remaining visual information through EGTM to preserve contextual coverage under a fixed token budget. Importantly, both components are modular: the consensus stage is agnostic to the specific saliency estimator (e.g., SCAP), and EGTM operates purely in encoder space while merging in the LLM projector space. In the main paper, we focus on empirical behavior and efficiency gains enabled by this design; the algorithms are provided here for completeness and reproducibility.

\begin{algorithm}
\caption{ConsensusDrop: Multimodal Consensus for Visual Token Reduction}
\label{alg:consensusdrop}
\begin{algorithmic}[1]
\Require Image $\mathcal{I}$, query $\mathcal{Q}$, Vision encoder $\mathcal{E}_v$, projector $\mathcal{P}$, LLM $\mathcal{L}$, Top-$K$ budget $K$, merge budget $M$

\State $\mathbf{s}^{(v)}, \mathbf{V}, \mathbf{K} \gets \mathcal{E}_v(\mathcal{I})$ \Comment{vision scores, tokens and keys}
\State $\tilde{\mathbf{V}} \gets \mathcal{P}(\mathbf{V})$ \Comment{project to text space}
\State $\mathbf{T} \gets \textsc{Tokenize}(\mathcal{Q})$ \Comment{LLM tokenizer}
\State $\mathbf{s}^{(c)} \gets \textsc{SCAP}(\mathbf{T}, \tilde{\mathbf{V}})$ \Comment{cross-modal scores}
\State $\mathbf{s}, \mathcal{I}_{\text{top}}  \gets \textsc{Fuser}(\mathbf{s}^{(v)}, \mathbf{s}^{(c)}, K)$ \Comment{fused scores and top-$K$ indices}
\State $\mathcal{I}_{\text{non}} \gets [1,\dots,N] \setminus \mathcal{I}_{\text{top}}$ \Comment{non-top-$K$ indices}
\State $\tilde{\mathbf{V}}_{\text{top}} \gets \tilde{\mathbf{V}}[\mathcal{I}_{\text{top}}]$ 
\State $\tilde{\mathbf{V}}_{\text{merge}} \gets \textsc{EGTM}(\tilde{\mathbf{V}}, \mathbf{V}, \mathbf{K}, \mathcal{I}_{\text{non}}, M)$ \Comment{merge non-top}
\State $\tilde{\mathbf{V}}_{\text{final}} \gets \textsc{Concat}(\tilde{\mathbf{V}}_{\text{top}}, \tilde{\mathbf{V}}_{\text{merge}})$
\State $y \gets \mathcal{L}(\tilde{\mathbf{V}}_{\text{final}}, \mathbf{T})$ \Comment{LLM inference}
\State \Return $y$ \Comment{final output}
\end{algorithmic}
\end{algorithm}

\begin{algorithm}
\caption{Encoder-Guided Token Merging (EGTM)}
\label{alg:egtm}
\begin{algorithmic}[1]
\Require Projected visual tokens $\tilde{\mathbf{V}} \in \mathbb{R}^{N \times d}$, Encoder patch features $\mathbf{V} \in \mathbb{R}^{N \times d_v}$, Encoder keys $\mathbf{K} \in \mathbb{R}^{H_v \times N \times d_k}$, Top-$K$ indices $\mathcal{I}_{\text{top}}$, merge budget $M$

\State $\mathcal{I}_{\text{non}} \gets [N] \setminus \mathcal{I}_{\text{top}}$
\State $\mathbf{V}_{\text{non}} \gets \textsc{Normalize}(\mathbf{V}[\mathcal{I}_{\text{non}}])$

\State $\mathcal{A}, \mathcal{U} \gets \textsc{FPS}(\mathbf{V}_{\text{non}}, M)$
\Comment{$|\mathcal{A}|=M$, $\mathcal{U}=\mathcal{I}_{\text{non}}\setminus\mathcal{A}$}

\State $\mathbf{G} \gets \textsc{Normalize}\!\left(
\frac{1}{H_v}\sum_{h=1}^{H_v} \mathbf{K}^{(h)} \right)$
\State $\mathbf{G}_{\text{non}} \gets \mathbf{G}[\mathcal{I}_{\text{non}}]$

\For{each $u \in \mathcal{U}$}
    \State $a^\star(u) \gets \arg\max_{a \in \mathcal{A}}
    \langle \mathbf{g}_u, \mathbf{g}_a \rangle$
\EndFor

\For{each $a \in \mathcal{A}$}
    \State $\mathcal{C}(a) \gets \{a\} \cup \{u \in \mathcal{U} : a^\star(u)=a\}$
    \State $\tilde{\mathbf{v}}^{\text{merge}}_a
    \gets \frac{1}{|\mathcal{C}(a)|}
    \sum_{j \in \mathcal{C}(a)} \tilde{\mathbf{v}}_j$
\EndFor

\State $\tilde{\mathbf{V}}_{\text{final}} \gets
\mathrm{Concat}(\tilde{\mathbf{V}}[\mathcal{I}_{\text{top}}],
\{\tilde{\mathbf{v}}^{\text{merge}}_a\}_{a \in \mathcal{A}})$

\State \Return $\tilde{\mathbf{V}}_{\text{final}} \in \mathbb{R}^{(K+M)\times d}$
\end{algorithmic}
\end{algorithm}

\section{Method and Implementation Details}
\label{app:methodAppendix}

\subsection{SCAP Aggregation Strategies}
\label{app:scap_strategies}

In the main text (Sec.~\ref{sec:scap}), we employ \textbf{all-token} aggregation as the default strategy to convert the raw text-to-vision attention block $\mathbf{A} \in \mathbb{R}^{L \times N}$ into the final cross-modal saliency vector $\mathbf{s}^{(c)} \in \mathbb{R}^N$. Here, we formally define the three variants explored in our ablation study.
Recall that $\mathbf{A}_{i,j}$ denotes the attention weight from the $i$-th text token to the $j$-th visual token, and $\mathrm{Norm}(\mathbf{z}) = \mathbf{z} / (\sum_k z_k + \epsilon)$ denotes row-wise normalization.

\vspace{5pt} 
\noindent \textbf{All-Token Aggregation (Default).} We average the normalized attention scores across all $L$ query tokens. This strategy captures the global semantic alignment between the visual features and the entire textual instruction:
\begin{equation}
    \mathbf{s}^{(c)}_{\text{all}} = \frac{1}{L} \sum_{i=1}^{L} \mathrm{Norm}(\mathbf{A}_{i,:}).
\end{equation}

\noindent \textbf{Last-Token Aggregation.} This strategy relies solely on the final text token (typically the instruction terminator or separator), hypothesizing that the causal attention mechanism aggregates the full context into the last position:
\begin{equation}
    \mathbf{s}^{(c)}_{\text{last}} = \mathrm{Norm}(\mathbf{A}_{L,:}).
\end{equation}

\noindent \textbf{Max-Token Aggregation.} This strategy selects the maximum attention score received by a visual token from \emph{any} text token. It prioritizes peak alignment signals (e.g., a specific keyword strongly attending to a patch) regardless of the other tokens:
\begin{equation}
    \mathbf{s}^{(c)}_{\text{max}} = \mathrm{Norm}\left(\max_{i=1}^{L} \mathbf{A}_{i,:}\right).
\end{equation}

\begin{figure*}
    \centering
    \includegraphics[width=\linewidth]{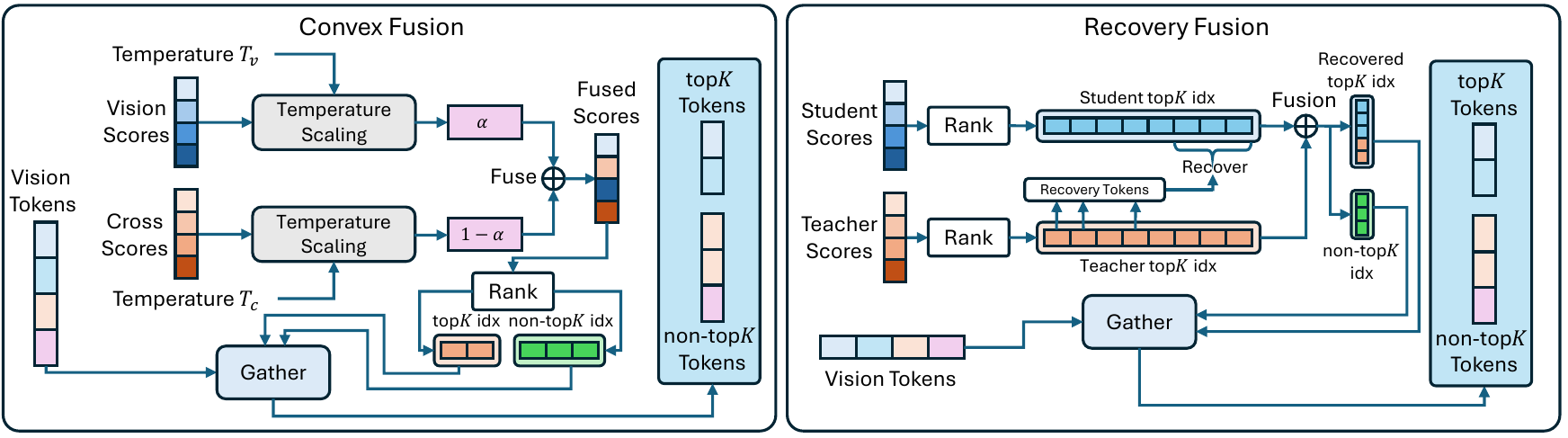}
    \caption{\textbf{Fuser module designs.}
    The Fuser module has two variants: \textbf{Convex} and \textbf{Recovery}.
    \emph{Left:} the Convex Fuser temperature-normalizes vision and cross-modal
    scores and forms a convex combination with weight $\alpha$ to obtain a
    unified importance distribution.
    \emph{Right:} the Recovery Fuser treats cross-modal scores as the student
    and vision scores as the teacher, selecting top-$K$ tokens from the student
    and recovering a small set of high-ranked teacher tokens based on a
    recovery rate.}
    \label{fig:fuser}
\end{figure*}

\subsection{Fuser Strategies}
\label{app:fuser_variants}

In Sec.~\ref{sec:fuser}, we introduced the Convex Fuser as our primary method. Here, we describe the alternative \textbf{Recovery Fuser} and visualize both designs. Both fusion strategies are illustrated in Figure~\ref{fig:fuser}.

% \noindent \textbf{Recovery Fuser Details.}
% While the Convex Fuser merges scores softly, the Recovery Fuser operates on hard selection logic. It treats the cross-modal saliency $\mathbf{s}^{(c)}$ as the \emph{student} (which captures user intent but may miss background context) and the unimodal vision saliency $\mathbf{s}^{(v)}$ as the \emph{teacher} (which captures structural completeness).
% Specifically, we first select the top-$K$ tokens based on $\mathbf{s}^{(c)}$. Then, we compute the "missed" information by comparing the selected set against the top tokens from $\mathbf{s}^{(v)}$. A recovery rate hyperparameter $\gamma$ determines how many additional tokens from the teacher's top-ranked set are explicitly added back ("recovered") to the final set.
% Although conceptually appealing, our empirical results showed that the Convex Fuser consistently outperformed the Recovery Fuser (likely due to smoother gradient properties and simpler hyperparameter tuning), leading to its selection as the default.

\noindent \textbf{Recovery Fuser.}
We additionally consider a recovery-based fuser inspired by the controlled
analysis in Sec.~\ref{sec:motiv}.
One modality is treated as a \emph{student} and the other as a \emph{teacher}.
Starting from the student’s top-$K$ tokens, we drop the lowest $rK$ tokens
and replace them with the highest-ranked, non-overlapping tokens from the
teacher, where $r \in [0,1]$ is a recovery rate.
This mechanism operates directly on ranked indices rather than fused
distributions.
Unless stated otherwise, we use the convex fuser as the default and employ
recovery fusion primarily to validate the analysis in Sec.~\ref{sec:motiv}.

\section{Experimental Setup and Additional Details}
\label{app:expsetup}

\subsection{Datasets}
\label{sec:detail_dataset}

We evaluate ConsensusDrop on a total of 13 widely used benchmarks, comprising 10 image-based benchmarks and 3 video-based benchmarks. These datasets cover a broad range of vision--language capabilities, including visual question answering, multimodal reasoning, hallucination detection, and video understanding. Unless otherwise stated, all inference protocols and evaluation metrics follow the standard settings used in LLaVA-1.5~\cite{llava, improvedllava} and Video-LLaVA~\cite{lin2023videollava}.

\subsubsection{Image Benchmarks}

We conduct experiments on 10 image benchmarks that are commonly used to evaluate large vision--language models. These include four visual question answering datasets and six multimodal reasoning and robustness benchmarks.

\noindent\textbf{VQAv2~\cite{goyal2017vqav2}.}
VQAv2 evaluates open-ended visual question answering with a strong emphasis on reducing language bias. The dataset consists of images from MSCOCO~\cite{lin2014microsoft}, where each question is paired with multiple images corresponding to different answers. We report accuracy on the test-dev split, which contains 107,394 image--question pairs, using the standard evaluation protocol.

\noindent\textbf{GQA~\cite{hudson2019gqa}.}
GQA focuses on compositional reasoning and structured visual understanding. Questions are generated from scene graphs derived from Visual Genome~\cite{krishna2017visual}, ensuring that each question corresponds to an explicit semantic reasoning path. We evaluate accuracy on the test-dev split, which contains 12,578 image--question pairs.

\noindent\textbf{ScienceQA (SQA-IMG)~\cite{lu2022sqa}.}
ScienceQA evaluates scientific reasoning using multiple-choice questions spanning natural science, social science, and language topics. We use the image-based subset (SQA-IMG), which contains 2,017 image--question pairs from the test set, following prior work \cite{vispruner-prune-viz, visionzip-prune-viz, sparsevlm-prune-cross}.

\noindent\textbf{TextVQA~\cite{singh2019textvqa}.}
TextVQA measures a model’s ability to reason over textual content embedded in images, such as signs and product labels. Images are sourced primarily from Open Images~\cite{krasin2017openimages}. We report accuracy on the validation set, which contains 5,000 image--question pairs.

\noindent\textbf{POPE~\cite{li2023pope}.}
POPE evaluates object hallucination in vision-language models using binary questions about object presence. Images are drawn from MSCOCO~\cite{lin2014microsoft}. We report the average F1 score across three evaluation splits, following the standard protocol, over 8,910 image--question pairs.

\noindent\textbf{MME~\cite{fu2024mme}.}
MME provides a comprehensive evaluation of multimodal perception and cognition across 14 subtasks, including OCR, object recognition, and fine-grained visual understanding. All questions are binary judgment tasks. We report the perception score, following prior work \cite{vispruner-prune-viz}, over 2,374 image--question pairs.

\noindent\textbf{MMBench and MMBench-CN~\cite{liu2025mmbench}.}
MMBench evaluates multimodal perception and reasoning using multiple-choice questions organized across hierarchical capability levels. We evaluate both the English (MMBench) and Chinese (MMBench-CN) versions, which contain 4,377 and 4,329 image--question pairs respectively.

\noindent\textbf{MM-Vet~\cite{yu2023mmvet}.}
MM-Vet focuses on integrated vision--language capabilities such as OCR, spatial reasoning, knowledge grounding, and mathematical reasoning. The benchmark includes 218 carefully curated image--question pairs and uses a GPT-based evaluator to score model outputs.

\noindent \textbf{SEEDBench~\cite{li2023seed}.}
SEEDBench is a large-scale multimodal benchmark designed to evaluate fine-grained visual understanding and reasoning in both images and videos. It consists of approximately 19K multiple-choice questions annotated by human assessors and spans 12 distinct evaluation aspects, covering object recognition, spatial understanding, temporal reasoning, and cross-modal comprehension. In this work, we use SEEDBench exclusively for ablation studies to analyze the behavior of different token selection and fusion components under controlled settings.

\subsubsection{Video Benchmarks}

To assess performance under higher visual redundancy, we also evaluate on three video question answering benchmarks used in Video-LLaVA~\cite{lin2023videollava}. Evaluation follows the Video-ChatGPT protocol~\cite{video-chatgpt}, using \textsc{gpt-3.5-turbo} as the scoring assistant. We follow FastV~\cite{fastv-prune-cross} and evaluate on the first 1,000 samples of each benchmark.

\noindent\textbf{TGIF-QA~\cite{tgif-qa}.}
TGIF-QA extends visual question answering to animated GIFs, requiring models to reason over both spatial and temporal information. We evaluate performance on the Frame QA task subset.

\noindent\textbf{MSVD-QA~\cite{msvd-qa}.}
MSVD-QA is built on the Microsoft Research Video Description corpus~\cite{MRVDC}, with question--answer pairs derived from video captions. The dataset contains 1,970 video clips and over 50K QA pairs.

\noindent\textbf{MSRVTT-QA~\cite{msvd-qa}.}
MSRVTT-QA is based on the MSRVTT dataset~\cite{msrvtt}, which contains more diverse and complex video scenes than MSVD \cite{MRVDC}. The benchmark includes 10K video clips and 243K question--answer pairs.

\subsection{Model Architectures}
\label{sec:detail_models}

We evaluate ConsensusDrop across representative state-of-the-art vision--language models spanning image and video understanding. All models are used in their publicly released configurations, without architectural modification.

\noindent\textbf{LLaVA-1.5~\cite{liu2024llava1.5}.}
LLaVA-1.5 is a widely adopted open-source vision--language model that combines a CLIP-based vision encoder with a Vicuna language model. Visual features extracted by the vision encoder are projected into the language embedding space via a linear projection layer, enabling the LLM to process visual tokens directly. Compared to the original LLaVA, LLaVA-1.5 increases the input image resolution from $224\times224$ to $336\times336$ and incorporates additional instruction tuning data, resulting in substantially improved performance across multimodal benchmarks.

\noindent\textbf{LLaVA-NeXT (LLaVA-1.6)~\cite{liu2024llavanext}.}
LLaVA-NeXT extends LLaVA-1.5 by supporting dynamic high-resolution image inputs. Instead of using a fixed resolution, the model adaptively selects an aspect ratio and partitions high-resolution images into multiple sub-images, each processed independently by the vision encoder. The resulting visual representations are concatenated before being passed to the language model. This design improves performance on tasks requiring fine-grained perception, such as OCR and spatial reasoning, while preserving the underlying architecture.

\noindent\textbf{Video-LLaVA~\cite{lin2023videollava}.}
Video-LLaVA extends the LLaVA architecture to video understanding by encoding individual video frames using the same vision encoder and aligning their representations before projection into the language model. Frame-level features are concatenated to form a unified visual token sequence, enabling joint reasoning over spatial and temporal information. After multimodal alignment training, Video-LLaVA supports both image and video-based question answering within a single unified framework.

\noindent\textbf{Qwen-VL~\cite{bai2023qwenvl}.}
Qwen-VL is an open-source vision--language model that pairs an OpenCLIP vision encoder with the Qwen language model. Unlike LLaVA’s linear projection, Qwen-VL employs a vision--language adapter based on cross-attention to map visual features into a fixed-length token sequence. The model is trained in multiple stages to align visual and textual representations, with Qwen-VL-Chat further fine-tuned for conversational multimodal tasks.

\subsection{Implementation Details}
\label{sec:impl_details}

\noindent\textbf{Static Cross-Attention Probe (SCAP).}
SCAP is instantiated as a lightweight replica of the first decoder layer’s attention module and is executed once during the prefill stage. It extracts query-aware text--vision saliency without maintaining key/value caches or modifying the LLM forward pass. As a result, SCAP does not interfere with optimized attention kernels (e.g., FlashAttention) used during subsequent decoding.

\noindent\textbf{Training-Free Deployment.}
ConsensusDrop introduces no additional trainable parameters and is applied entirely at inference time. SCAP parameters are deep-copied from the base model and remain frozen in all experiments, ensuring that performance gains arise solely from improved token selection rather than task-specific adaptation.

\subsection{Efficiency Analysis Setting}
\label{subsec:efficiency-anal-set}

All efficiency measurements (Section \ref{sec:efficiency}) are conducted on a single NVIDIA RTX~6000 Ada GPU (48\,GB) using CUDA~12.1, PyTorch~2.1.2, and the Hugging Face Transformers library. Unless otherwise stated, we evaluate LLaVA-1.5-7B with SDPA (scaled dot-product attention) enabled and identical generation settings across methods. 

\noindent\textbf{Time to First Token (TTFT).}
TTFT measures the prefill latency, defined as the elapsed time from input submission to the generation of the first output token. This metric captures the cost of processing the full multimodal prompt and initializing the KV cache. We report TTFT using GPU-synchronized wall-clock timing, averaged over multiple runs after warm-up.

\noindent\textbf{Time per Output Token (TPOT).}
TPOT measures the steady-state decode latency per generated token. We compute TPOT by generating a fixed-length output sequence with greedy decoding and KV caching enabled, and report the average per-token latency excluding the prefill phase. This metric reflects the decoding cost after the initial prompt has been processed.

\noindent\textbf{KV Cache Footprint.}
The KV cache memory footprint is estimated analytically based on the retained visual token count after pruning. For LLaVA-1.5-7B (LLaMA-7B backbone), the cache size is computed as $\text{KV}~(\text{MB}) = \frac{2 \times L \times H \times S \times D \times B}{2^{20}}$,
where $L=32$ is the number of transformer layers, $H=32$ the number of key-value heads, $D=128$ the head dimension, $S$ the sequence length (visual and text tokens), and $B=2$ bytes for FP16 precision. We report the KV cache size immediately after prefill to isolate the impact of visual token reduction. All reported KV cache sizes correspond to the \emph{net} memory footprint aggregated across all transformer layers. For methods that prune tokens within the LLM, with layer-wise pruning ratios that vary across stages
(\cite{sparsevlm-prune-cross, pyramiddrop-prune-cross}), this estimate does not capture potential KV
cache memory fragmentation, which can increase effective memory usage in practice.

\noindent\textbf{Evaluation Consistency.}
All methods (FastV \cite{fastv-prune-cross}, SparseVLM \cite{sparsevlm-prune-cross}, VisionZip \cite{visionzip-prune-viz}, VisPruner \cite{vispruner-prune-viz}, and ConsensusDrop) are evaluated under identical conditions, including input images, prompts, retention budgets, decoding strategy, and hardware, using their respective official codebase. Baseline results correspond to the full visual token sequence (576 tokens) for LLaVA-1.5-7B.

\section{Ablation Studies}
\label{app:ablations}

We conduct extensive ablation studies on LLaVA-1.5-7B \cite{llava, improvedllava} to analyze the contribution of each component in ConsensusDrop. All experiments retain 128 visual tokens (from 576) and evaluate on five benchmarks: MME \cite{mme}, POPE \cite{li2023pope}, ScienceQA \cite{lu2022sqa}, SEED-Bench \cite{li2023seed}, and TextVQA \cite{singh2019textvqa}. Unless otherwise noted, we use the default configuration: Convex Fuser with $\alpha = 0.7$, $\tau_v = \tau_c =1.0$, and EGTM with $K=108$ selected tokens and $M=20$ anchor tokens. The default vision-side saliency is the CLS-to-patch attention scores (extracted from the penultimate layer of the vision encoder); the default cross-modal saliency is the aggregate query-to-patch scores extracted from SCAP (Section \ref{sec:scap}).

\subsection{Fusion Weight ($\alpha$)}
\label{app:ablation:alpha}

The convex fusion weight $\alpha$ controls the balance between vision-side saliency ($\mathbf{s}^{(v)}$) and cross-modal saliency ($\mathbf{s}^{(c)}$) in the fused score: $\mathbf{s} = \alpha \cdot \mathbf{s}^{(v)} + (1-\alpha) \cdot \mathbf{s}^{(c)}$. We evaluate $\alpha \in \{0.3, 0.5, 0.7\}$, where higher values weight the vision encoder's saliency more heavily.

\begin{table}
  \centering
  \footnotesize
  \setlength{\tabcolsep}{4pt}
    \resizebox{\columnwidth}{!}{%
  \begin{tabular}{c ccccc}
    \toprule
    $\alpha$ & \textbf{MME} & \textbf{POPE} & \textbf{ScienceQA} & \textbf{SEED} & \textbf{TextVQA} \\
    \midrule
    0.3 & 1382.3 & 82.0 & 68.8 & 60.9 & 55.4 \\
    0.5 & 1414.5 & 83.0 & 69.3 & 62.0 & 55.4 \\
    \rowcolor{lightgray!50}
    0.7 & \textbf{1451.1} & \textbf{85.1} & \textbf{69.6} & \textbf{62.1} & 55.1 \\
    \bottomrule
  \end{tabular}}
  \caption{Effect of fusion weight $\alpha$. Higher vision weight ($\alpha = 0.7$) yields consistently stronger performance, supporting the hypothesis that vision-encoder saliency provides a reliable base signal that cross-modal attention refines.}
  \label{tab:ablation_alpha}
\end{table}

\begin{figure}
  \centering
  \begin{subfigure}[t]{0.48\linewidth}
    \includegraphics[width=\linewidth]{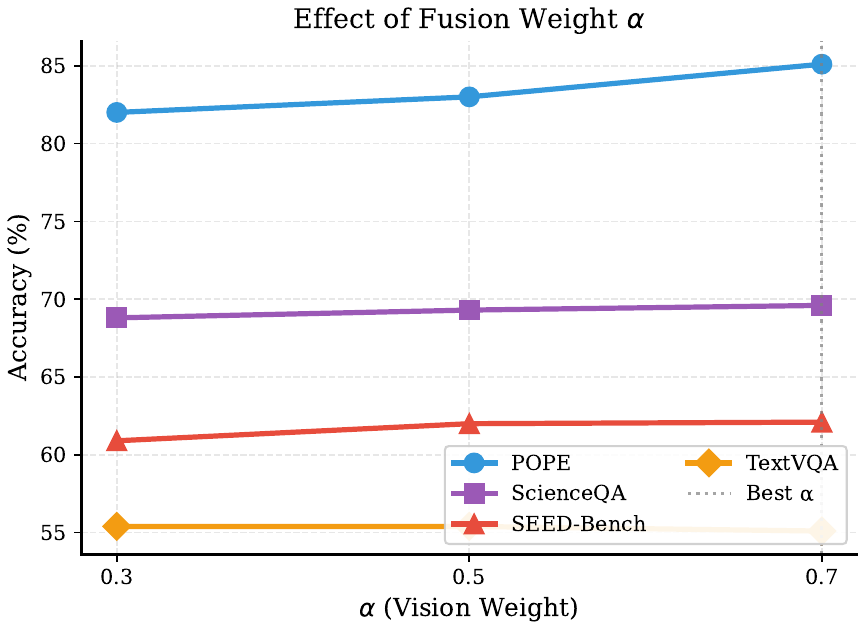}
    \caption{Accuracy vs.\ fusion weight $\alpha$.}
  \end{subfigure}
  \hfill
  \begin{subfigure}[t]{0.48\linewidth}
    \includegraphics[width=\linewidth]{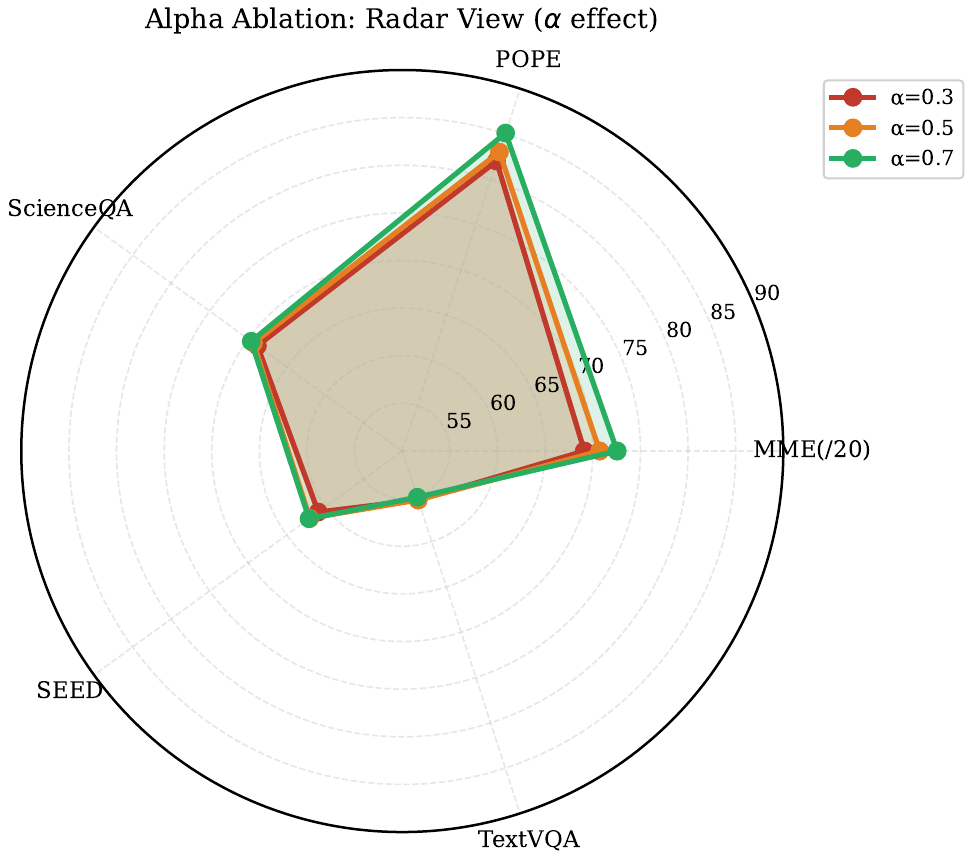}
    \caption{Radar plot across all benchmarks.}
  \end{subfigure}
  \caption{Effect of fusion weight $\alpha$ across benchmarks. Higher vision weight consistently improves performance, with $\alpha=0.7$ achieving the best trade-off.}
  \label{fig:ablation_alpha}
\end{figure}

As shown in Tab.~\ref{tab:ablation_alpha} and Fig.~\ref{fig:ablation_alpha}, performance improves roughly monotonically as $\alpha$ increases from 0.3 to 0.7. The largest gains appear on POPE (+3.1 F1) and MME (+68.8), both of which require accurate object grounding \cite{li2023pope, mme}. This suggests that vision-encoder saliency captures spatially coherent regions critical for hallucination-sensitive tasks, while cross-modal attention contributes complementary query-specific refinement. The monotonic improvement validates our design intuition: the vision encoder's self-attention already encodes semantic groupings (objects, textures, boundaries), making it a reliable ``base signal'' that cross-modal attention should refine rather than override.

\subsection{Temperature Scaling ($\tau_v$, $\tau_c$)}
\label{app:ablation:temperature}

Temperature scaling sharpens or smooths the saliency distributions before fusion. We test $\tau_v, \tau_c \in \{0.7, 1.3\}$, where lower temperatures produce sharper (more peaked) distributions and higher temperatures yield softer distributions. This ablation probes whether ConsensusDrop benefits from amplifying high-confidence tokens (low $\tau$) or preserving distributional entropy (high $\tau$).

\begin{table}
  \centering
  \footnotesize
  \setlength{\tabcolsep}{4pt}
    \resizebox{\columnwidth}{!}{%

  \begin{tabular}{cc ccccc}
    \toprule
    $\tau_v$ & $\tau_c$ & \textbf{MME} & \textbf{POPE} & \textbf{ScienceQA} & \textbf{SEED} & \textbf{TextVQA} \\
    \midrule
    0.7 & 0.7 & 1394.2 & 82.0 & 69.3 & 60.6 & 55.1 \\
    0.7 & 1.3 & 1418.3 & 82.3 & 68.7 & 61.3 & 55.3 \\
    1.3 & 0.7 & 1409.4 & 82.6 & 69.4 & 61.2 & 55.4 \\
    1.3 & 1.3 & \textbf{1416.9} & \textbf{83.8} & 69.1 & \textbf{61.9} & 55.3 \\
    \bottomrule
  \end{tabular}}
  \caption{Effect of temperature scaling. Performance varies by $<$2\% across all configurations, indicating robustness to this hyperparameter.}
  \label{tab:ablation_temperature}
\end{table}

\begin{figure}
  \centering
  \includegraphics[width=0.95\linewidth]{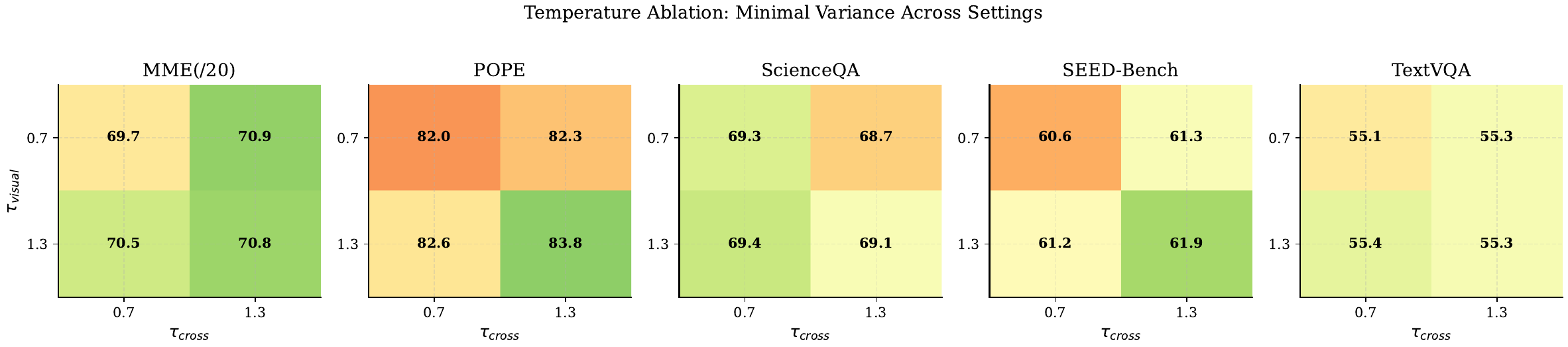}
  \caption{Temperature ablation heatmaps across all benchmarks. Minimal variance ($<$2\%) indicates robustness to temperature scaling, with MME values shown as $\times 1/20$ for visual comparability.}
  \label{fig:ablation_temperature}
\end{figure}

Tab.~\ref{tab:ablation_temperature} and Fig.~\ref{fig:ablation_temperature} demonstrate that ConsensusDrop is largely insensitive to temperature scaling: the maximum variation is 1.8\% on POPE and 2.1\% on SEED-Bench. This robustness is desirable for practical deployment, as it eliminates the need for dataset-specific tuning. The slight preference for higher temperatures ($\tau_v = \tau_c = 1.3$) on POPE and SEED-Bench suggests that preserving distributional entropy marginally helps hallucination-sensitive tasks, possibly by avoiding over-concentration on a few dominant tokens. We adopt $\tau_v = \tau_c = 1.0$ as the default for simplicity, as it lies within the robust operating region.

\subsection{Recovery Fuser: Student Modality and Recovery Rate}
\label{app:ablation:recovery}

The recovery fuser implements a student--teacher correction mechanism (Sec.~\ref{sec:fuser}), where the student modality's bottom-ranked tokens are replaced by high-ranked tokens from the teacher. We ablate (i) which modality serves as student (visual or cross-modal), and (ii) the recovery rate $\gamma \in \{0.1, 0.3, 0.5\}$, which controls what fraction of the student's lowest-ranked tokens are swapped with the teacher's highest-ranked tokens.

\begin{table}
  \centering
  \footnotesize
  \setlength{\tabcolsep}{4pt}
    \resizebox{\columnwidth}{!}{%

  \begin{tabular}{cc ccccc}
    \toprule
    \textbf{Student} & $\gamma$ & \textbf{MME} & \textbf{POPE} & \textbf{ScienceQA} & \textbf{SEED} & \textbf{TextVQA} \\
    \midrule
    \multirow{3}{*}{Visual} 
      & 0.1 & 1402.9 & \textbf{83.8} & 68.7 & \textbf{61.3} & \textbf{55.3} \\
      & 0.3 & 1404.8 & 82.9 & 69.1 & 60.7 & 55.2 \\
      & 0.5 & 1390.7 & 81.7 & \textbf{69.2} & 59.9 & 55.2 \\
    \midrule
    \multirow{3}{*}{Cross} 
      & 0.1 & 1320.4 & 80.4 & 69.0 & 58.9 & 54.1 \\
      & 0.3 & 1398.2 & 81.3 & 68.9 & 59.3 & 54.9 \\
      & 0.5 & \textbf{1412.0} & 82.2 & 68.6 & 60.1 & 55.1 \\
    \bottomrule
  \end{tabular}}
  \caption{Recovery fuser ablation. Using vision as the student with minimal recovery ($\gamma=0.1$) yields the best hallucination-sensitive metrics (POPE, SEED-Bench), while cross-as-student requires higher recovery rates to compensate for its weaker base signal.}
  \label{tab:ablation_recovery}
\end{table}

\begin{figure}
  \centering
  \includegraphics[width=0.6\linewidth]{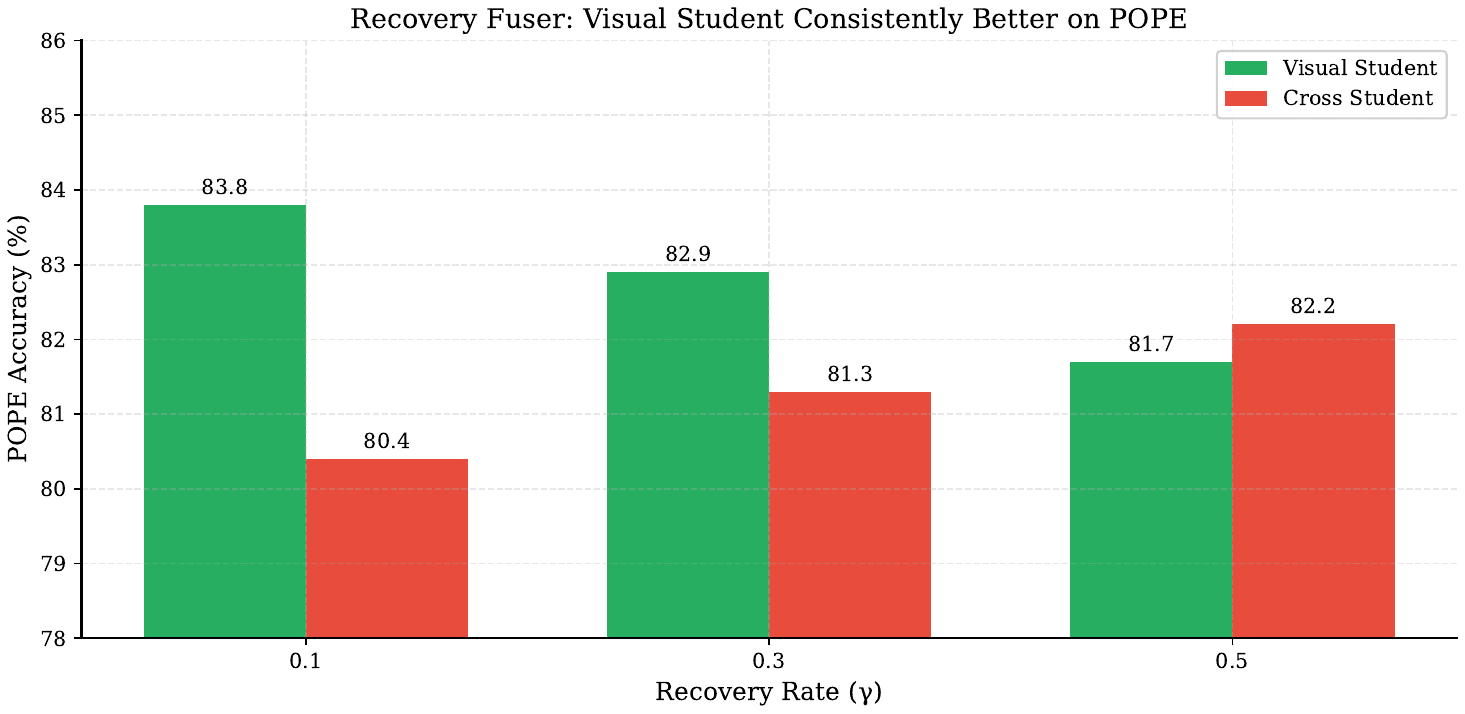}
  \caption{Recovery fuser: POPE F1 vs.\ recovery rate $\gamma$. Visual-as-student (green) consistently outperforms cross-as-student (red), with the gap widening at lower recovery rates.}
  \label{fig:ablation_recovery}
\end{figure}

Tab.~\ref{tab:ablation_recovery} and Fig.~\ref{fig:ablation_recovery} reveal a clear asymmetry: \emph{visual-as-student} consistently outperforms \emph{cross-as-student} across all recovery rates, with the gap most pronounced at low recovery rates (POPE: 83.8 vs.\ 80.4 at $\gamma=0.1$). This aligns with our hypothesis that the vision encoder provides a stable base signal, and cross-modal attention should act as a corrective teacher rather than the primary selector. 

Interestingly, higher recovery rates ($\gamma=0.5$) degrade POPE and SEED-Bench when visual is the student, suggesting that excessive correction disrupts the encoder's coherent spatial signal. Conversely, cross-as-student \emph{improves} with higher recovery rates, indicating that cross-modal saliency alone is insufficient and requires substantial correction from the vision teacher. This asymmetry provides mechanistic evidence for our design choice of vision-weighted fusion.

\subsection{Anchor Tokens (EGTM)}
\label{app:ablation:anchor}

EGTM partitions the token budget into $K$ directly selected tokens and $M$ merged anchor tokens, such that $K + M = 128$. We sweep $M \in \{0, 10, 20, 30, 40, 50\}$ (correspondingly, $K \in \{128, 118, 108, 98, 88, 78\}$) to study the trade-off between selection fidelity and contextual preservation through merging.

\begin{table}
  \centering
  \footnotesize
  \setlength{\tabcolsep}{4pt}
    \resizebox{\columnwidth}{!}{%

  \begin{tabular}{cc ccccc}
    \toprule
    $K$ & $M$ & \textbf{MME} & \textbf{POPE} & \textbf{ScienceQA} & \textbf{SEED} & \textbf{TextVQA} \\
    \midrule
    128 & 0  & 1410.2 & 81.3 & 69.4 & 61.5 & 55.0 \\
    118 & 10 & 1415.8 & 82.8 & 69.6 & 61.8 & 55.1 \\
    \rowcolor{lightgray!50}
    108 & 20 & \textbf{1441.2} & \textbf{84.9} & 69.5 & \textbf{63.2} & \textbf{57.0} \\
    98  & 30 & 1430.0 & 83.6 & 69.1 & 62.0 & 55.0 \\
    88  & 40 & 1420.6 & 84.3 & 69.0 & 62.2 & 55.4 \\
    78  & 50 & 1418.7 & 84.7 & \textbf{69.3} & 62.1 & 55.5 \\
    \bottomrule
  \end{tabular}}
  \caption{EGTM anchor token ablation. $M=20$ achieves the optimal balance, improving POPE by +3.6 F1 and TextVQA by +2.0\% over pure selection ($M=0$).}
  \label{tab:ablation_anchor}
\end{table}

\begin{figure}
  \centering
  \begin{subfigure}[t]{0.48\linewidth}
    \includegraphics[width=\linewidth]{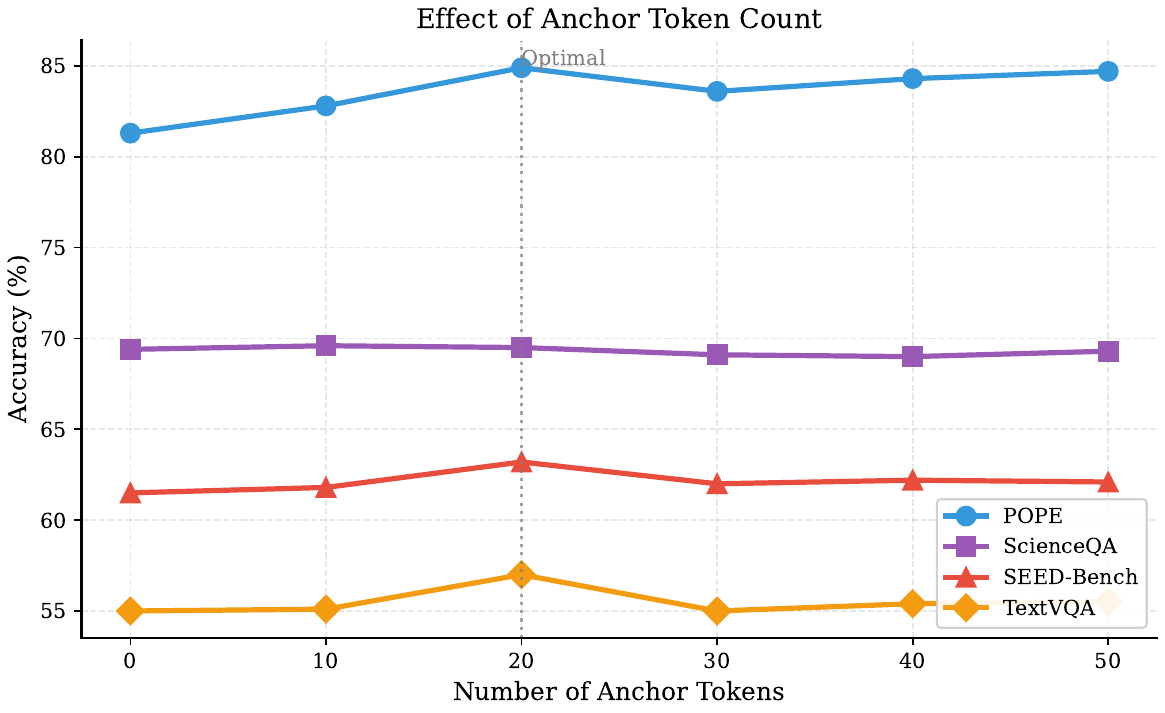}
    \caption{Accuracy vs.\ anchor count $M$.}
  \end{subfigure}
  \hfill
  \begin{subfigure}[t]{0.48\linewidth}
    \includegraphics[width=\linewidth]{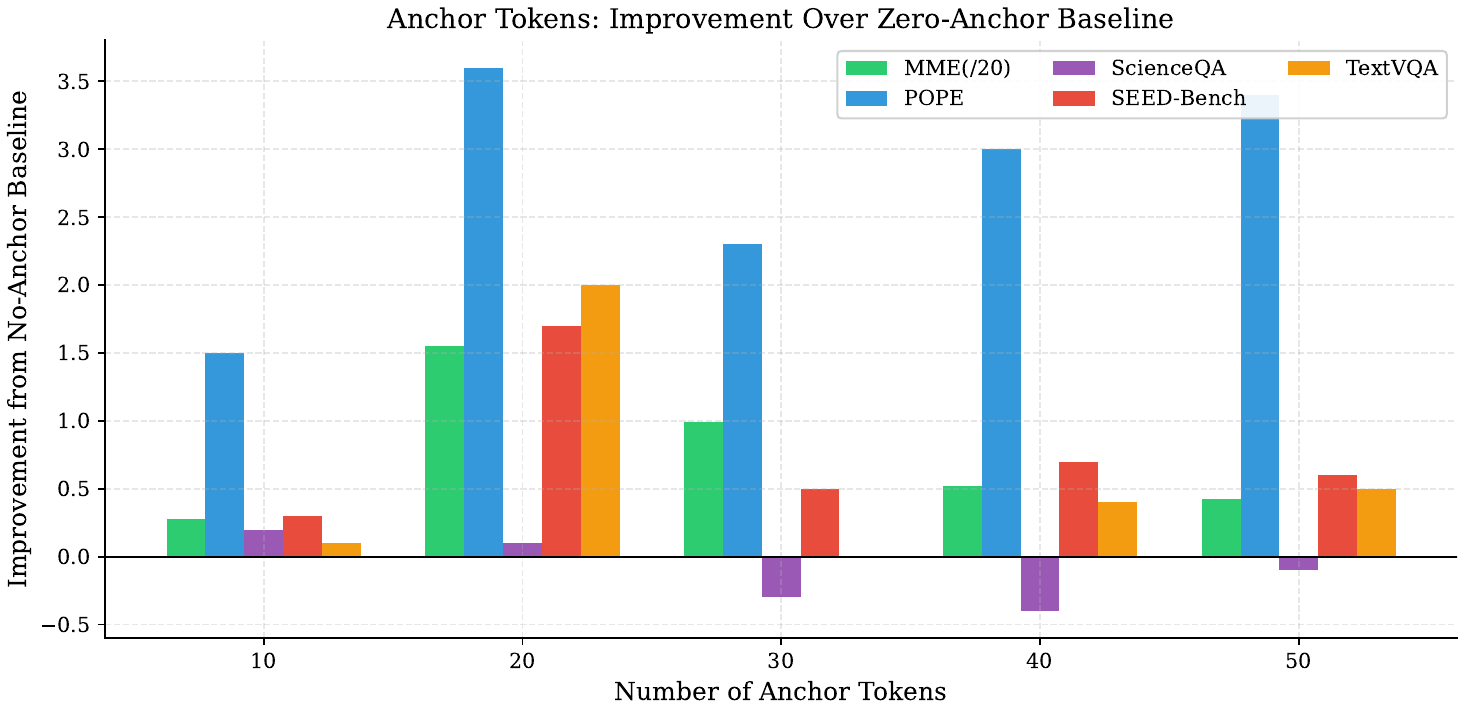}
    \caption{Improvement over $M=0$ baseline.}
  \end{subfigure}
  \caption{EGTM anchor token trade-off. (a) All benchmarks peak near $M=20$. (b) Merging provides up to +3.6\% improvement over pure selection, validating the value of contextual preservation.}
  \label{fig:ablation_anchor}
\end{figure}

Tab.~\ref{tab:ablation_anchor} and Fig.~\ref{fig:ablation_anchor} demonstrate that EGTM provides consistent gains over pure top-$K$ selection. The sweet spot occurs at $M=20$: compared to $M=0$ (no merging), this configuration improves POPE by +3.6 F1, SEED-Bench by +1.7\%, and TextVQA by +2.0\%. Performance degrades slightly for $M > 20$, as excessive merging dilutes the information preserved in anchor tokens---the merged representation becomes an over-smoothed average rather than a focused summary.

This validates our EGTM design: a modest merge budget recovers contextual information from discarded tokens (background, peripheral objects) without sacrificing the precision of top-$K$ selection for salient foreground regions. 

\subsection{Score Type}
\label{app:ablation:score}

We ablate the choice of saliency extraction for both modalities: (i) \emph{vision score type} $\in$ \{tokens, class\}, where ``tokens'' computes mean patch-to-patch attention and ``class'' uses CLS-to-patch attention from the vision encoder; and (ii) \emph{cross-modal score type} $\in$ \{all, last, max\}, where ``all'' averages attention from all text tokens to each visual token, ``last'' uses only the final text token's attention, and ``max'' takes the element-wise maximum across text tokens.

\begin{table}
  \centering
  \footnotesize
  \setlength{\tabcolsep}{3.5pt}
    \resizebox{\columnwidth}{!}{%
  \begin{tabular}{cc ccccc c}
    \toprule
    \textbf{Vision} & \textbf{Cross} & \textbf{MME} & \textbf{POPE} & \textbf{ScienceQA} & \textbf{SEED} & \textbf{TextVQA} & \textbf{Avg.} \\
    \midrule
    tokens & all  & 1431.1 & 83.9 & 68.8 & 61.5 & 55.5 & 67.4 \\
    \rowcolor{lightgray!50}
    class  & all  & \textbf{1467.3} & 83.7 & 69.1 & \textbf{63.4} & \textbf{56.2} & \textbf{68.1} \\
    tokens & last & 1443.5 & 82.3 & 68.0 & 61.8 & 55.8 & 67.0 \\
    class  & last & 1420.5 & \textbf{83.7} & \textbf{69.2} & 61.9 & 55.2 & 67.5 \\
    tokens & max  & 1436.5 & 83.6 & 69.7 & 61.0 & 55.6 & 67.5 \\
    class  & max  & 1424.1 & 83.5 & 69.1 & 62.0 & 55.2 & 67.5 \\
    \bottomrule
  \end{tabular}}
  \caption{Score type ablation. The ``class+all'' configuration achieves the highest average accuracy (+0.7\% over second-best) with the strongest MME, SEED-Bench, and TextVQA scores.}
  \label{tab:ablation_score}
\end{table}

\begin{figure}
  \centering
  \includegraphics[width=0.85\linewidth]{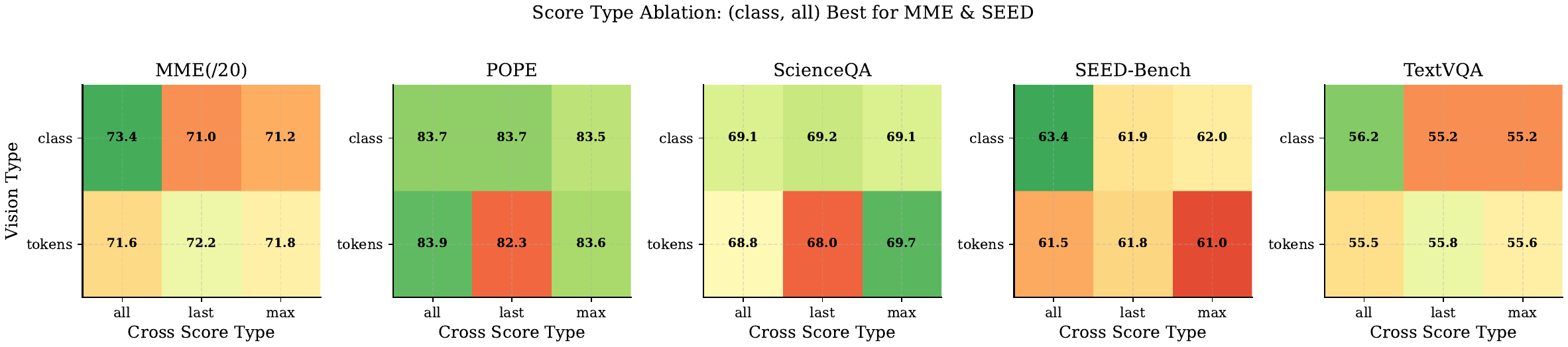}
  \caption{Score type ablation heatmap. The ``class+all'' configuration (highlighted) achieves the strongest performance across most benchmarks, with MME values shown as $\times 1/20$.}
  \label{fig:ablation_score}
\end{figure}

Tab.~\ref{tab:ablation_score} and Fig.~\ref{fig:ablation_score} show that the ``class+all'' configuration achieves the best overall performance. CLS-to-patch attention provides a globally-informed saliency signal that outperforms local patch-to-patch attention, particularly on MME (+36.2 over tokens+all) and SEED-Bench (+1.9\%). This is expected: the CLS token aggregates global context during vision encoding, making its attention pattern a better proxy for ``image-level importance'' than local patch interactions \cite{vit}.

Averaging cross-modal attention across all text tokens (``all'') outperforms ``last'' and ``max'', likely because it captures the aggregate query intent. Using only the last token (``last'') may overweight positional recency, while ``max'' can be sensitive to spurious attention spikes on individual text tokens.

\subsection{Summary}
\label{app:ablation:summary}

\begin{figure}
  \centering
  \includegraphics[width=0.8\columnwidth]{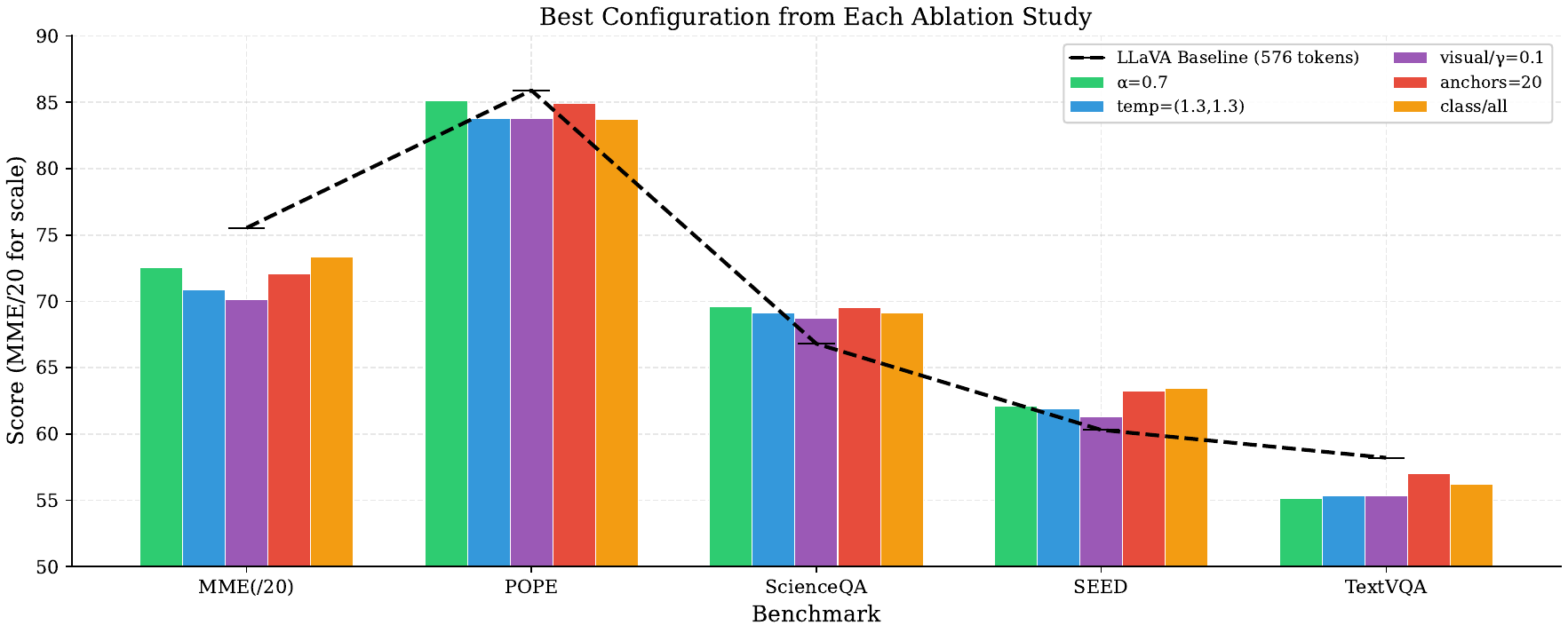}
  \caption{Grouped bar comparison with LLaVA baseline (dashed line). All best configurations approach the LLaVA baseline (576 tokens) while using only 128 tokens (78\% reduction). Score type (class+all) and anchor tokens ($M=20$) achieve the highest absolute scores.}
  \label{fig:ablation_summary_bar}
  \vspace{-2mm}
\end{figure}

Fig.~\ref{fig:ablation_summary_radar} and Fig.~\ref{fig:ablation_summary_bar} summarize the best configuration from each ablation study. Across all experiments, the default ConsensusDrop configuration ($\alpha=0.7$, $\tau_v=\tau_c=1.0$, visual student with $\gamma=0.1$, $K=108$, $M=20$, class+all scoring) consistently achieves strong performance, retaining over 95\% of baseline accuracy on hallucination-sensitive benchmarks while reducing visual tokens by 78\%. Key findings include:

\begin{figure}
  \centering
  \includegraphics[width=0.8\columnwidth]{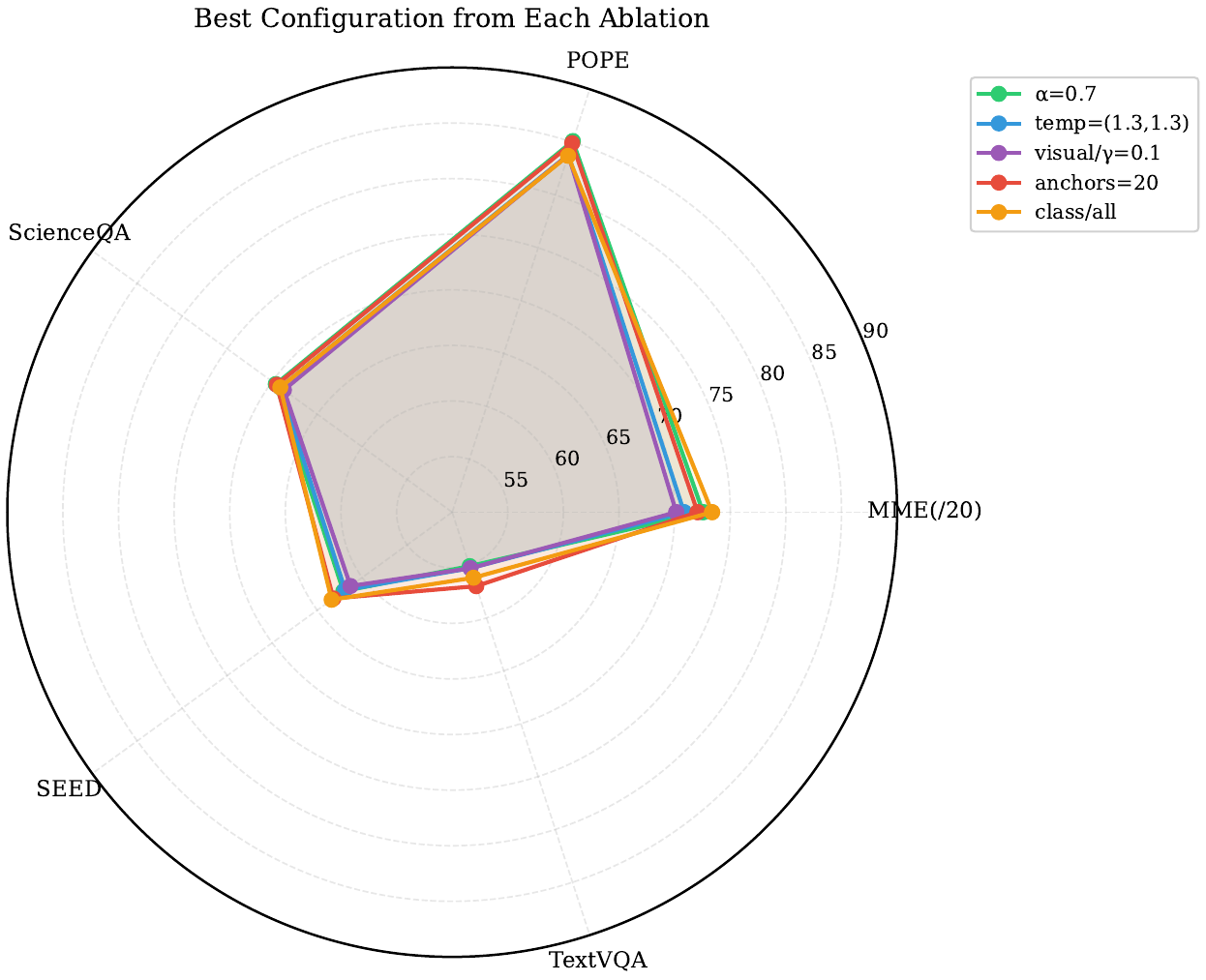}
  \caption{Radar view showing tight clustering across all optimal configurations.}
  \label{fig:ablation_summary_radar}
  \vspace{-2mm}
\end{figure}

\begin{itemize}
  \item \textbf{Vision-weighted fusion} ($\alpha=0.7$) outperforms balanced or cross-weighted fusion by +68.8 MME and +3.1 POPE F1, indicating that encoder saliency provides a reliable base signal that cross-modal attention should refine.
  \item \textbf{Temperature scaling} has minimal impact ($<$2\% variation across all benchmarks), demonstrating robustness and eliminating the need for dataset-specific tuning.
  \item \textbf{Visual-as-student} in the recovery fuser outperforms cross-as-student by up to +3.4 POPE F1 at low recovery rates, supporting the asymmetric role of vision (base) and cross-modal (refinement).
  \item \textbf{EGTM with $M=20$ anchors} provides +3.6 POPE F1 and +2.0 TextVQA over pure top-$K$ selection, validating the value of preserving contextual information through token merging.
  \item \textbf{CLS-to-patch attention} (``class'') combined with full-text averaging (``all'') yields the best score extraction strategy, leveraging global context from both modalities.
\end{itemize}

\begin{table*}[t]
  \centering
  \caption{Efficiency comparison on LLaVA-1.5-13B (576 visual tokens baseline). Larger models exhibit proportionally greater efficiency gains from token pruning. $\dagger$SparseVLM uses progressive layer-wise pruning; see Table~\ref{tab:efficiency-7b} caption.}
  \label{tab:efficiency-13b}
  \resizebox{\linewidth}{!}{%
    \begin{tabular}{l c c c c c c c c}
    \toprule
    \textbf{Method} & \textbf{Reduction} & \textbf{\# Tokens} & \textbf{TTFT (ms)} & \textbf{Speedup} & \textbf{FLOPs (T)} & \textbf{KV Cache (MB)} & \textbf{GPU Mem (GB)} & \textbf{Gen. Time (ms)} \\
    \midrule
    Baseline & 0\% & 576 & 156.5 & 1.00$\times$ & 16.56 & 489 & 26.70 & 824.6 \\
    \midrule
    \multicolumn{9}{c}{\textit{25\% Token Reduction}} \\
    \midrule
    \textbf{ConsensusDrop (Ours)} & 25\% & 432 & 138.8 & 1.13$\times$ & 12.84 & 377 & 26.66 & 800.4 \\
    VisionZip & 25\% & 432 & 135.1 & 1.16$\times$ & 12.84 & 377 & 26.47 & 792.5 \\
    VisPruner & 25\% & 432 & 112.9 & 1.39$\times$ & 13.24 & 377 & 25.87 & 762.7 \\
    SparseVLM$^\dagger$ & 25\% & 432 & 92.0 & 1.70$\times$ & 10.61 & 334 & 32.40 & 960.7 \\
    \midrule
    \multicolumn{9}{c}{\textit{50\% Token Reduction}} \\
    \midrule
    \textbf{ConsensusDrop (Ours)} & 50\% & 288 & 106.7 & 1.47$\times$ & 9.13 & 264 & 26.42 & 758.9 \\
    VisionZip & 50\% & 288 & 103.5 & 1.51$\times$ & 9.13 & 264 & 26.22 & 752.6 \\
    VisPruner & 50\% & 288 & 84.7 & 1.85$\times$ & 9.52 & 264 & 25.64 & 700.9 \\
    SparseVLM$^\dagger$ & 50\% & 288 & 93.1 & 1.68$\times$ & 6.75 & 234 & 32.39 & 962.7 \\
    \midrule
    \multicolumn{9}{c}{\textit{75\% Token Reduction}} \\
    \midrule
    \textbf{ConsensusDrop (Ours)} & 75\% & 144 & 82.5 & 1.90$\times$ & 5.45 & 152 & 26.40 & 729.1 \\
    VisionZip & 75\% & 144 & 80.6 & 1.94$\times$ & 5.45 & 152 & 26.20 & 724.4 \\
    VisPruner & 75\% & 144 & 70.5 & 2.22$\times$ & 5.83 & 152 & 25.41 & 639.0 \\
    SparseVLM$^\dagger$ & 75\% & 144 & 83.9 & 1.87$\times$ & 4.19 & 155 & 32.30 & 952.8 \\
    \midrule
    \multicolumn{9}{c}{\textit{90\% Token Reduction}} \\
    \midrule
    \textbf{ConsensusDrop (Ours)} & 90\% & 57 & 66.5 & 2.35$\times$ & 3.23 & 84 & 26.40 & 710.6 \\
    VisionZip & 90\% & 57 & 62.4 & 2.51$\times$ & 3.23 & 84 & 26.20 & 706.6 \\
    VisPruner & 90\% & 57 & 64.4 & 2.43$\times$ & 3.60 & 84 & 25.37 & 601.7 \\
    SparseVLM$^\dagger$ & 90\% & 57 & 77.3 & 2.02$\times$ & 2.99 & 109 & 32.28 & 948.3 \\
    \midrule
    \multicolumn{9}{c}{\textit{95\% Token Reduction}} \\
    \midrule
    \textbf{ConsensusDrop (Ours)} & 95\% & 28 & 64.3 & 2.43$\times$ & 2.49 & 61 & 26.40 & 706.1 \\
    VisionZip & 95\% & 28 & 60.2 & 2.60$\times$ & 2.49 & 61 & 26.20 & 702.2 \\
    VisPruner & 95\% & 28 & 64.5 & 2.43$\times$ & 2.86 & 61 & 25.37 & 589.2 \\
    SparseVLM$^\dagger$ & 95\% & 28 & 75.7 & 2.07$\times$ & 2.61 & 93 & 32.31 & 944.4 \\
    \bottomrule
    \end{tabular}%
  }
\end{table*}

\section{Additional Efficiency Analysis}
\label{app:efficiency}

This section provides comprehensive efficiency benchmarks comparing ConsensusDrop against VisionZip~\cite{visionzip-prune-viz}, VisPruner~\cite{vispruner-prune-viz}, and SparseVLM~\cite{sparsevlm-prune-cross} across LLaVA-1.5-7B and LLaVA-1.5-13B models. The experimental settings are identical to those described in Section~\ref{subsec:efficiency-anal-set}. All methods are evaluated under matched token retention budgets.

\subsection{Efficiency Metrics Summary}

We report metrics as defined in Section \ref{subsec:efficiency-anal-set}, along with ~\textbf{Peak GPU Memory}, the maximum memory consumption during inference. For each metric, we compute both absolute values and relative improvements over the baseline (576 visual tokens).

\subsection{LLaVA-1.5-7B Results}

Table~\ref{tab:efficiency-7b} presents detailed efficiency metrics for LLaVA-1.5-7B across five token reduction ratios (25\%--95\%). At moderate reduction levels (50\%), ConsensusDrop achieves a \textbf{1.35$\times$ TTFT speedup} with \textbf{46\% KV cache reduction}, closely matching VisionZip \cite{visionzip-prune-viz} while outperforming SparseVLM in generation latency. At aggressive reduction (90\%), ConsensusDrop delivers \textbf{1.71$\times$ TTFT speedup} and \textbf{83\% KV cache reduction}.

Notably, while VisPruner \cite{visionzip-prune-viz} exhibits marginally faster TTFT at some operating points, this advantage comes at the cost of its iterative pruning loop, which incurs additional computational overhead reflected in higher FLOPs. SparseVLM's \cite{sparsevlm-prune-cross} progressive layer-wise pruning strategy results in significantly higher peak GPU memory (+3.7~GB over ConsensusDrop at 50\% reduction) due to its multi-stage KV cache management.

\begin{table*}[t]
  \centering
  \caption{Efficiency comparison on LLaVA-1.5-7B (576 visual tokens baseline). ConsensusDrop achieves competitive TTFT speedups while maintaining lower memory overhead than SparseVLM. $\dagger$SparseVLM uses progressive layer-wise pruning; token count shown is the target budget (actual final-layer tokens are fewer due to its multi-stage schedule).}
  \label{tab:efficiency-7b}
  \resizebox{\linewidth}{!}{%
    \begin{tabular}{l c c c c c c c c}
    \toprule
    \textbf{Method} & \textbf{Reduction} & \textbf{\# Tokens} & \textbf{TTFT (ms)} & \textbf{Speedup} & \textbf{FLOPs (T)} & \textbf{KV Cache (MB)} & \textbf{GPU Mem (GB)} & \textbf{Gen. Time (ms)} \\
    \midrule
    Baseline & 0\% & 576 & 90.0 & 1.00$\times$ & 8.47 & 313 & 14.50 & 471.6 \\
    \midrule
    \multicolumn{9}{c}{\textit{25\% Token Reduction}} \\
    \midrule
    \textbf{ConsensusDrop (Ours)} & 25\% & 432 & 82.2 & 1.09$\times$ & 6.57 & 241 & 14.48 & 458.0 \\
    VisionZip & 25\% & 432 & 82.5 & 1.09$\times$ & 6.57 & 241 & 14.48 & 464.7 \\
    VisPruner & 25\% & 432 & 57.4 & 1.57$\times$ & 6.95 & 241 & 13.78 & 436.2 \\
    SparseVLM$^\dagger$ & 25\% & 432 & 60.4 & 1.49$\times$ & 5.61 & 212 & 18.15 & 626.7 \\
    \midrule
    \multicolumn{9}{c}{\textit{50\% Token Reduction}} \\
    \midrule
    \textbf{ConsensusDrop (Ours)} & 50\% & 288 & 66.9 & 1.35$\times$ & 4.67 & 169 & 14.48 & 446.7 \\
    VisionZip & 50\% & 288 & 65.2 & 1.38$\times$ & 4.67 & 169 & 14.34 & 440.7 \\
    VisPruner & 50\% & 288 & 50.8 & 1.77$\times$ & 5.05 & 169 & 13.60 & 400.9 \\
    SparseVLM$^\dagger$ & 50\% & 288 & 61.1 & 1.47$\times$ & 3.63 & 148 & 18.15 & 625.9 \\
    \midrule
    \multicolumn{9}{c}{\textit{75\% Token Reduction}} \\
    \midrule
    \textbf{ConsensusDrop (Ours)} & 75\% & 144 & 52.6 & 1.71$\times$ & 2.78 & 97 & 14.48 & 430.7 \\
    VisionZip & 75\% & 144 & 50.9 & 1.77$\times$ & 2.78 & 97 & 14.36 & 437.1 \\
    VisPruner & 75\% & 144 & 46.8 & 1.92$\times$ & 3.16 & 97 & 13.52 & 365.5 \\
    SparseVLM$^\dagger$ & 75\% & 144 & 62.6 & 1.44$\times$ & 2.32 & 96 & 18.15 & 614.1 \\
    \midrule
    \multicolumn{9}{c}{\textit{90\% Token Reduction}} \\
    \midrule
    \textbf{ConsensusDrop (Ours)} & 90\% & 57 & 47.5 & 1.89$\times$ & 1.64 & 54 & 14.44 & 430.5 \\
    VisionZip & 90\% & 57 & 45.9 & 1.96$\times$ & 1.64 & 54 & 14.36 & 434.3 \\
    VisPruner & 90\% & 57 & 48.7 & 1.85$\times$ & 2.02 & 54 & 13.52 & 344.1 \\
    SparseVLM$^\dagger$ & 90\% & 57 & 63.2 & 1.42$\times$ & 1.62 & 56 & 18.15 & 619.6 \\
    \midrule
    \multicolumn{9}{c}{\textit{95\% Token Reduction}} \\
    \midrule
    \textbf{ConsensusDrop (Ours)} & 95\% & 28 & 46.0 & 1.96$\times$ & 1.27 & 39 & 14.48 & 426.8 \\
    VisionZip & 95\% & 28 & 42.7 & 2.11$\times$ & 1.27 & 39 & 14.36 & 434.4 \\
    VisPruner & 95\% & 28 & 46.3 & 1.94$\times$ & 1.64 & 39 & 13.52 & 337.0 \\
    SparseVLM$^\dagger$ & 95\% & 28 & 45.2 & 1.99$\times$ & 1.51 & 41 & 18.15 & 613.8 \\
    \bottomrule
    \end{tabular}%
  }
\end{table*}

\begin{figure}
  \centering
  \includegraphics[width=\columnwidth]{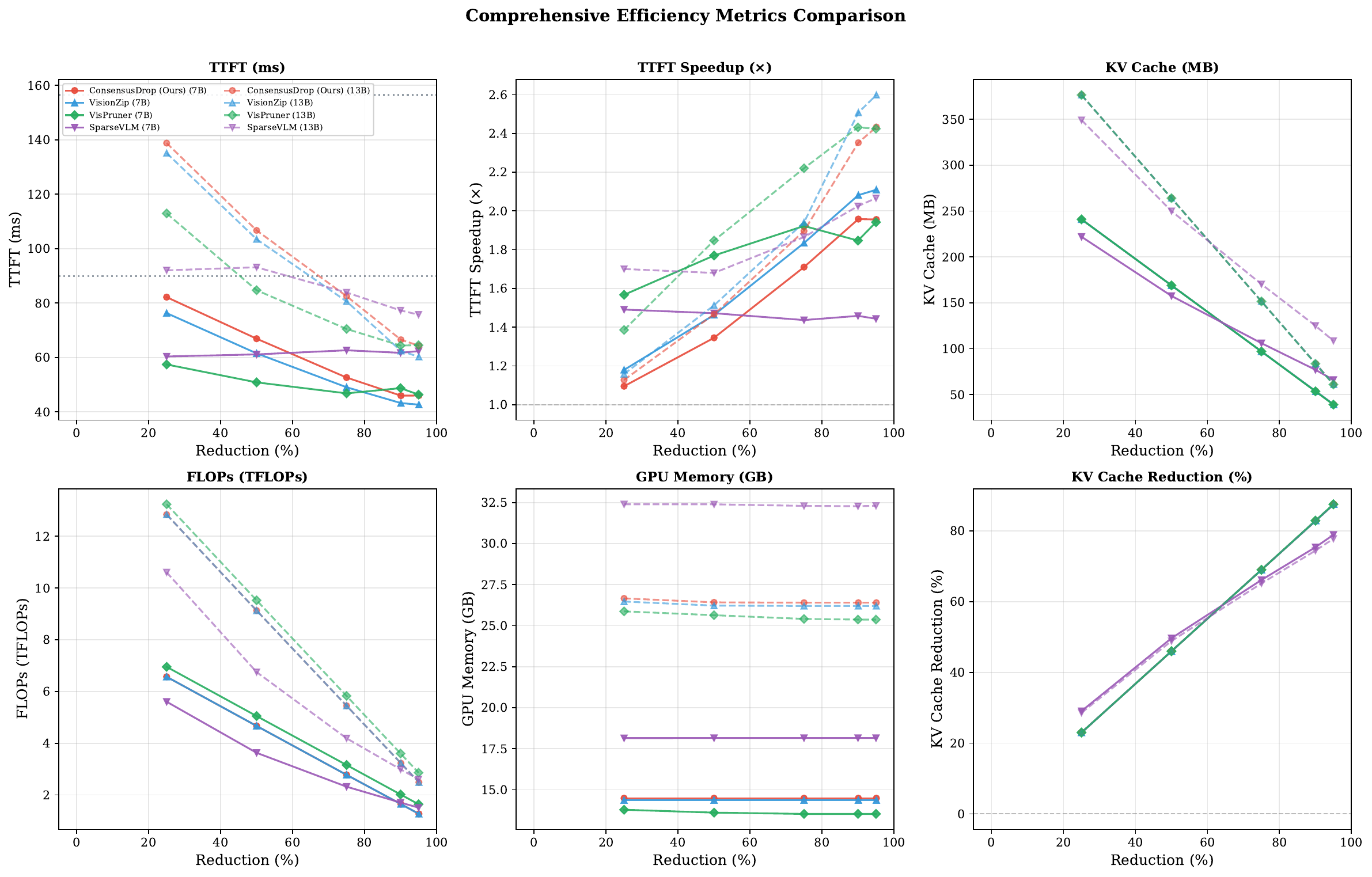}
  \caption{Comprehensive efficiency metrics comparison across token reduction ratios. Each subplot shows a different metric: \textbf{(a)}~TTFT latency, \textbf{(b)}~TTFT speedup over baseline, \textbf{(c)}~KV cache memory footprint, \textbf{(d)}~computational FLOPs, \textbf{(e)}~peak GPU memory, and \textbf{(f)}~KV cache reduction percentage. Solid lines indicate LLaVA-1.5-7B; dashed lines indicate LLaVA-1.5-13B. ConsensusDrop (red) achieves efficiency profiles comparable to VisionZip while avoiding the memory overhead of SparseVLM.}
  \label{fig:comprehensive-efficiency}
\end{figure}

\subsection{LLaVA-1.5-13B Results}

Table~\ref{tab:efficiency-13b} extends the analysis to the larger LLaVA-1.5-13B model. The efficiency gains from visual token pruning scale consistently with model size. At 50\% reduction, ConsensusDrop achieves \textbf{1.56$\times$ TTFT speedup} and \textbf{46\% KV cache reduction}. The larger model amplifies the memory advantages of ConsensusDrop over SparseVLM, with the latter requiring \textbf{+6.0~GB additional GPU memory} at the 50\% reduction level.

\subsection{Efficiency vs. Accuracy Trade-off}

The key insight from our efficiency analysis is that \textbf{ConsensusDrop achieves Pareto-optimal accuracy--efficiency trade-offs}. While VisPruner demonstrates faster TTFT at certain operating points, this comes at the cost of (i)~higher computational FLOPs due to its iterative pruning loop, and (ii)~reduced accuracy (and inference performance) at aggressive reduction ratios (Section \ref{sec:exp}). Similarly, while SparseVLM's progressive pruning can achieve competitive TTFT, it requires substantially more GPU memory and exhibits slower end-to-end generation times.

ConsensusDrop's efficiency profile is characterized by:
\begin{itemize}
    \item \textbf{Consistent TTFT speedups} (1.35--2.0$\times$) that scale with token reduction ratio
    \item \textbf{Linear KV cache reduction} directly proportional to token reduction
    \item \textbf{Minimal memory overhead} compared to the baseline, unlike SparseVLM's +3--6~GB overhead
    \item \textbf{Stable generation latency} that does not degrade with aggressive pruning
\end{itemize}

\begin{figure}
  \centering
  \includegraphics[width=\columnwidth]{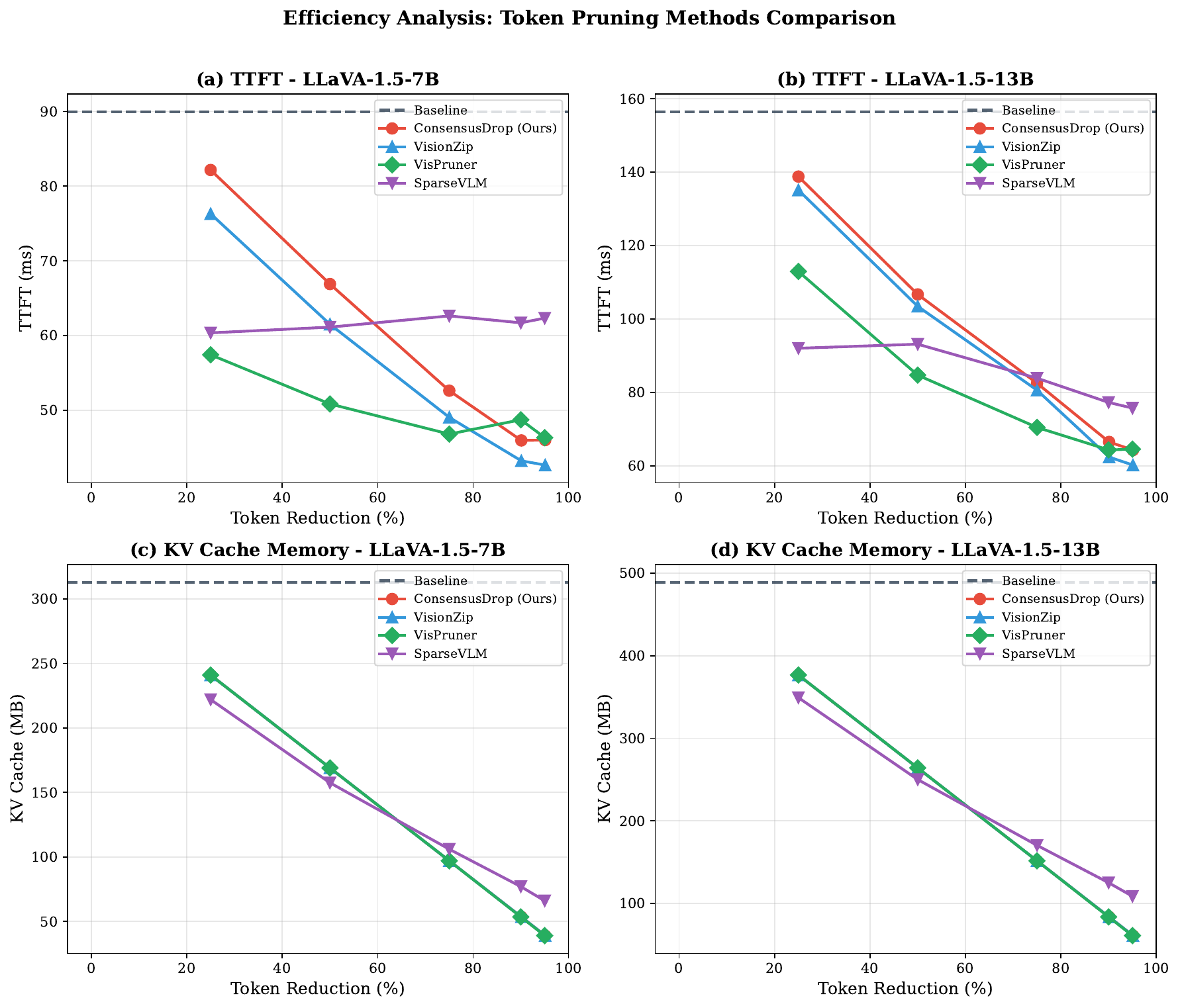}
  \caption{Efficiency comparison across token reduction ratios. \textbf{(a, b)}~TTFT decreases consistently with token reduction for all methods. \textbf{(c, d)}~KV cache memory scales linearly with retained tokens. ConsensusDrop (green/red/blue; coinciding) achieves competitive efficiency while avoiding the memory overhead of SparseVLM (purple) and the computational overhead of VisPruner.}
  \label{fig:efficiency-summary}
\end{figure}

Figure~\ref{fig:efficiency-summary} visualizes these trends across both model sizes, demonstrating that ConsensusDrop maintains favorable efficiency characteristics while preserving accuracy (Section \ref{sec:exp}).

\begin{figure}
  \centering
  \includegraphics[width=\columnwidth]{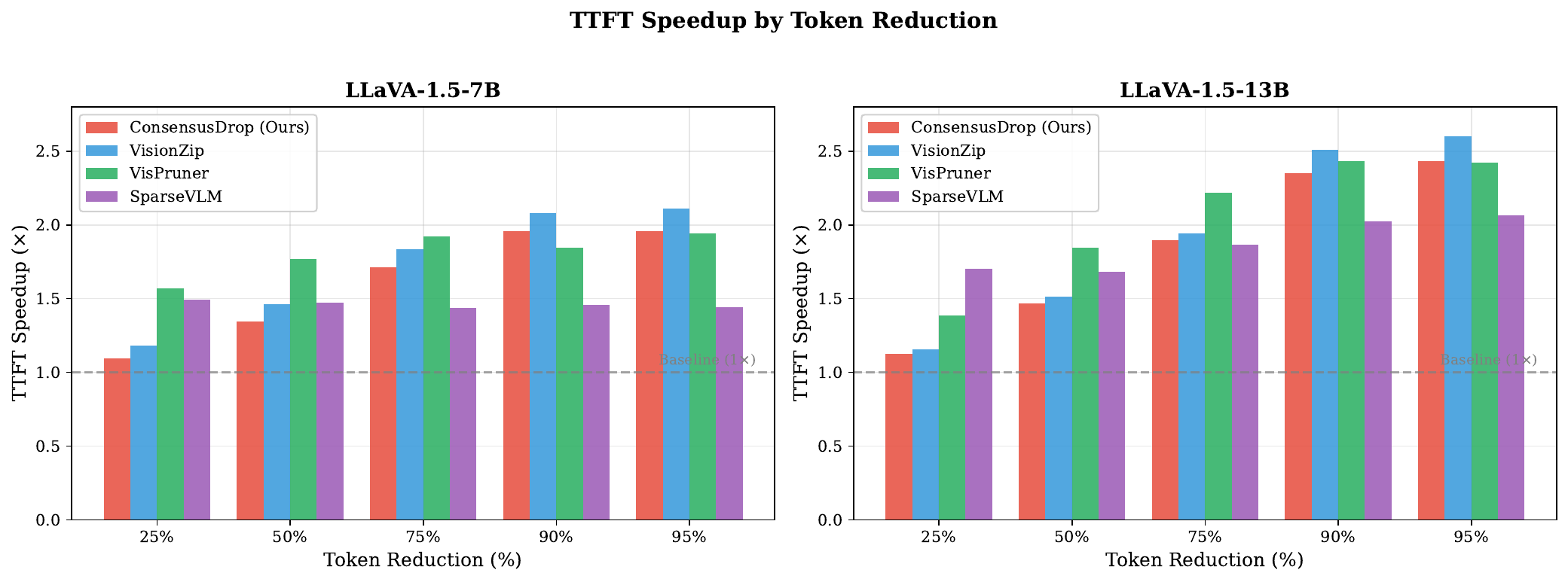}
  \caption{TTFT speedup comparison at fixed reduction levels. ConsensusDrop achieves speedups comparable to VisionZip while providing superior accuracy through multimodal consensus. SparseVLM's speedups plateau due to its progressive pruning overhead.}
  \label{fig:ttft-speedup-bars}
\end{figure}

Figure~\ref{fig:ttft-speedup-bars} provides a grouped comparison of TTFT speedups across methods at fixed reduction levels. At moderate reduction (50\%), all methods achieve meaningful speedups (1.35--1.77$\times$ for 7B), with differences becoming more pronounced at higher reduction ratios. Importantly, ConsensusDrop's speedup profile closely tracks that of VisionZip, confirming that our multimodal consensus mechanism introduces negligible overhead relative to vision-only pruning.

\label{app:visuals}

\section{Qualitative Visualizations}
\label{app:visuals}

\input{visualizations_for_paper/qualitative_figures}

\section{Limitations and Discussion}
\label{app:limitations}
ConsensusDrop relies on the availability of meaningful cross-modal attention signals; for prompts with extremely weak or ambiguous visual grounding, the benefit of consensus-based fusion may diminish. While SCAP enables efficient pre-LLM extraction of cross-modal cues, it remains a lightweight approximation of full LLM cross-attention and may not fully capture higher-layer semantic interactions. Our approach focuses on visual token reduction and does not address redundancy in text tokens or inter-layer token dynamics within the LLM. Finally, although encoder-guided token merging preserves semantic context in practice, aggressive compression may still lead to loss of fine-grained spatial details in highly localized visual reasoning tasks.

\section{Reproducibility}
\label{app:reproducibility}
Our implementation builds upon the official open-source codebases of LLaVA \cite{llava}, Video-LLaVA \cite{lin2023videollava}, and Qwen-VL \cite{qwen-vl}, with all model variants, preprocessing pipelines, and evaluation protocols preserved unless explicitly stated. We will release the full ConsensusDrop codebase, including inference scripts, configuration files, and evaluation utilities, to enable reproduction of all reported results. The main paper and appendix provide exhaustive details on experimental settings, token retention budgets, efficiency measurement, and ablation configurations to ensure transparency and reproducibility.

\section{Use of Large Language Models}
\label{app:llm-usage}
Large Language Models were used as auxiliary tools to assist with code refactoring, experiment orchestration across ablation settings, and minor editing for clarity and grammar after the core methods, implementations, and experiments were developed by the authors. All algorithmic design, experimental decisions, result interpretation, and final writing remain the sole responsibility of the authors.

%% file: visualizations_for_paper/qualitative_figures.tex
We present qualitative visualizations demonstrating ConsensusDrop's token selection mechanism (examples from VQAv2 \cite{goyal2017vqav2} test-dev2015 split, using LLaVA-1.5-7B as the base VLM). Our consensus fusion approach combines vision-based saliency scores with cross-modal attention to identify tokens that are both visually salient and semantically relevant to the question. 

\subsection{Token Selection Comparison}
\begin{figure*}
\centering
\includegraphics[width=\textwidth]{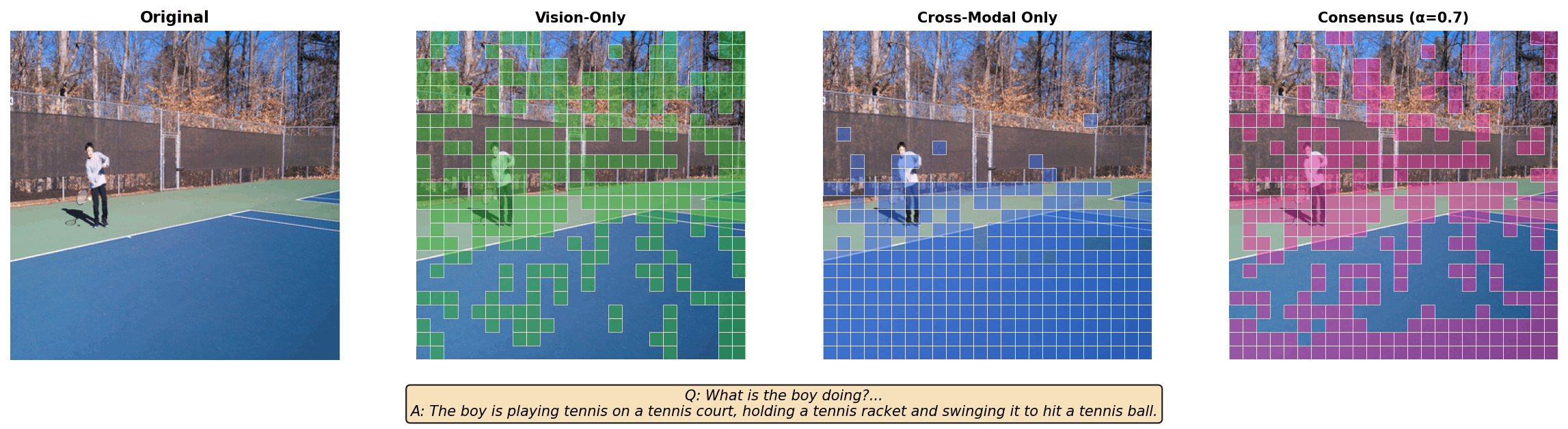}
\includegraphics[width=\textwidth]{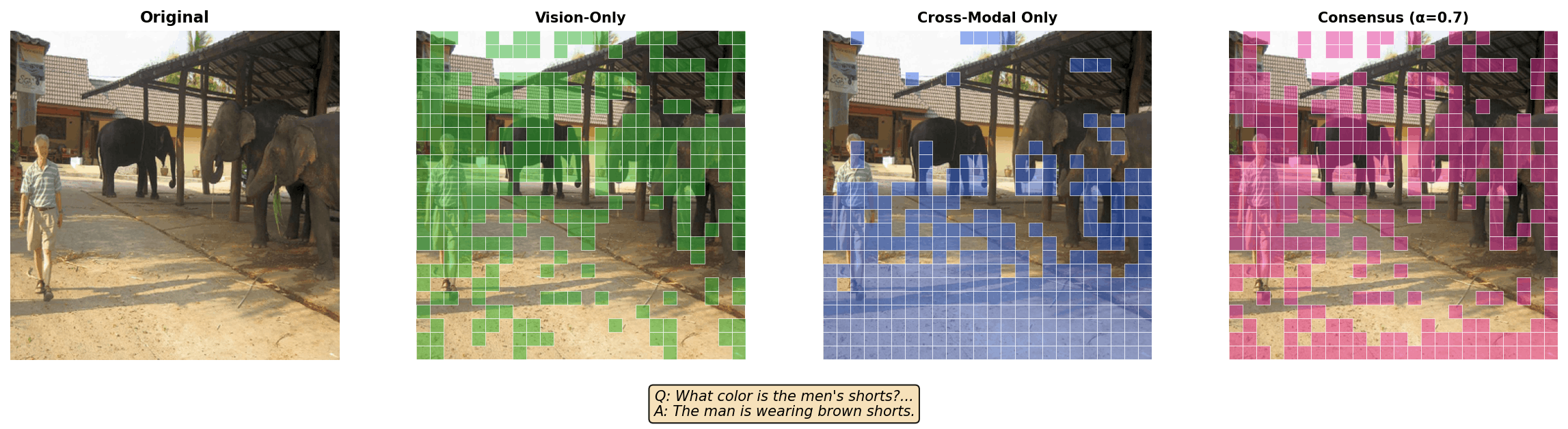}
\\[2mm]
\includegraphics[width=\textwidth]{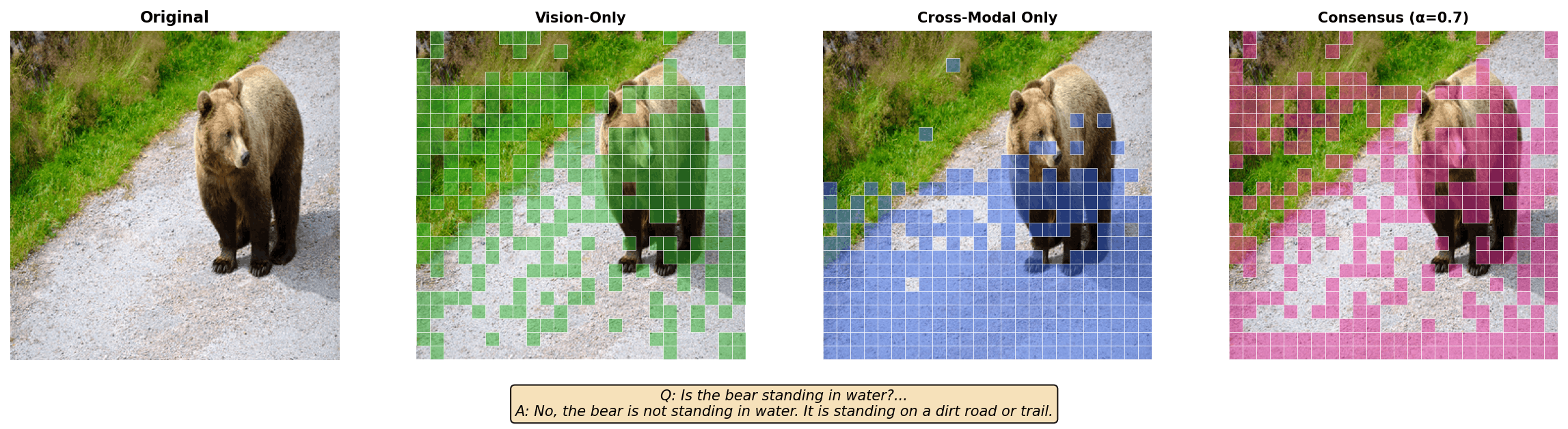}
\includegraphics[width=\textwidth]{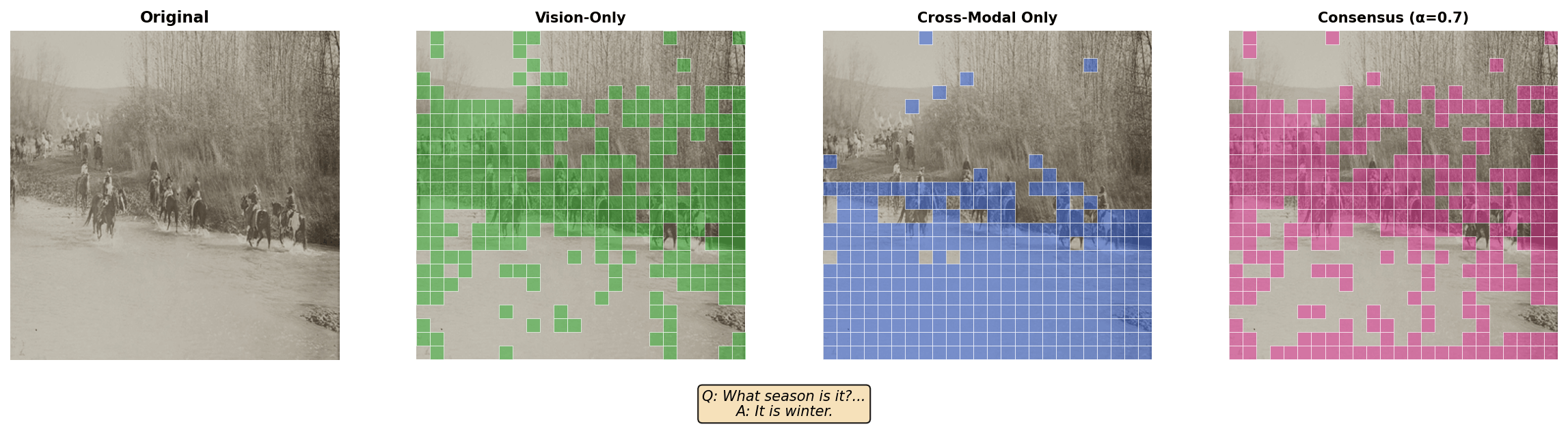}

\caption{Token selection comparison at $K=288$ (50\% reduction). Each panel shows: Original image, Vision-Only selection (green), Cross-Modal Only selection (blue), and Consensus fusion (magenta). Questions and model answers are shown below each comparison.}
\label{fig:comparison_examples}
\end{figure*}

\begin{figure*}
\centering
\includegraphics[width=\textwidth]{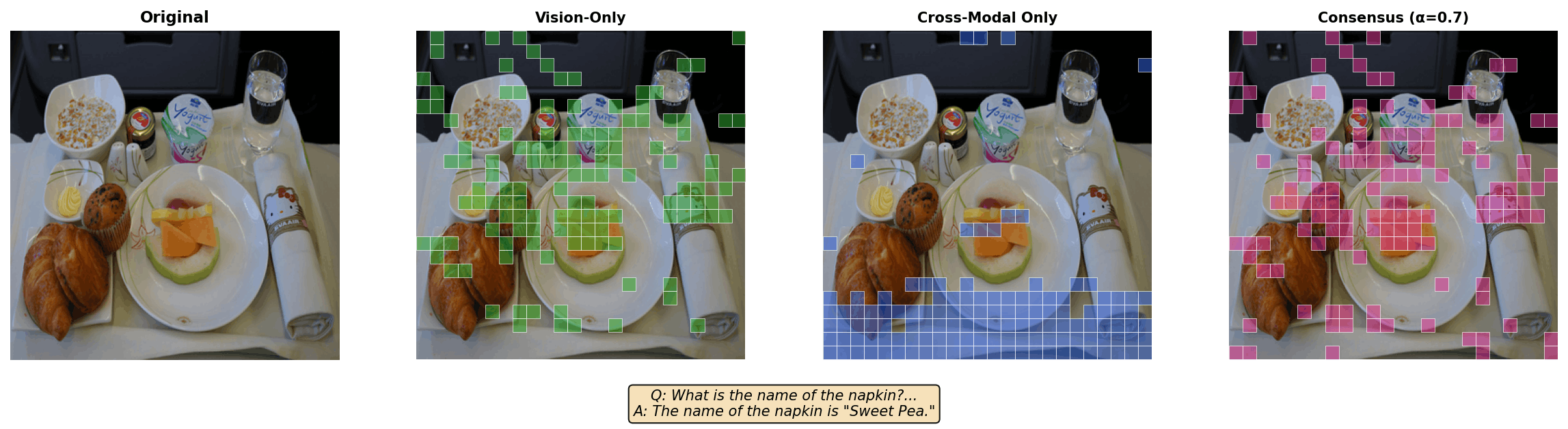}
\includegraphics[width=\textwidth]{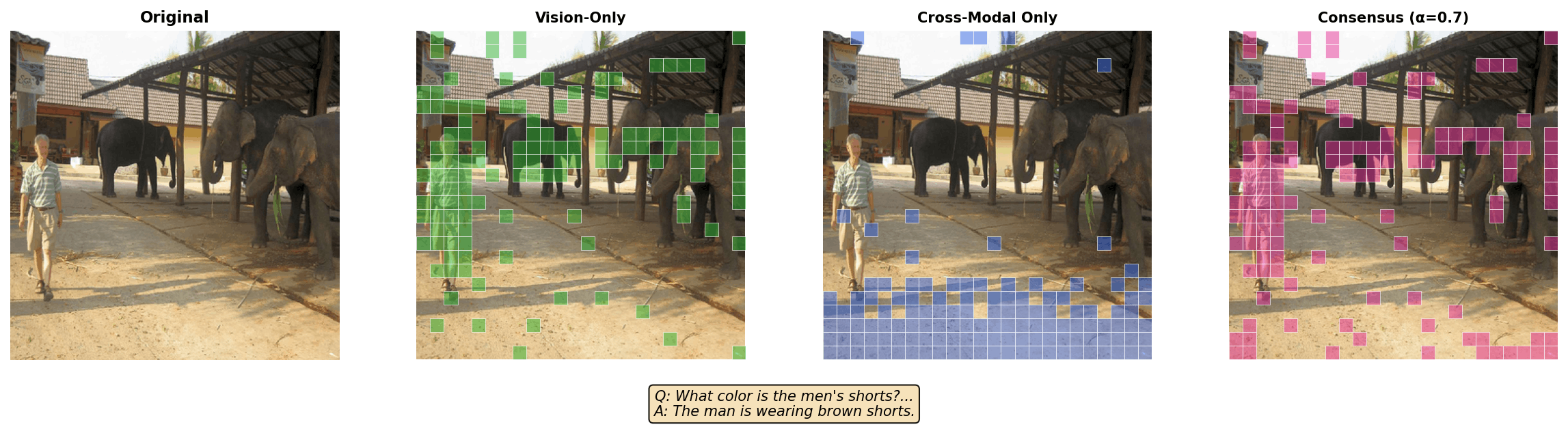}
\\[2mm]
\includegraphics[width=\textwidth]{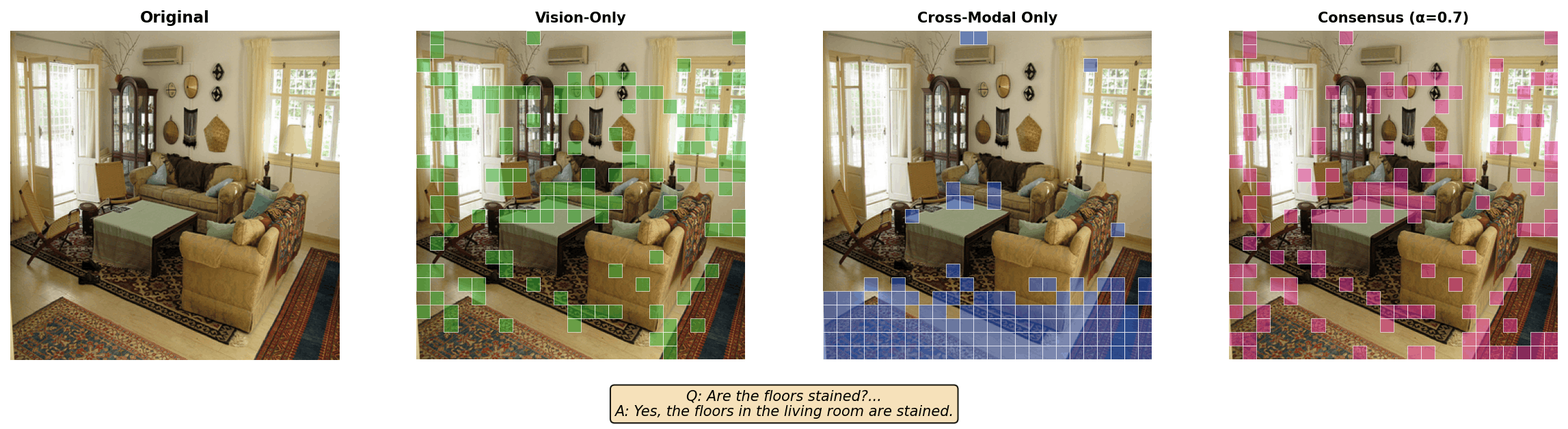}
\includegraphics[width=\textwidth]{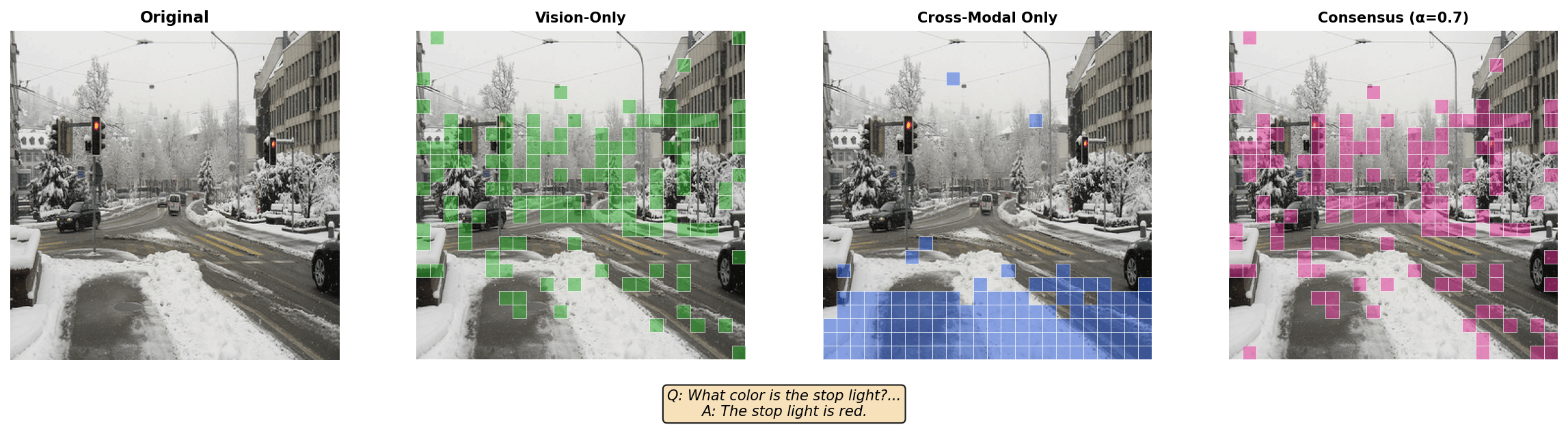}

\caption{Token selection comparison at $K=128$ (78\% reduction). At higher compression, the disagreement between vision and cross-modal signals becomes more pronounced, making consensus fusion even more critical.}
\label{fig:comparison_examples_k128}
\end{figure*}

Figure~\ref{fig:comparison_examples} and \ref{fig:comparison_examples_k128} shows side-by-side comparisons of token selection strategies: \textbf{Vision-Only} selects tokens based on visual saliency from the CLIP encoder, \textbf{Cross-Modal Only} uses attention scores from text-to-image attention, and \textbf{Consensus} combines both through our convex fusion ($\alpha=0.7$). The visualizations reveal that vision-only selection often focuses on globally salient regions (e.g., high-contrast edges, textures) while missing question-relevant details. Cross-modal attention, affected by positional biases (RoPE), tends to over-weight late positions (bottom image patches). Our consensus fusion balances these complementary signals, retaining tokens that are both visually informative and semantically aligned with the question.

\subsection{High-Disagreement Examples}
\begin{figure}
\centering
% Heatmap
\begin{subfigure}[b]{0.48\textwidth}
\includegraphics[width=\textwidth]{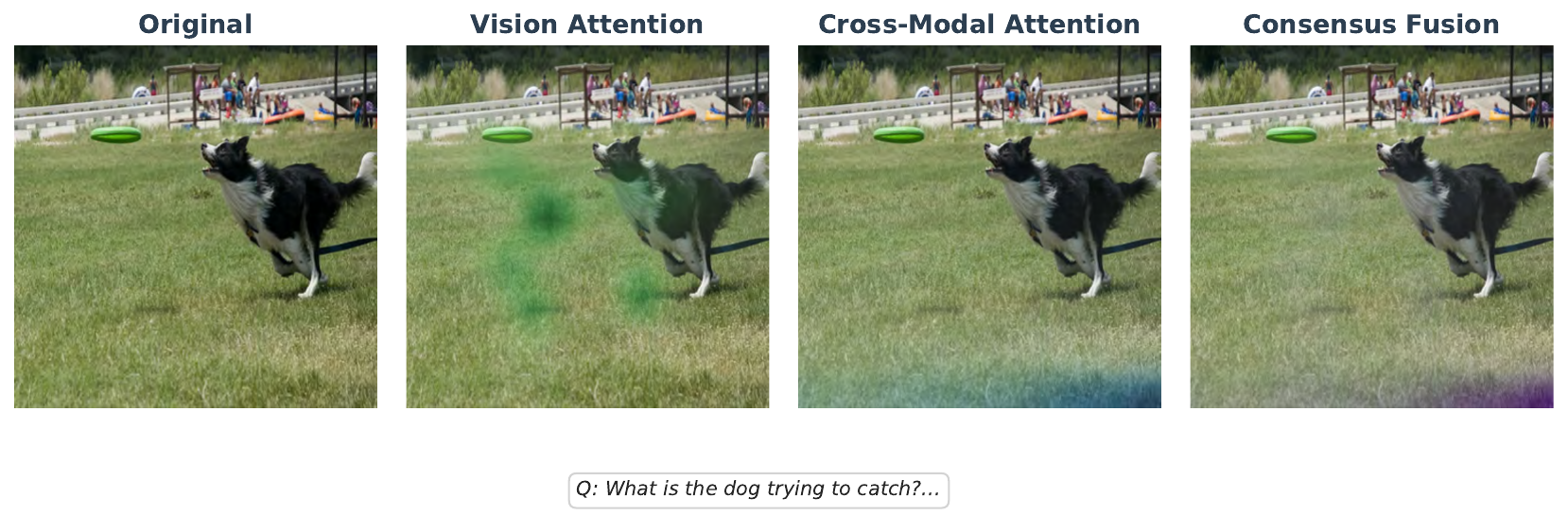}
\caption{Attention heatmaps}
\end{subfigure}
\hfill
% Venn
\begin{subfigure}[b]{0.48\textwidth}
\includegraphics[width=\textwidth]{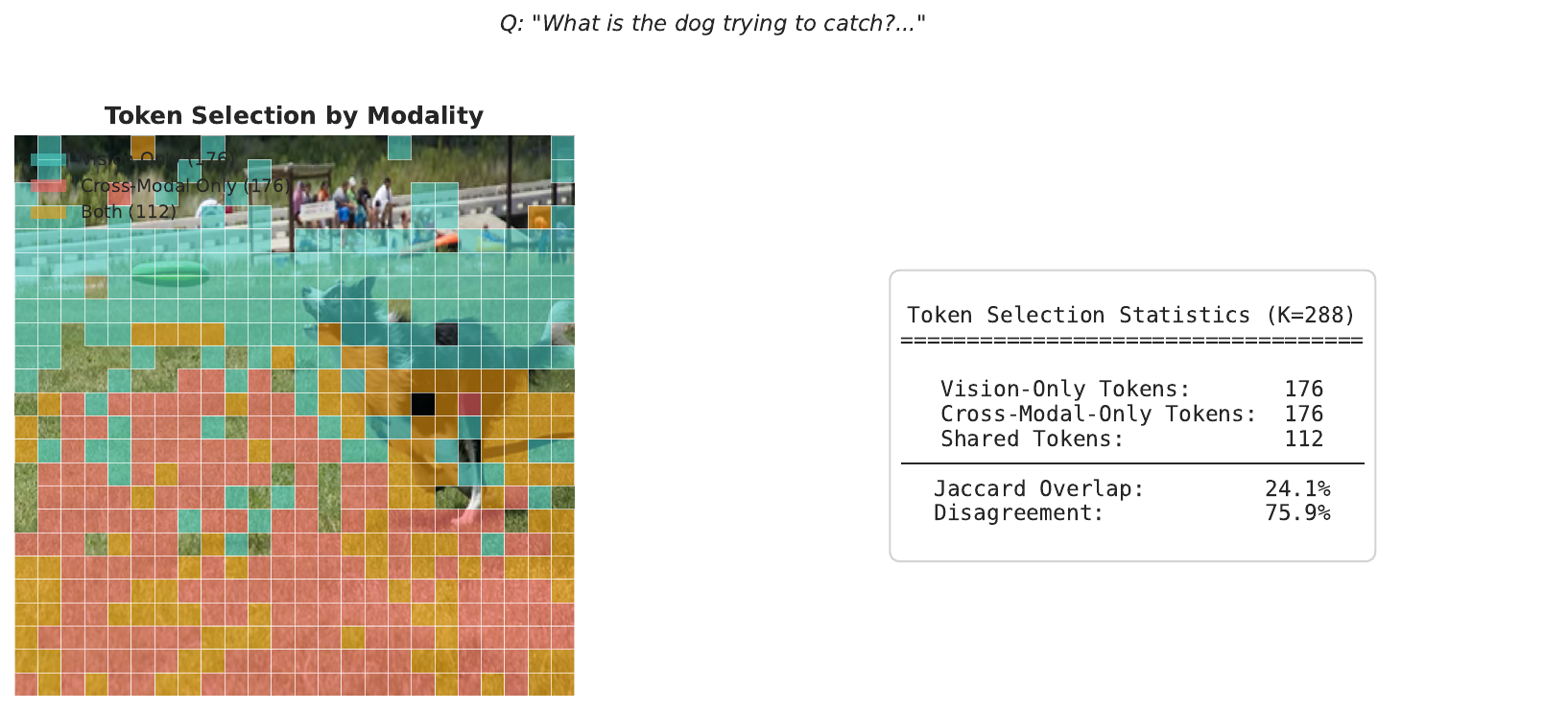}
\caption{Token set overlap}
\end{subfigure}
\\[2mm]
% Minimal
\begin{subfigure}[b]{0.48\textwidth}
\includegraphics[width=\textwidth]{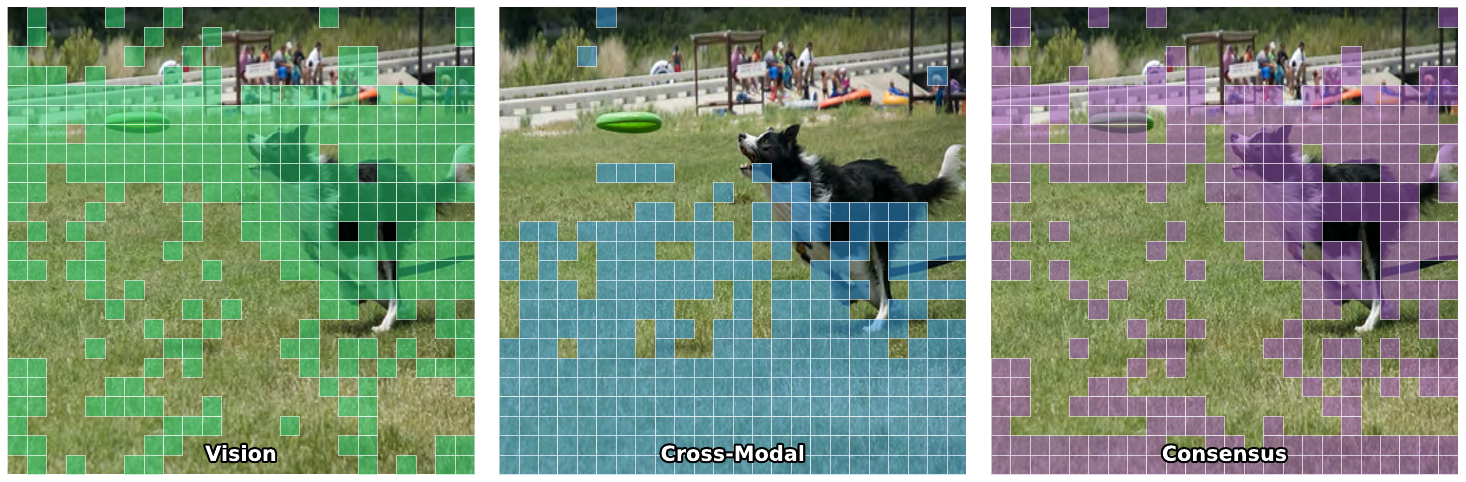}
\caption{Minimal comparison}
\end{subfigure}
\hfill
% Multi-K
\begin{subfigure}[b]{0.48\textwidth}
\includegraphics[width=\textwidth]{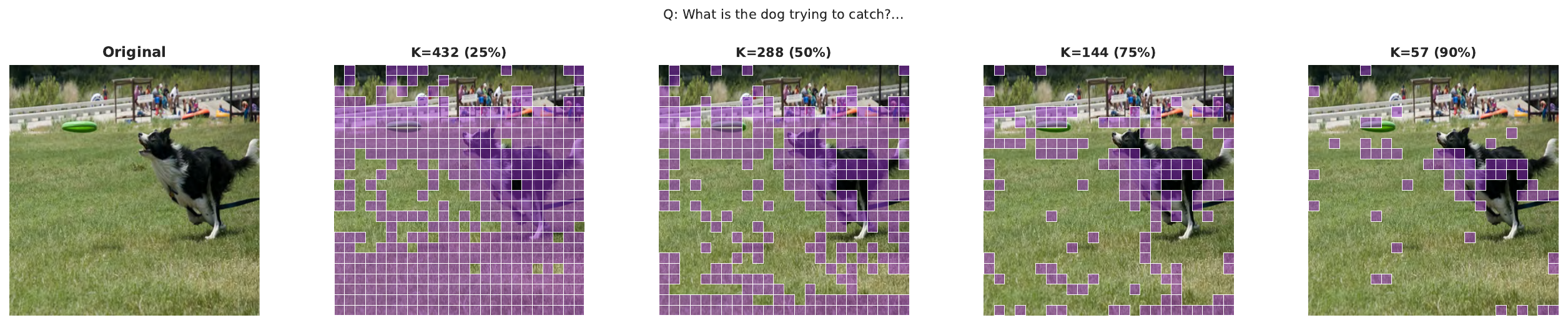}
\caption{Across retention ratios}
\end{subfigure}
\caption[Token selection]{\textbf{Q:} ``What is the dog trying to catch?" \textbf{A:} ``The dog is trying to catch a frisbee...." (Disagreement: 61.1\%)}
\label{fig:k288_detailed_1}
\end{figure}

\begin{figure}
\centering
% Heatmap
\begin{subfigure}[b]{0.48\textwidth}
\includegraphics[width=\textwidth]{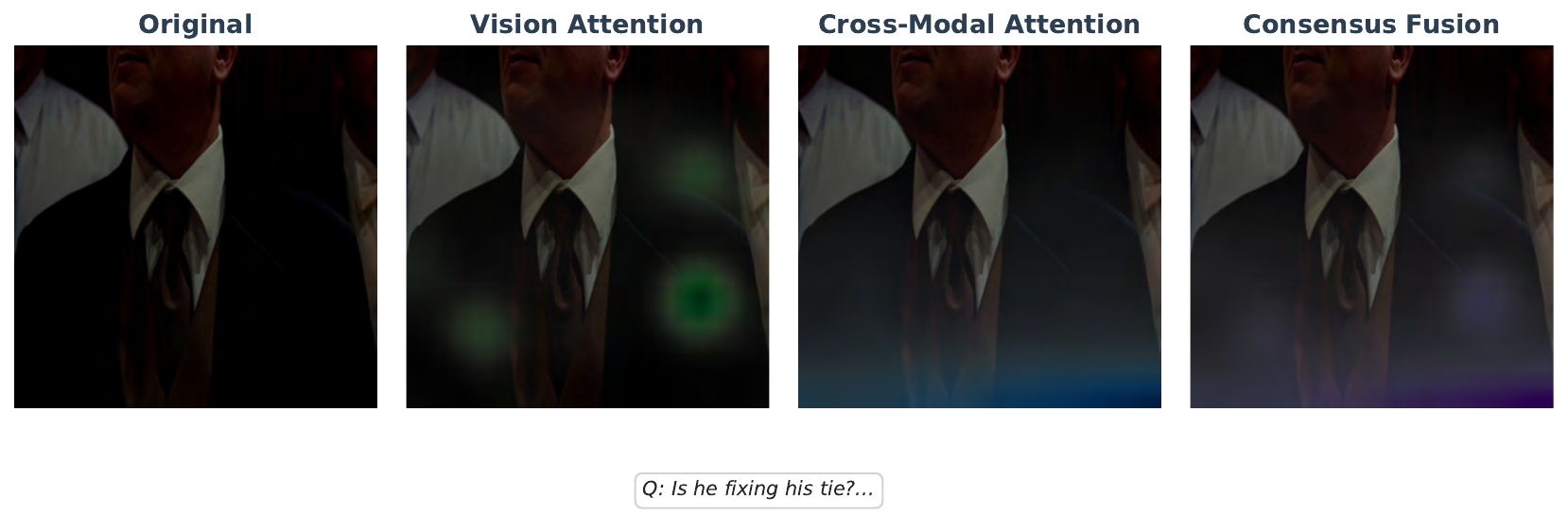}
\caption{Attention heatmaps}
\end{subfigure}
\hfill
% Venn
\begin{subfigure}[b]{0.48\textwidth}
\includegraphics[width=\textwidth]{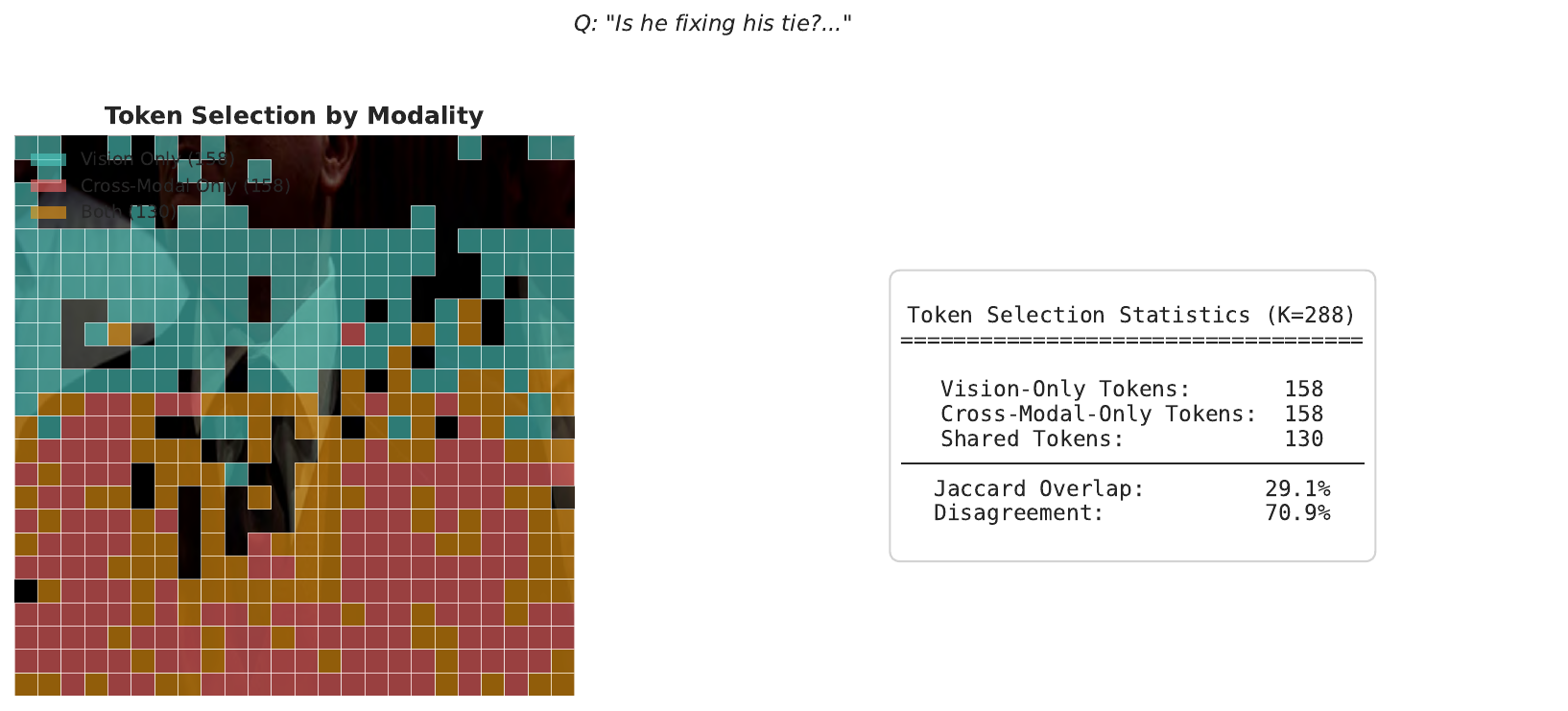}
\caption{Token set overlap}
\end{subfigure}
\\[2mm]
% Minimal
\begin{subfigure}[b]{0.48\textwidth}
\includegraphics[width=\textwidth]{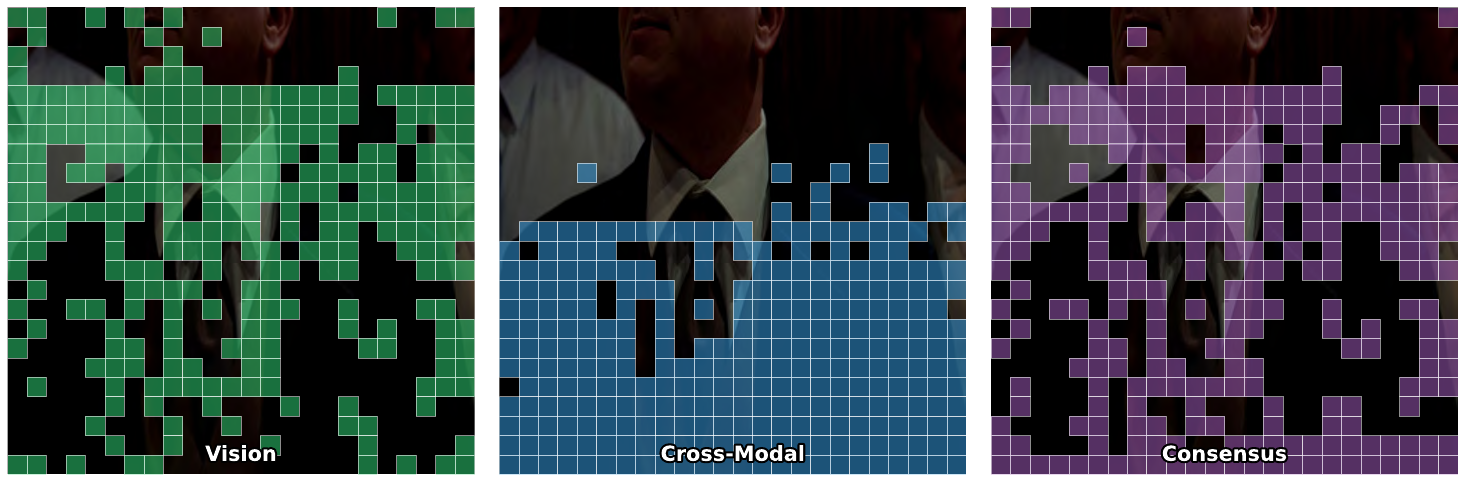}
\caption{Minimal comparison}
\end{subfigure}
\hfill
% Multi-K
\begin{subfigure}[b]{0.48\textwidth}
\includegraphics[width=\textwidth]{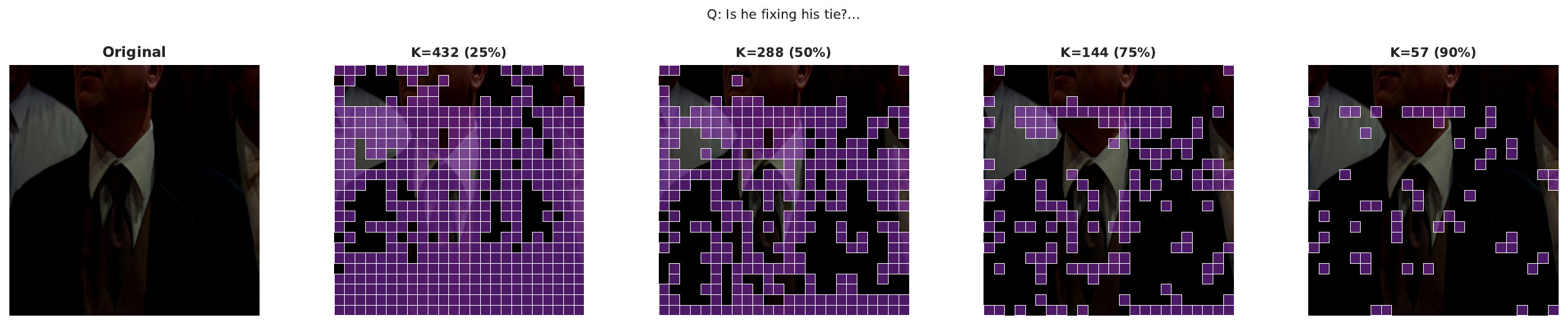}
\caption{Across retention ratios}
\end{subfigure}
\caption[Token selection]{\textbf{Q:} ``Is he fixing his tie?" \textbf{A:} ``Yes, the man in the image is fixing his tie, which is a common activity before a..." (Disagreement: 54.9\%)}
\label{fig:k288_detailed_2}
\end{figure}

\begin{figure}
\centering
% Heatmap
\begin{subfigure}[b]{0.48\textwidth}
\includegraphics[width=\textwidth]{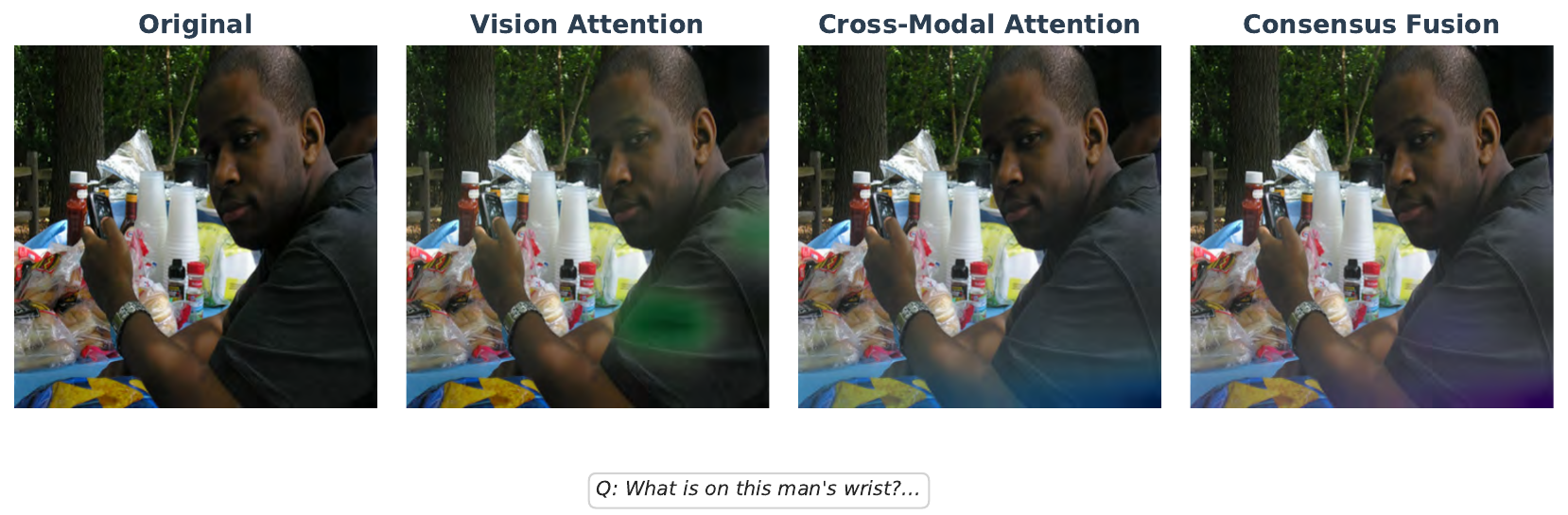}
\caption{Attention heatmaps}
\end{subfigure}
\hfill
% Venn
\begin{subfigure}[b]{0.48\textwidth}
\includegraphics[width=\textwidth]{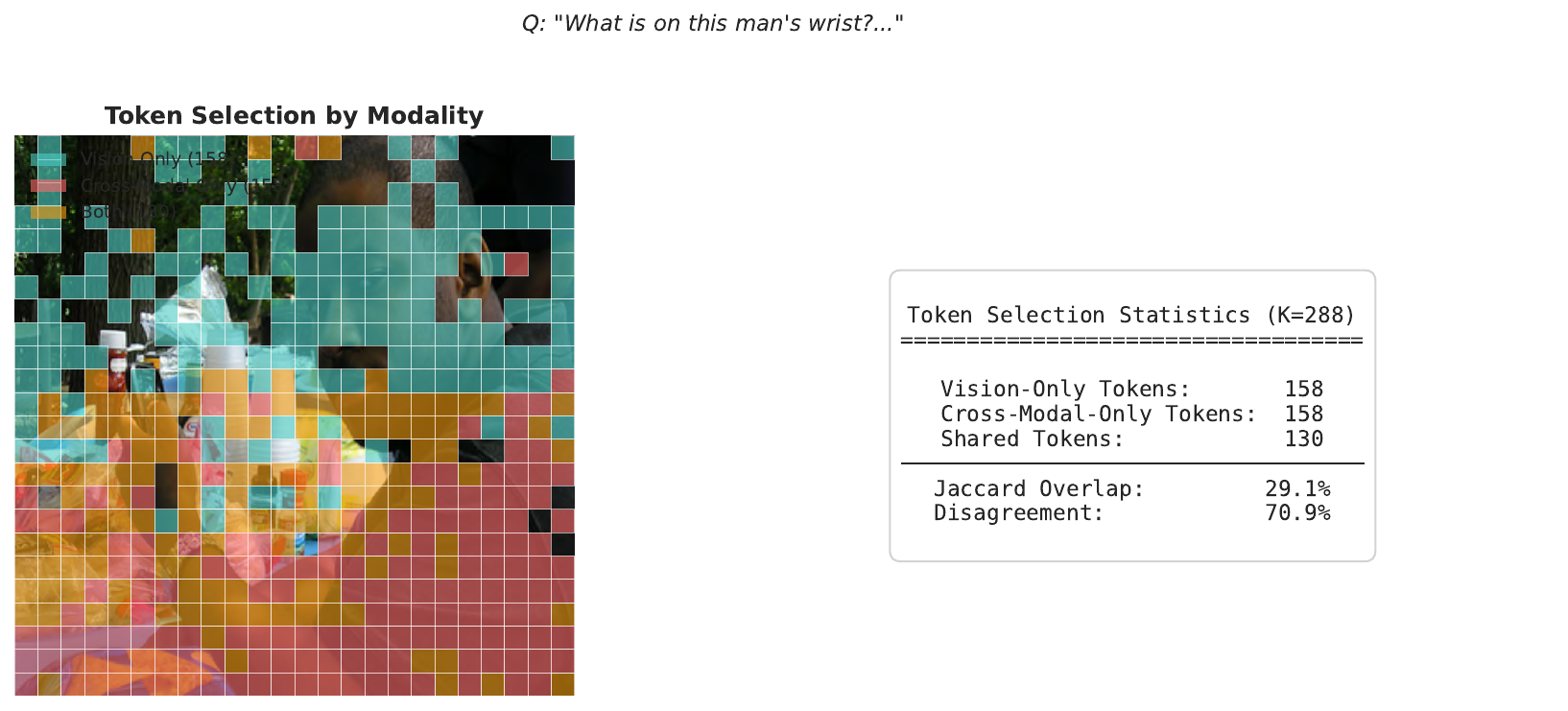}
\caption{Token set overlap}
\end{subfigure}
\\[2mm]
% Minimal
\begin{subfigure}[b]{0.48\textwidth}
\includegraphics[width=\textwidth]{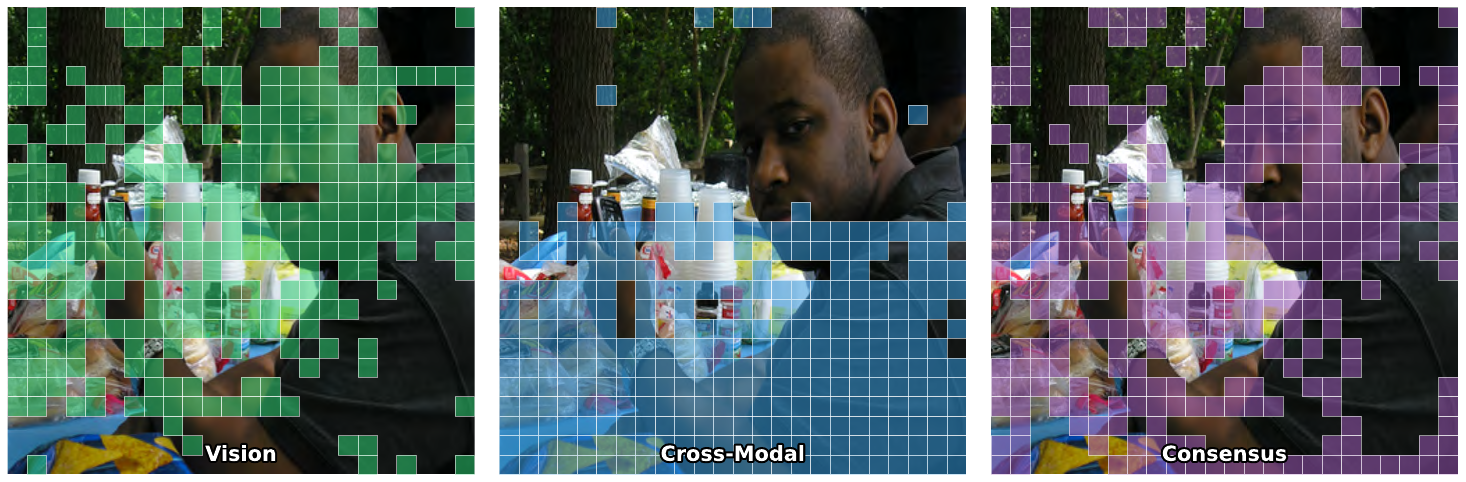}
\caption{Minimal comparison}
\end{subfigure}
\hfill
% Multi-K
\begin{subfigure}[b]{0.48\textwidth}
\includegraphics[width=\textwidth]{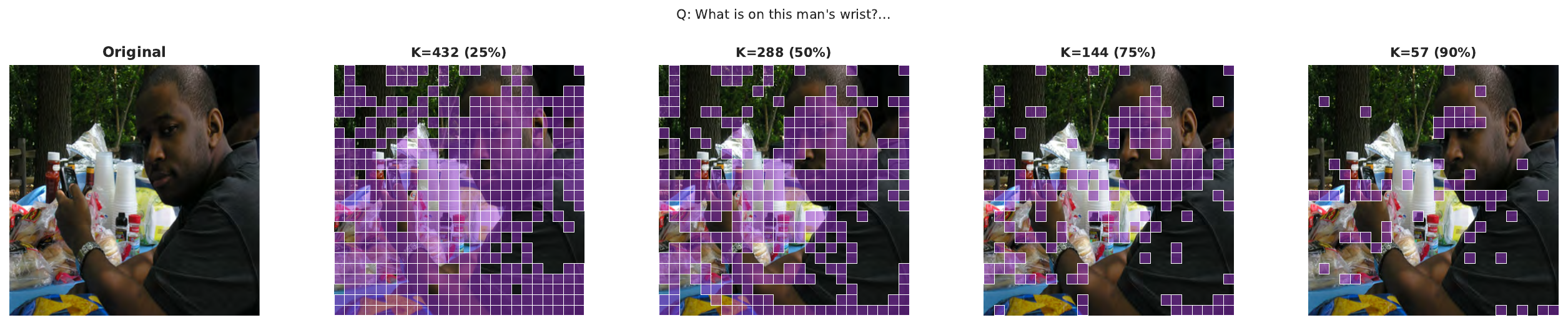}
\caption{Across retention ratios}
\end{subfigure}
\caption[Token selection]{\textbf{Q:} ``What is on this man's wrist?" \textbf{A:} ``The man is wearing a watch on his wrist...." (Disagreement: 54.9\%)}
\label{fig:k288_detailed_3}
\end{figure}

\begin{figure}
\centering
% Heatmap
\begin{subfigure}[b]{0.48\textwidth}
\includegraphics[width=\textwidth]{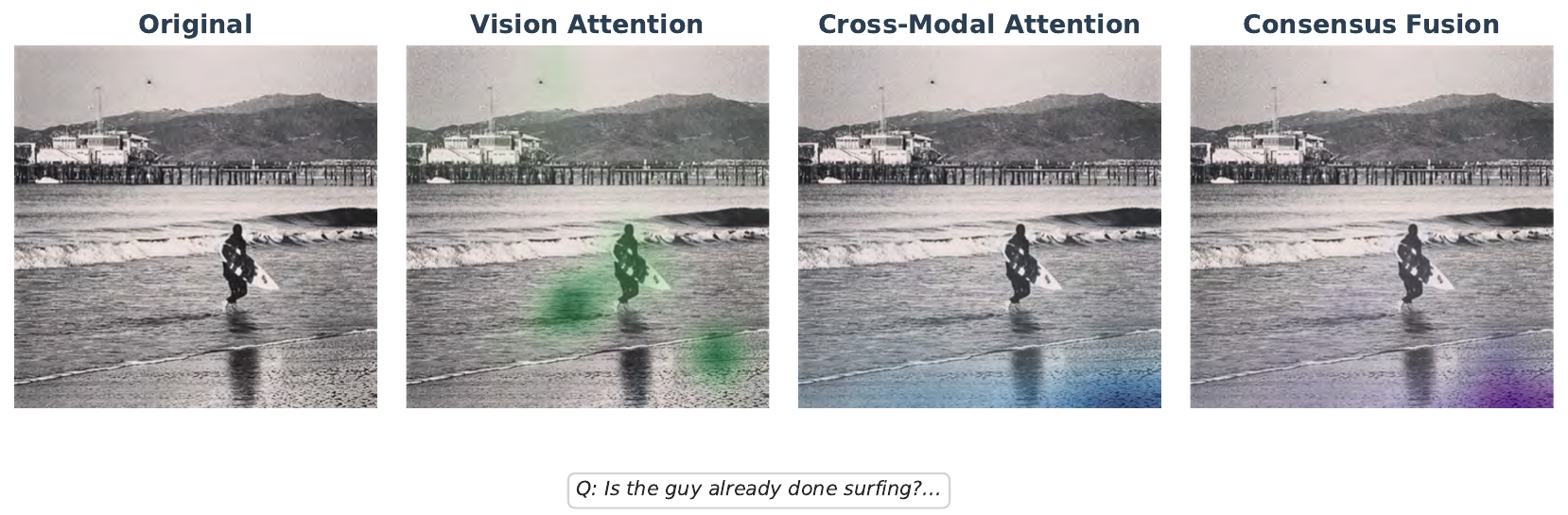}
\caption{Attention heatmaps}
\end{subfigure}
\hfill
% Venn
\begin{subfigure}[b]{0.48\textwidth}
\includegraphics[width=\textwidth]{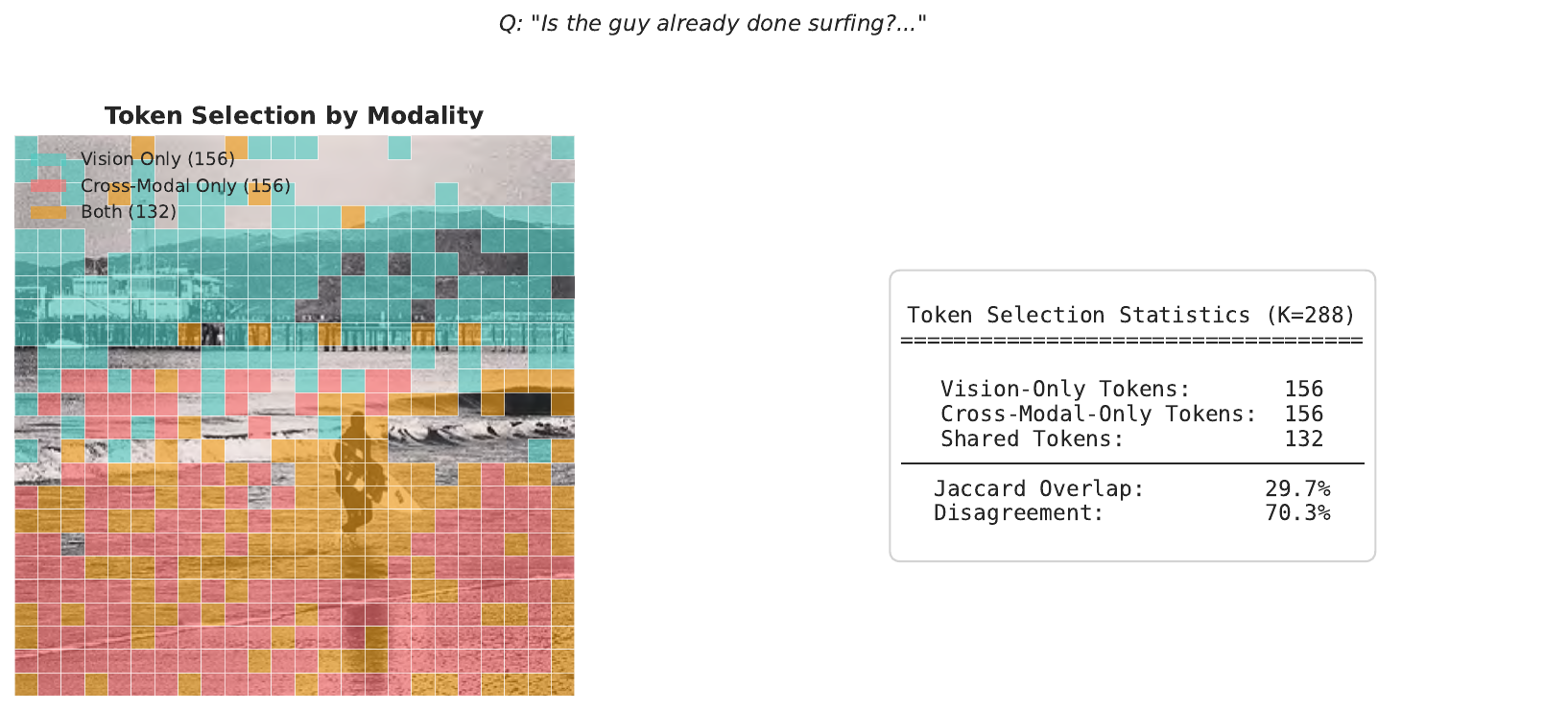}
\caption{Token set overlap}
\end{subfigure}
\\[2mm]
% Minimal
\begin{subfigure}[b]{0.48\textwidth}
\includegraphics[width=\textwidth]{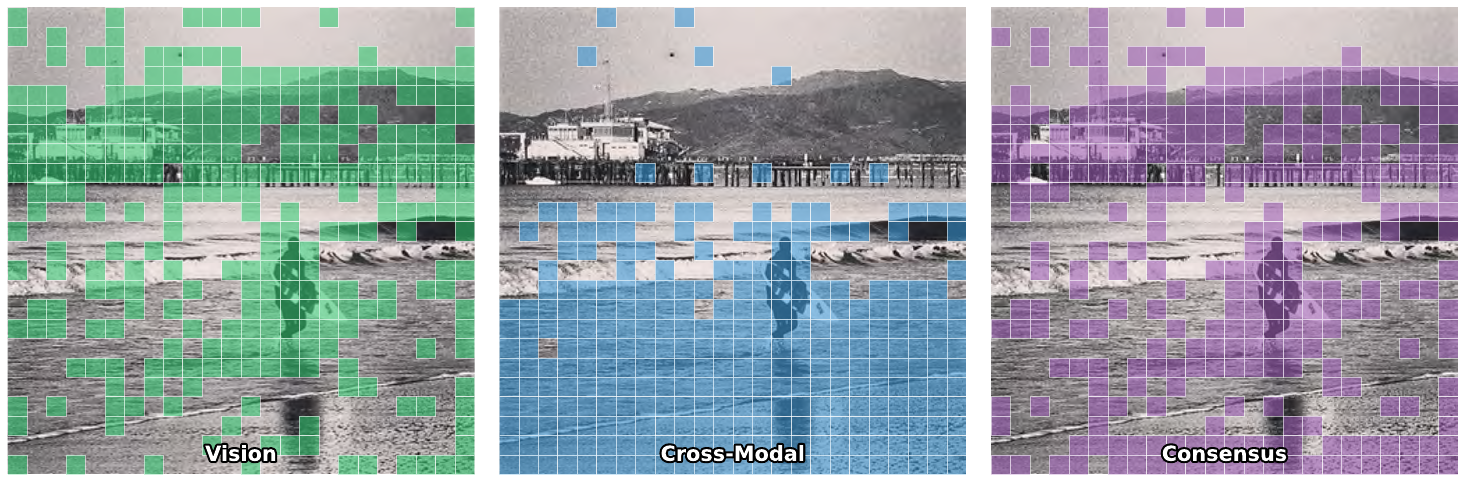}
\caption{Minimal comparison}
\end{subfigure}
\hfill
% Multi-K
\begin{subfigure}[b]{0.48\textwidth}
\includegraphics[width=\textwidth]{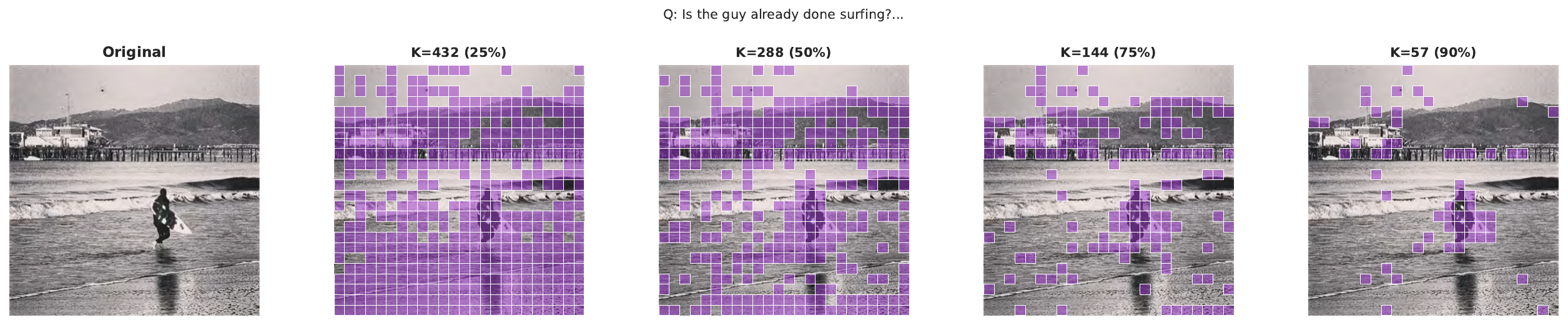}
\caption{Across retention ratios}
\end{subfigure}
\caption[Token selection]{\textbf{Q:} ``Is the guy already done surfing?" \textbf{A:} ``Yes, the guy is already done surfing and is walking out of the water with his su..." (Disagreement: 54.2\%)}
\label{fig:k288_detailed_4}
\end{figure}

\begin{figure}
\centering
% Heatmap
\begin{subfigure}[b]{0.48\textwidth}
\includegraphics[width=\textwidth]{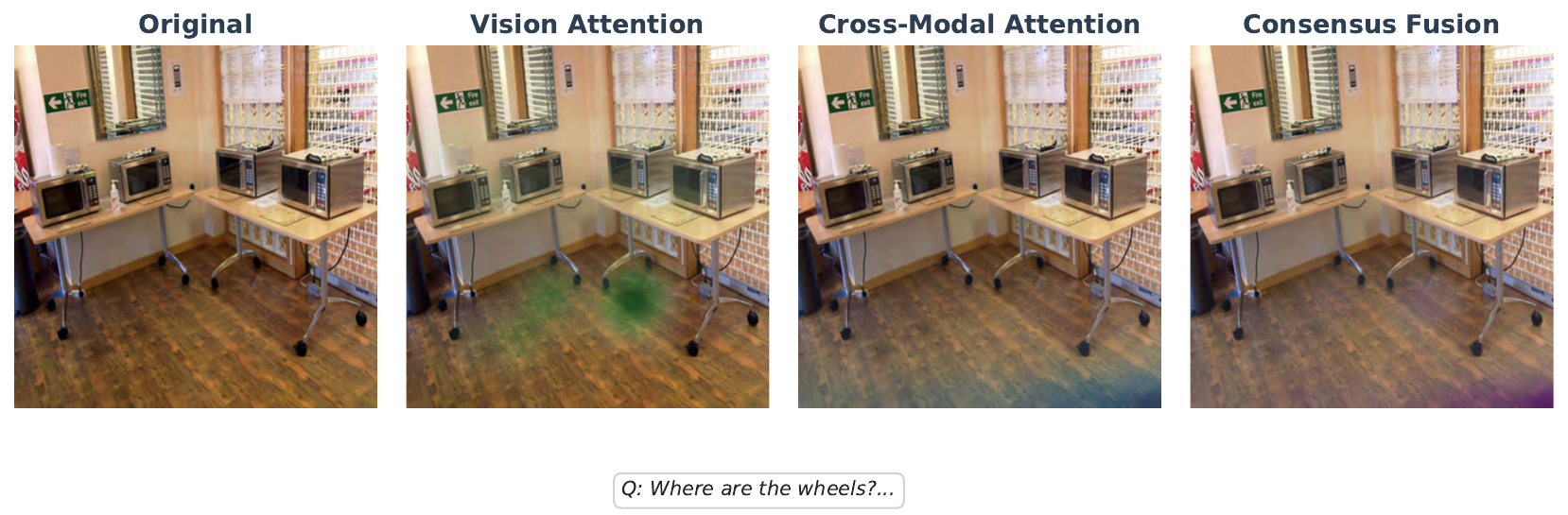}
\caption{Attention heatmaps}
\end{subfigure}
\hfill
% Venn
\begin{subfigure}[b]{0.48\textwidth}
\includegraphics[width=\textwidth]{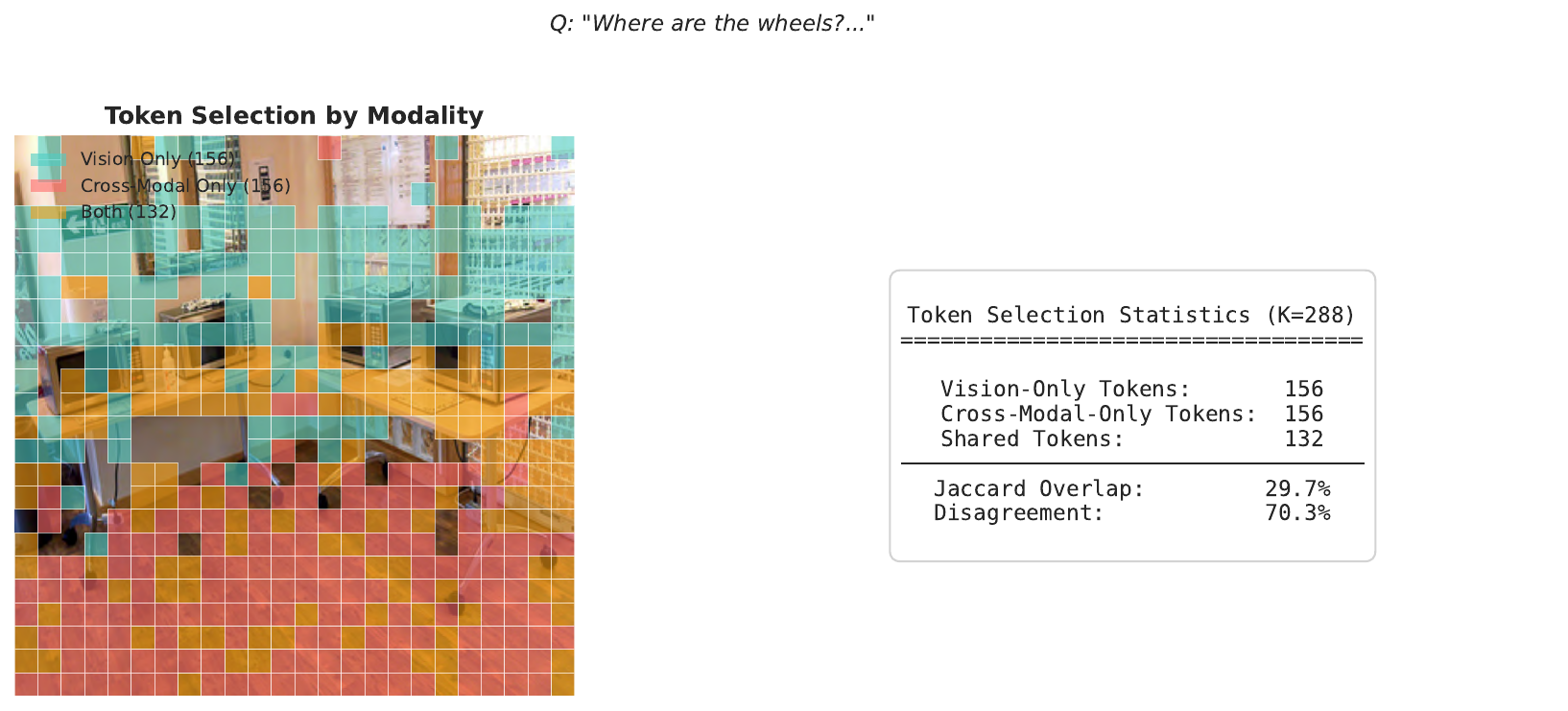}
\caption{Token set overlap}
\end{subfigure}
\\[2mm]
% Minimal
\begin{subfigure}[b]{0.48\textwidth}
\includegraphics[width=\textwidth]{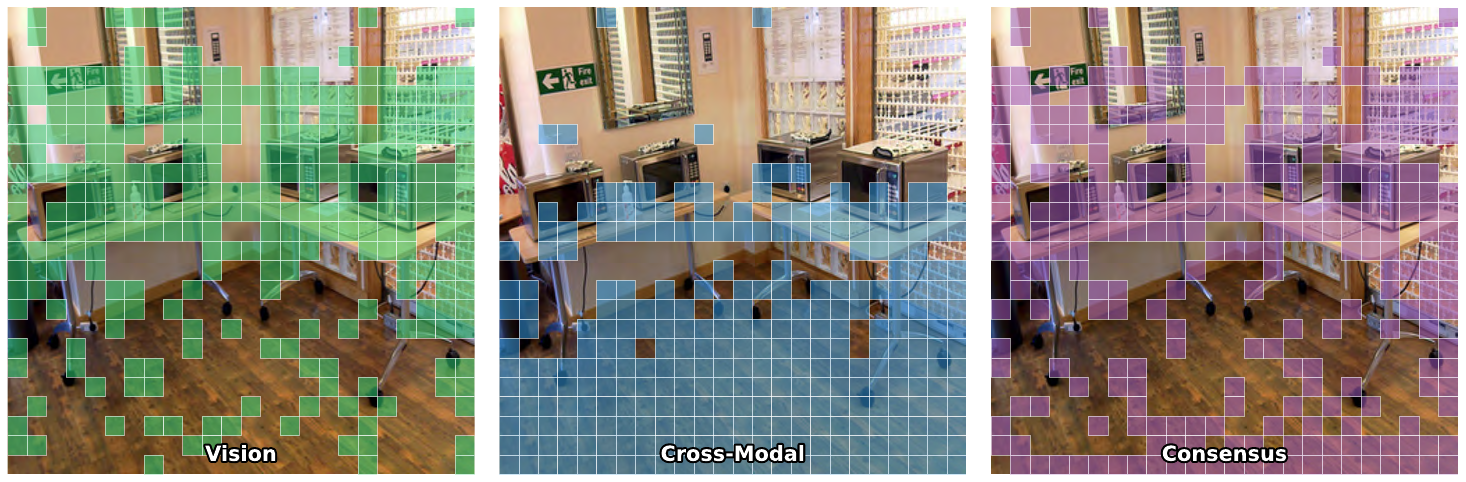}
\caption{Minimal comparison}
\end{subfigure}
\hfill
% Multi-K
\begin{subfigure}[b]{0.48\textwidth}
\includegraphics[width=\textwidth]{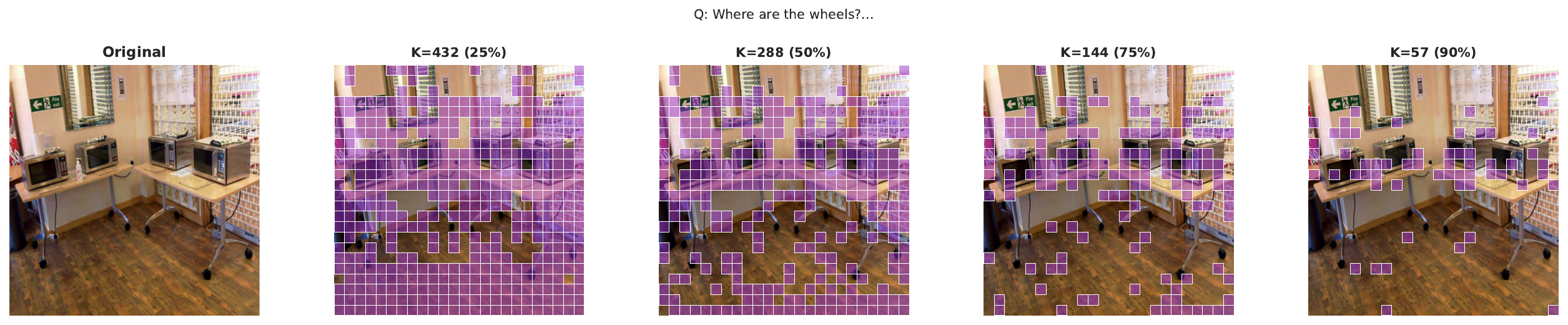}
\caption{Across retention ratios}
\end{subfigure}
\caption[Token selection]{\textbf{Q:} ``Where are the wheels?" \textbf{A:} ``The wheels are on the table...." (Disagreement: 54.2\%)}
\label{fig:k288_detailed_5}
\end{figure}

\begin{figure}
\centering
% Heatmap
\begin{subfigure}[b]{0.48\textwidth}
\includegraphics[width=\textwidth]{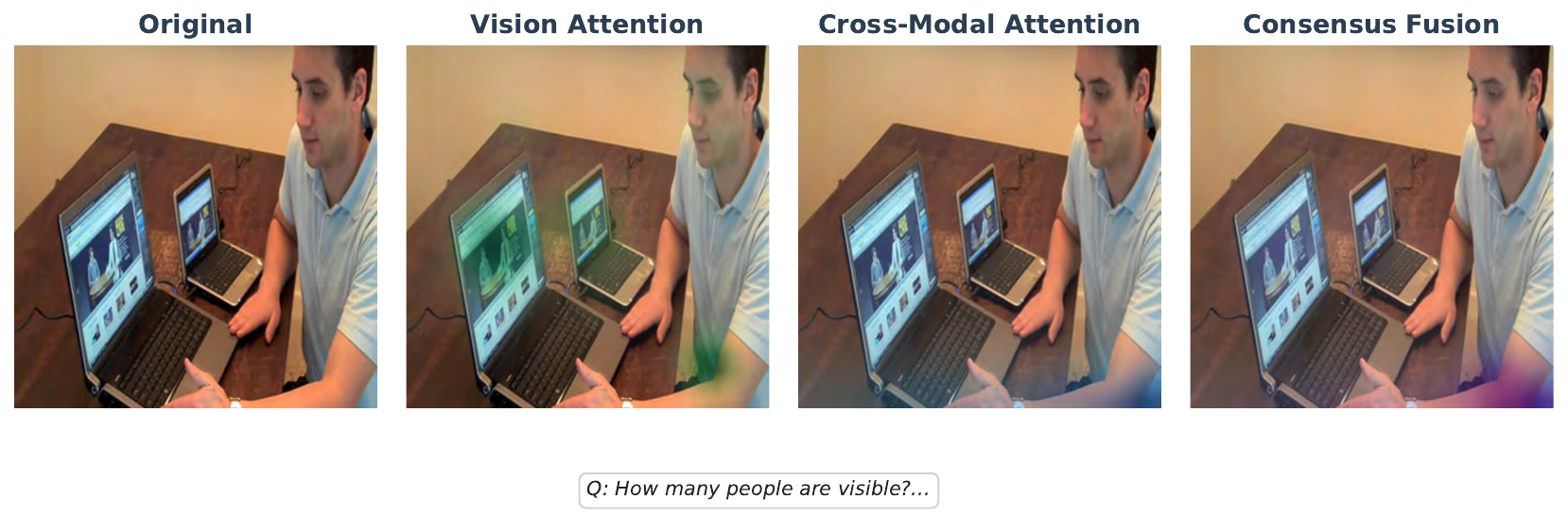}
\caption{Attention heatmaps}
\end{subfigure}
\hfill
% Venn
\begin{subfigure}[b]{0.48\textwidth}
\includegraphics[width=\textwidth]{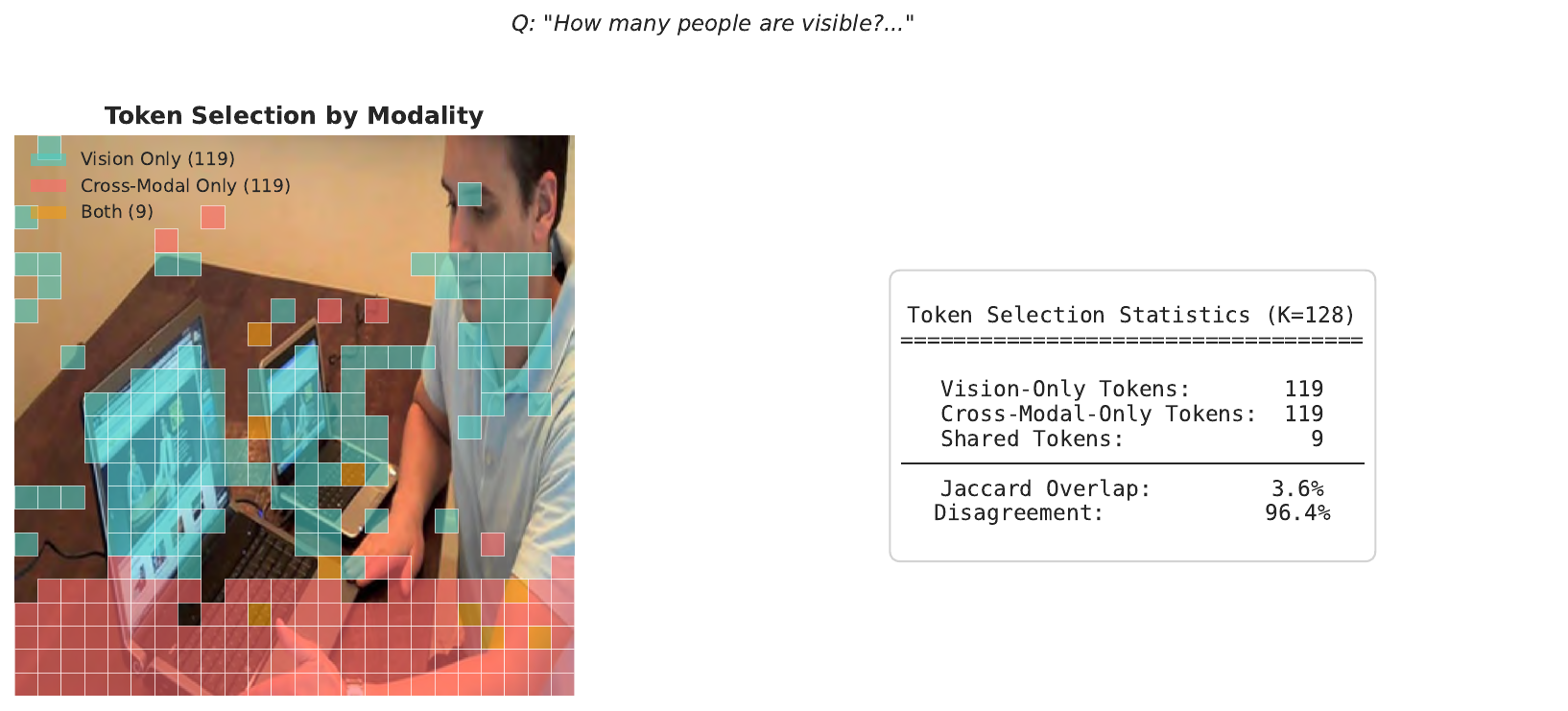}
\caption{Token set overlap}
\end{subfigure}
\\[2mm]
% Minimal
\begin{subfigure}[b]{0.48\textwidth}
\includegraphics[width=\textwidth]{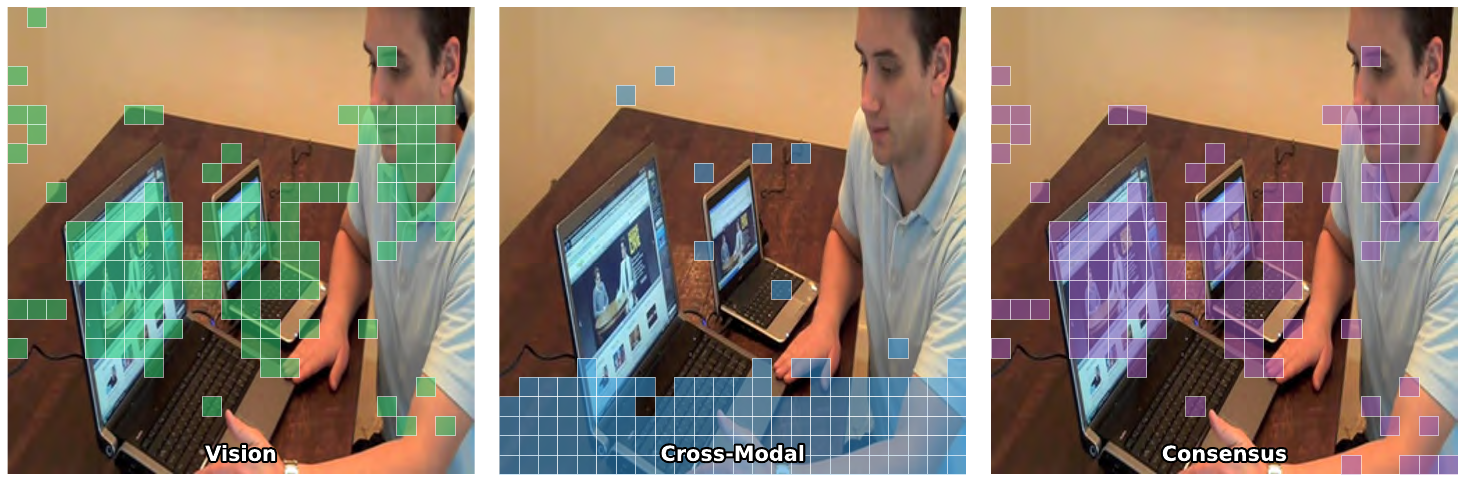}
\caption{Minimal comparison}
\end{subfigure}
\hfill
% Multi-K
\begin{subfigure}[b]{0.48\textwidth}
\includegraphics[width=\textwidth]{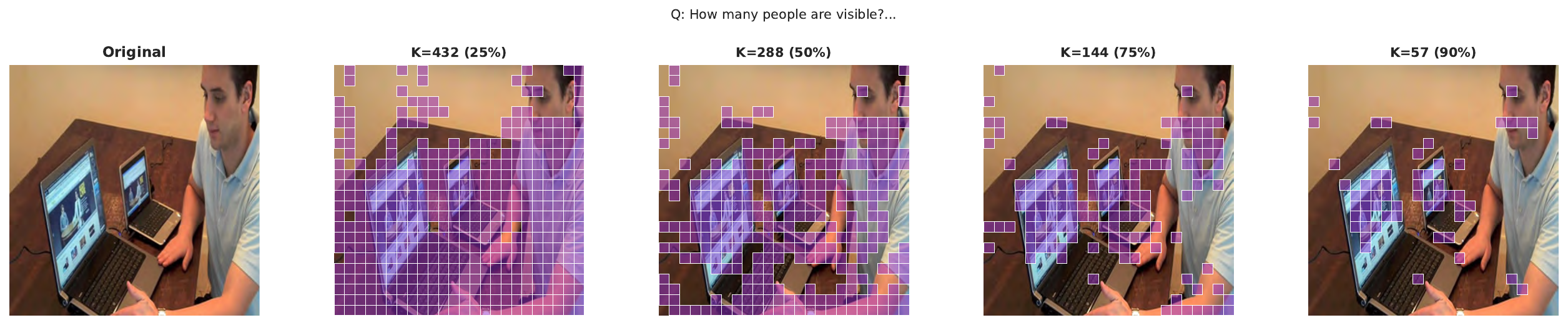}
\caption{Across retention ratios}
\end{subfigure}
\caption[Token selection]{\textbf{Q:} ``How many people are visible?" \textbf{A:} ``There is only one person visible in the image...." (Disagreement: 93.0\%)}
\label{fig:k128_detailed_1}
\end{figure}

\begin{figure}
\centering
% Heatmap
\begin{subfigure}[b]{0.48\textwidth}
\includegraphics[width=\textwidth]{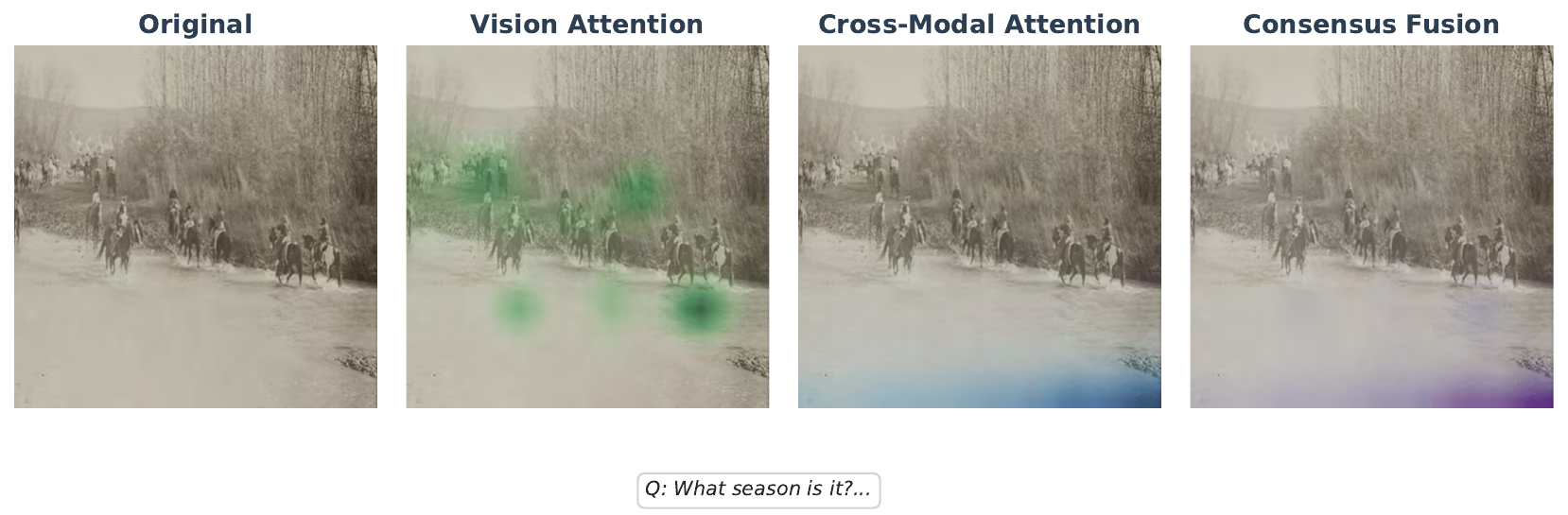}
\caption{Attention heatmaps}
\end{subfigure}
\hfill
% Venn
\begin{subfigure}[b]{0.48\textwidth}
\includegraphics[width=\textwidth]{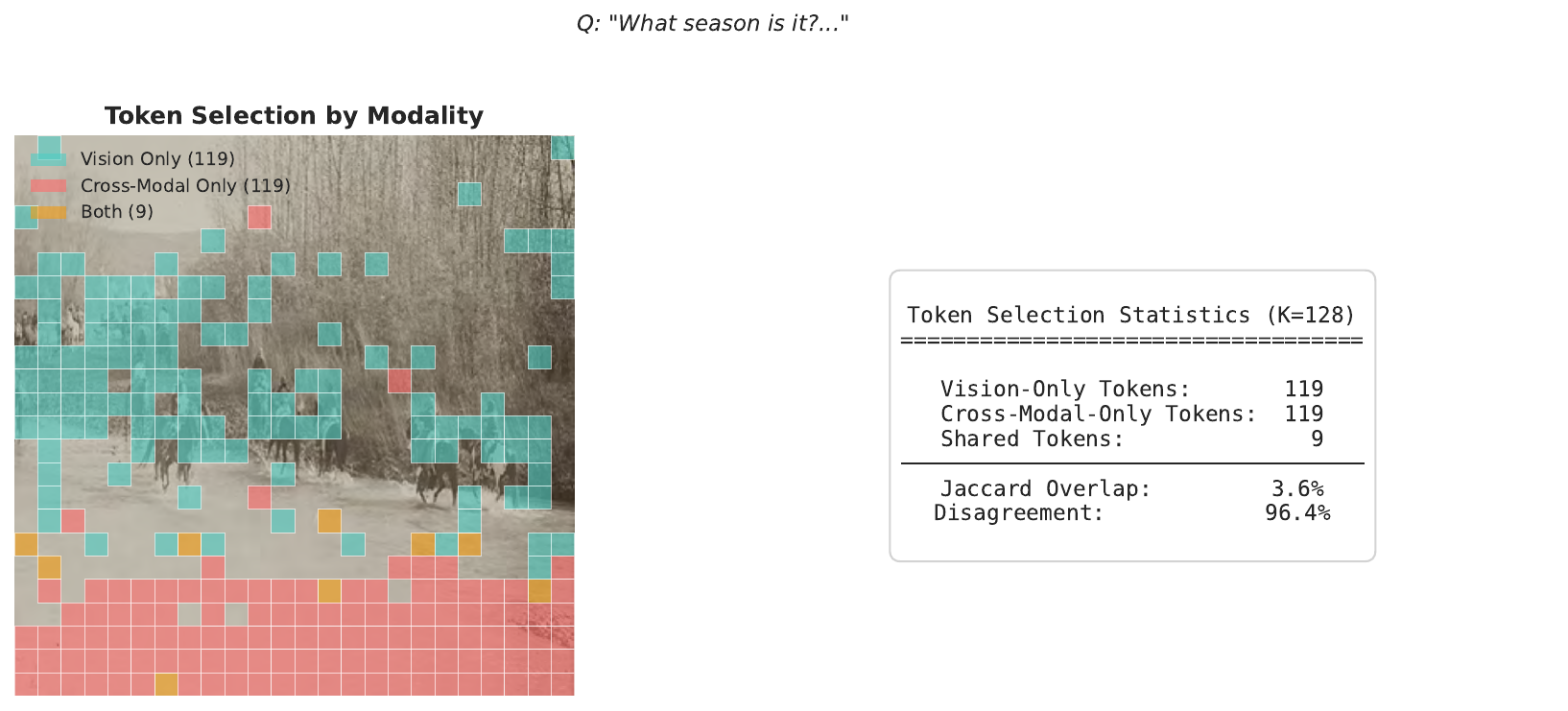}
\caption{Token set overlap}
\end{subfigure}
\\[2mm]
% Minimal
\begin{subfigure}[b]{0.48\textwidth}
\includegraphics[width=\textwidth]{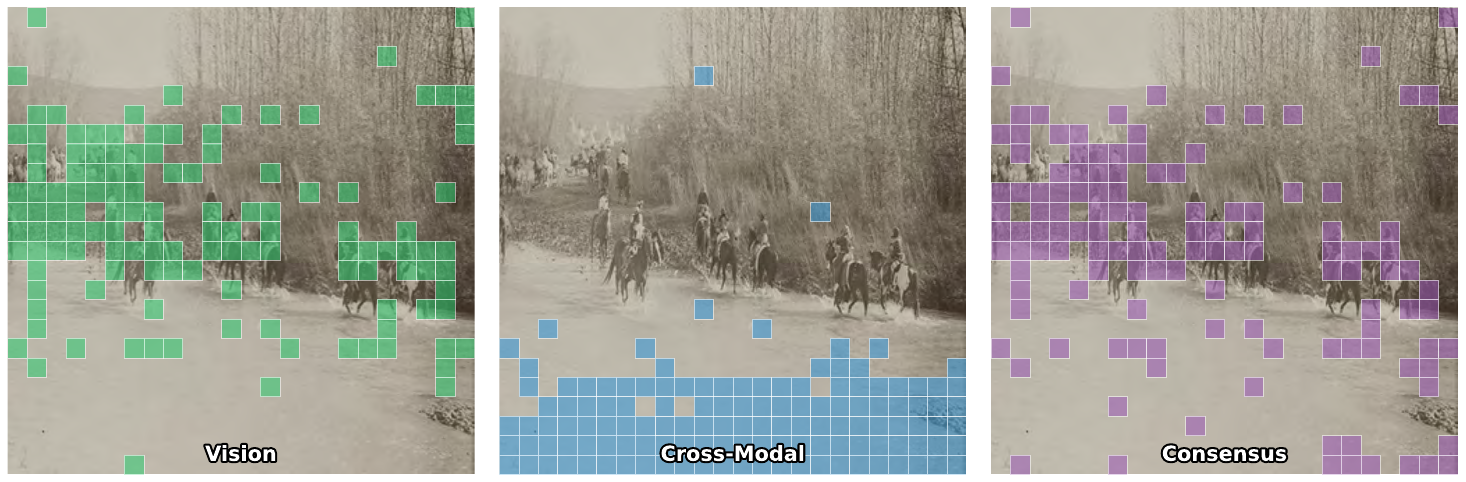}
\caption{Minimal comparison}
\end{subfigure}
\hfill
% Multi-K
\begin{subfigure}[b]{0.48\textwidth}
\includegraphics[width=\textwidth]{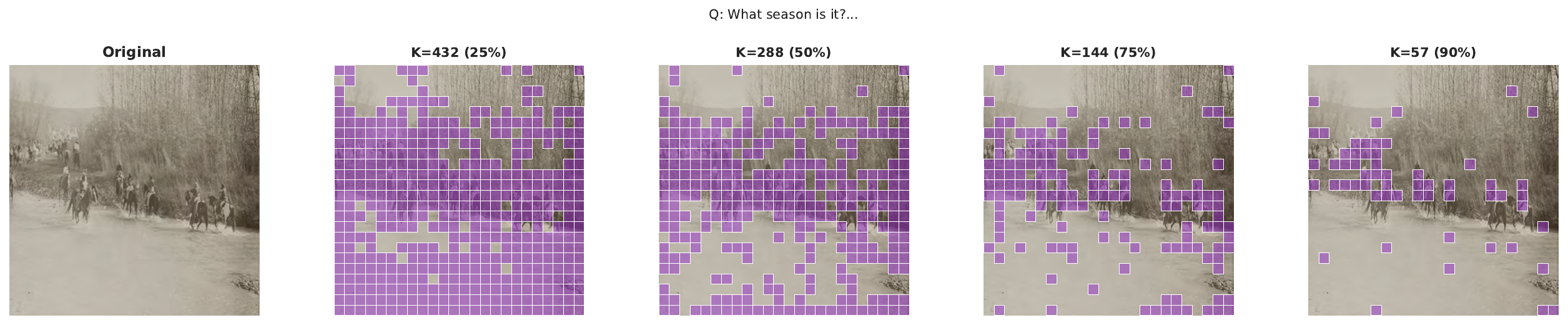}
\caption{Across retention ratios}
\end{subfigure}
\caption[Token selection]{\textbf{Q:} ``What season is it?" \textbf{A:} ``It is winter...." (Disagreement: 93.0\%)}
\label{fig:k128_detailed_2}
\end{figure}

\begin{figure}
\centering
% Heatmap
\begin{subfigure}[b]{0.48\textwidth}
\includegraphics[width=\textwidth]{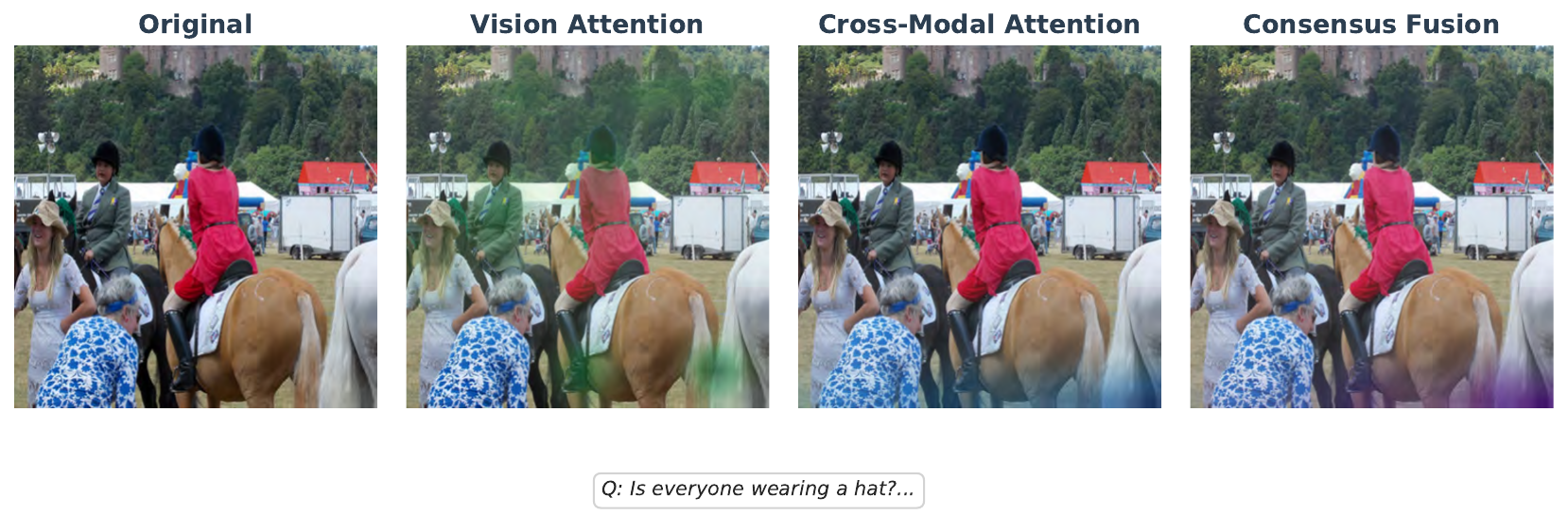}
\caption{Attention heatmaps}
\end{subfigure}
\hfill
% Venn
\begin{subfigure}[b]{0.48\textwidth}
\includegraphics[width=\textwidth]{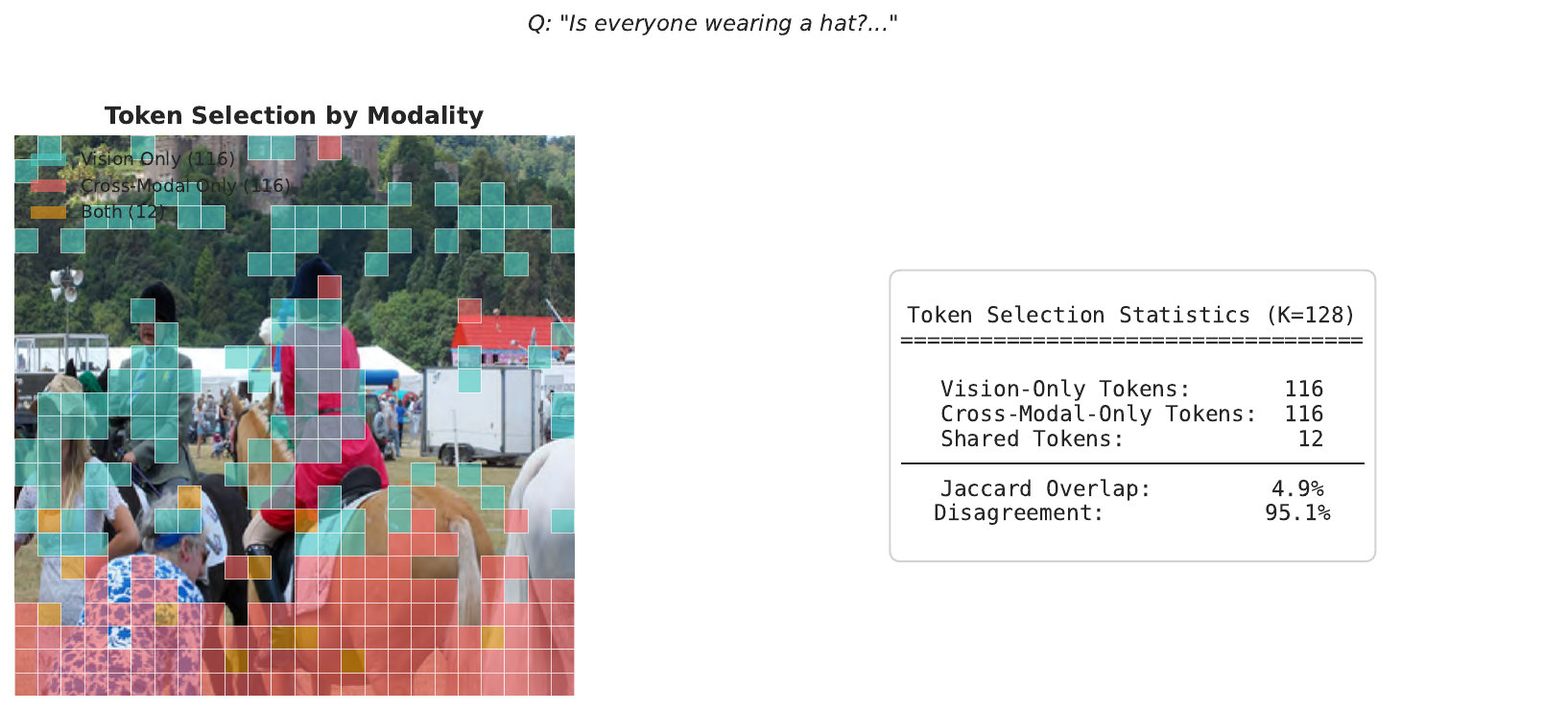}
\caption{Token set overlap}
\end{subfigure}
\\[2mm]
% Minimal
\begin{subfigure}[b]{0.48\textwidth}
\includegraphics[width=\textwidth]{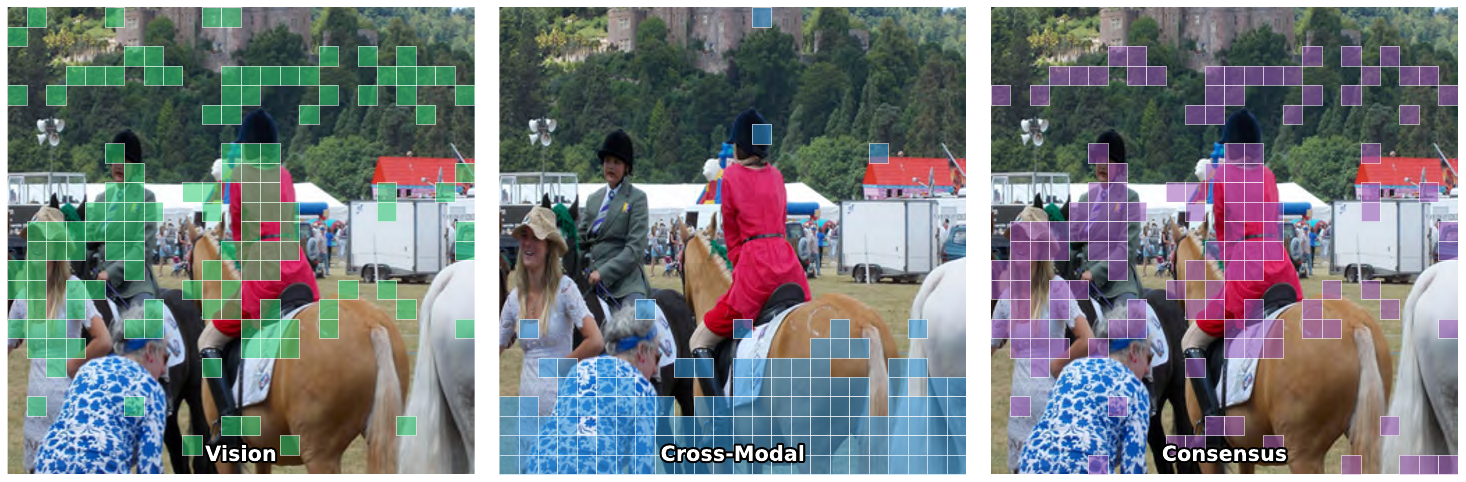}
\caption{Minimal comparison}
\end{subfigure}
\hfill
% Multi-K
\begin{subfigure}[b]{0.48\textwidth}
\includegraphics[width=\textwidth]{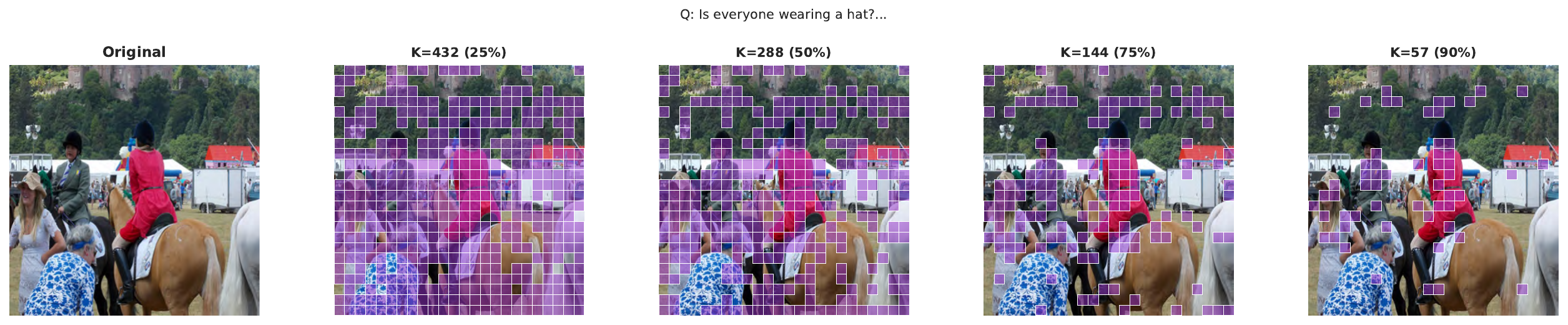}
\caption{Across retention ratios}
\end{subfigure}
\caption[Token selection]{\textbf{Q:} ``Is everyone wearing a hat?" \textbf{A:} ``Yes, everyone in the image is wearing a hat...." (Disagreement: 90.6\%)}
\label{fig:k128_detailed_3}
\end{figure}

\begin{figure}
\centering
% Heatmap
\begin{subfigure}[b]{0.48\textwidth}
\includegraphics[width=\textwidth]{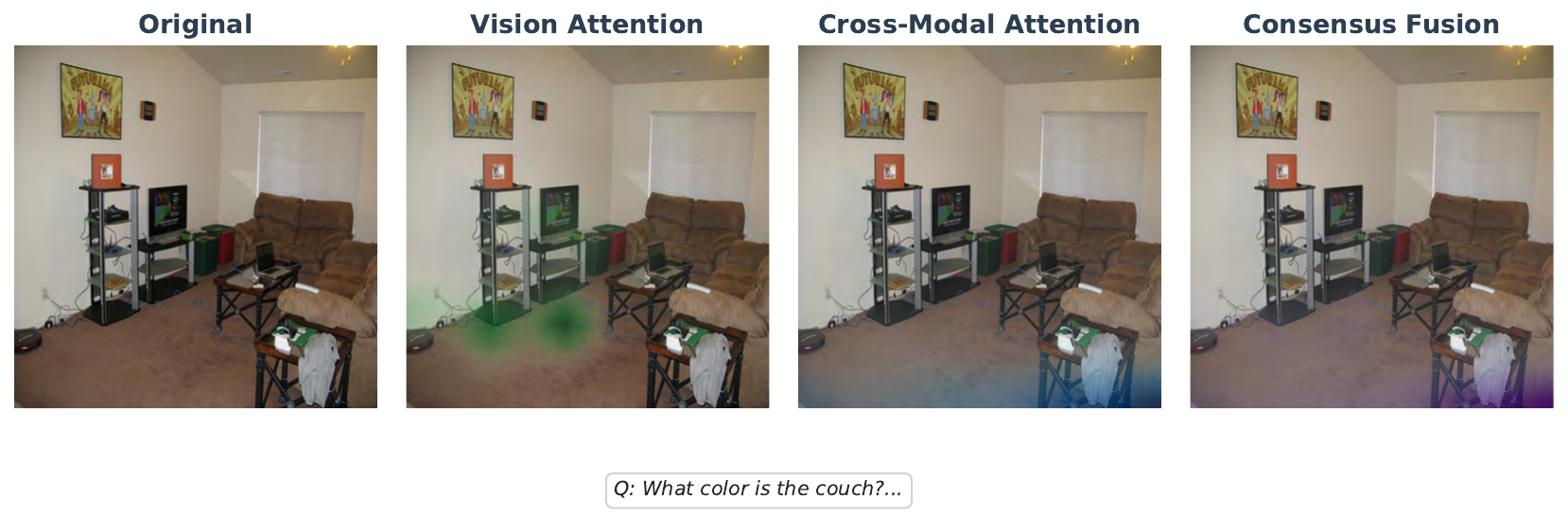}
\caption{Attention heatmaps}
\end{subfigure}
\hfill
% Venn
\begin{subfigure}[b]{0.48\textwidth}
\includegraphics[width=\textwidth]{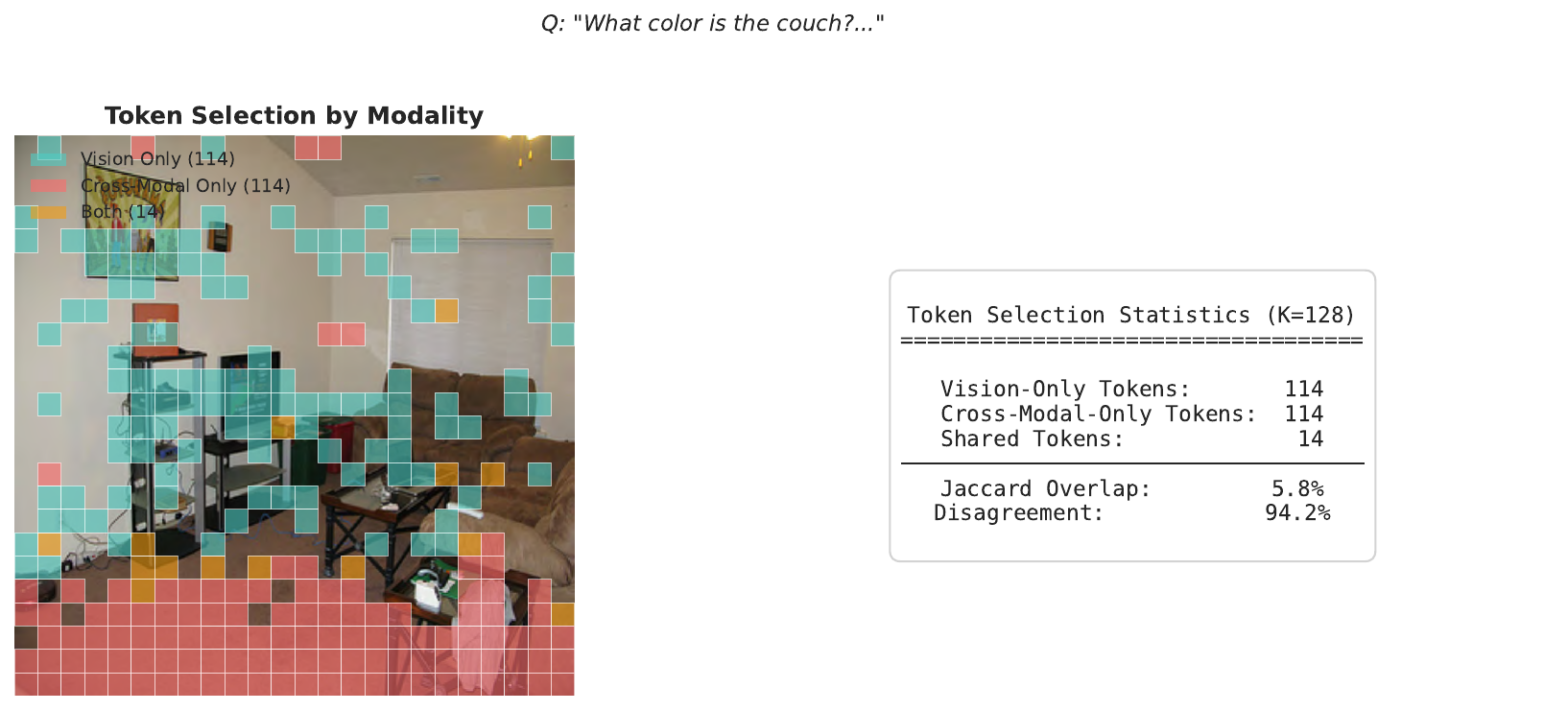}
\caption{Token set overlap}
\end{subfigure}
\\[2mm]
% Minimal
\begin{subfigure}[b]{0.48\textwidth}
\includegraphics[width=\textwidth]{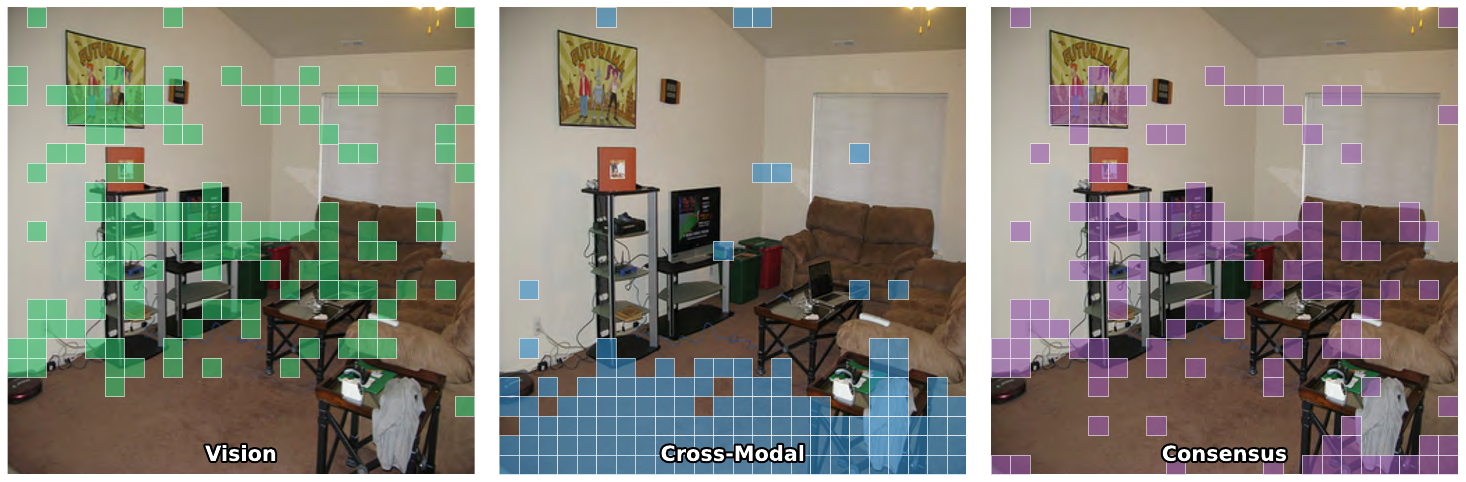}
\caption{Minimal comparison}
\end{subfigure}
\hfill
% Multi-K
\begin{subfigure}[b]{0.48\textwidth}
\includegraphics[width=\textwidth]{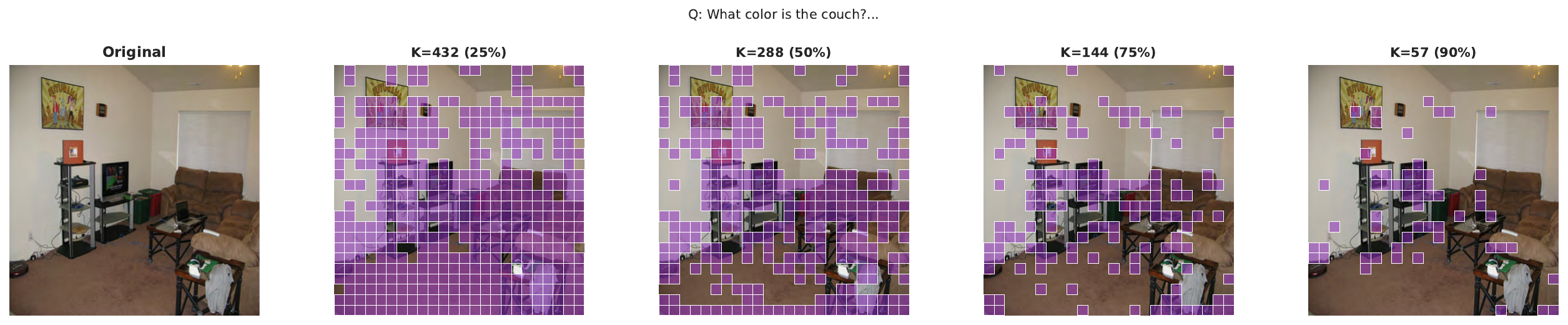}
\caption{Across retention ratios}
\end{subfigure}
\caption[Token selection]{\textbf{Q:} ``What color is the couch?" \textbf{A:} ``The couch is brown...." (Disagreement: 89.1\%)}
\label{fig:k128_detailed_4}
\end{figure}

\begin{figure}
\centering
% Heatmap
\begin{subfigure}[b]{0.48\textwidth}
\includegraphics[width=\textwidth]{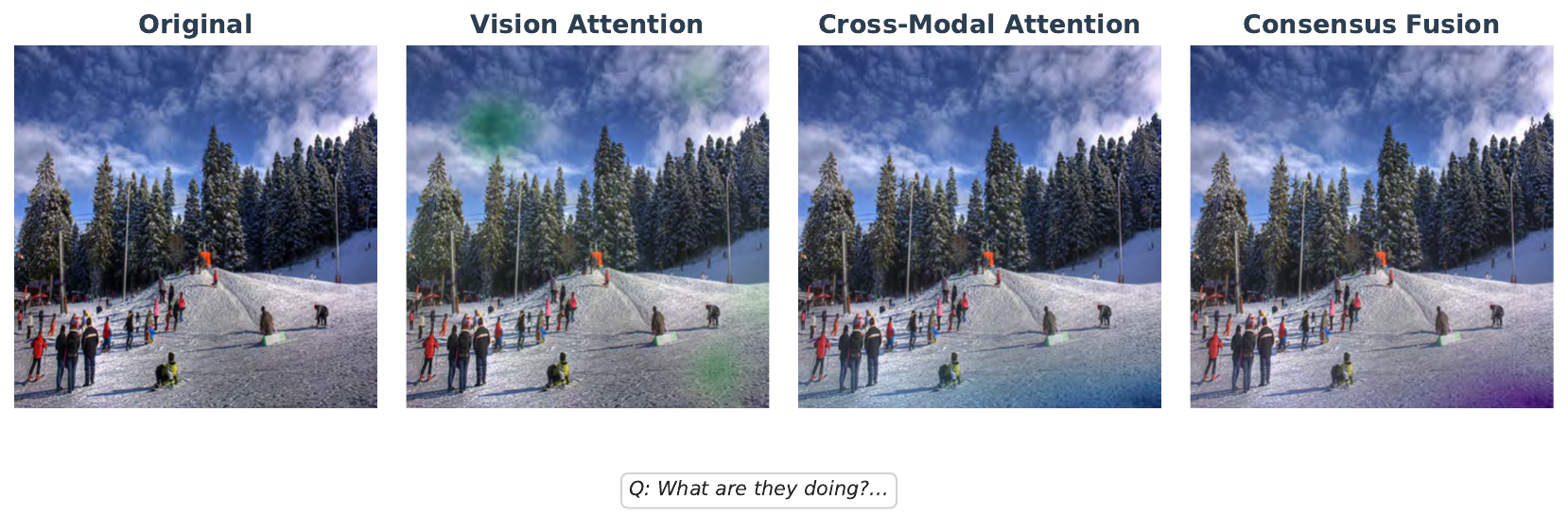}
\caption{Attention heatmaps}
\end{subfigure}
\hfill
% Venn
\begin{subfigure}[b]{0.48\textwidth}
\includegraphics[width=\textwidth]{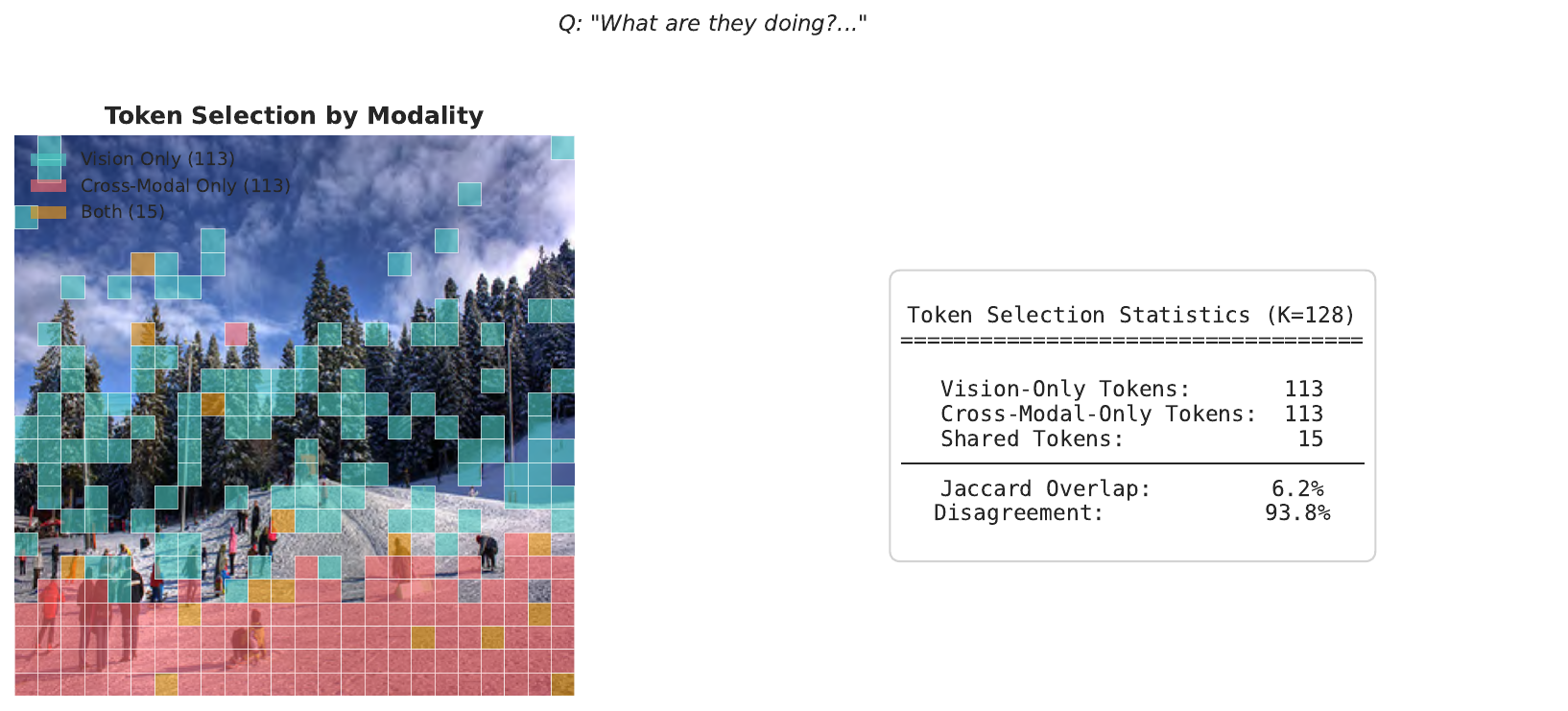}
\caption{Token set overlap}
\end{subfigure}
\\[2mm]
% Minimal
\begin{subfigure}[b]{0.48\textwidth}
\includegraphics[width=\textwidth]{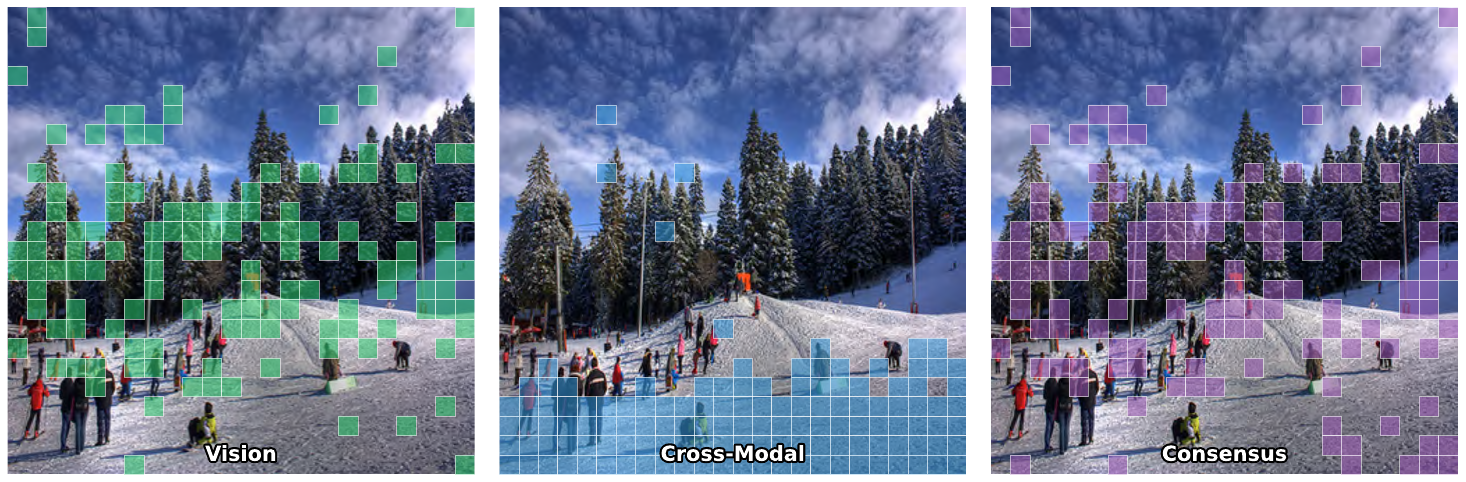}
\caption{Minimal comparison}
\end{subfigure}
\hfill
% Multi-K
\begin{subfigure}[b]{0.48\textwidth}
\includegraphics[width=\textwidth]{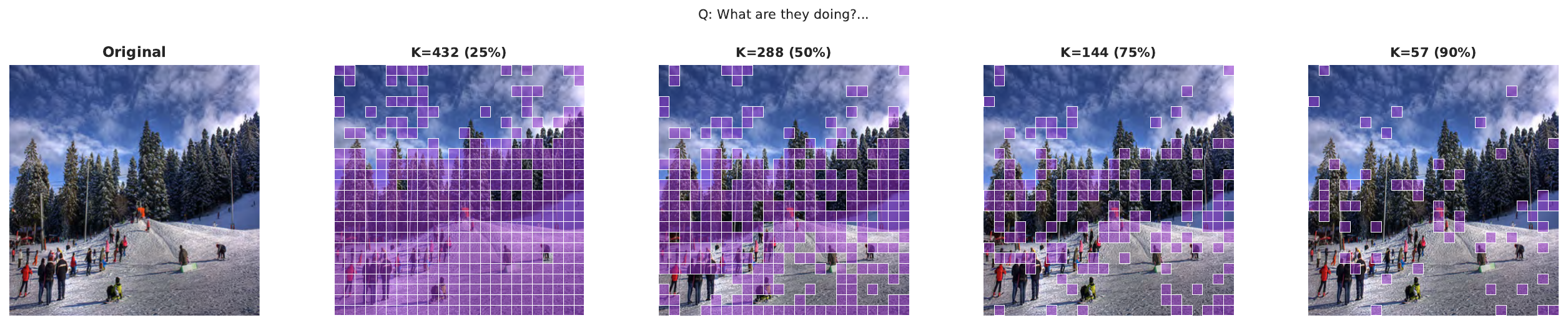}
\caption{Across retention ratios}
\end{subfigure}
\caption[Token selection]{\textbf{Q:} ``What are they doing?" \textbf{A:} ``The people in the image are skiing and snowboarding on a snow-covered slope...." (Disagreement: 88.3\%)}
\label{fig:k128_detailed_5}
\end{figure}

We analyze examples with the highest modality disagreement---cases where vision-only and cross-modal attention select substantially different token sets (Figure \ref{fig:k288_detailed_1} to \ref{fig:k128_detailed_5}). These examples best demonstrate the value of consensus fusion. At higher compression rates, careful token selection becomes even more critical. These examples demonstrate effectiveness of ConsensusDrop under aggressive pruning.

\subsection{Key Observations}
\begin{itemize}
    \item \textbf{Complementary signals:} Vision-based scores capture global saliency, while cross-modal attention identifies question-relevant regions. Neither alone is sufficient.
    \item \textbf{Positional bias in cross-modal:} Cross-modal attention exhibits positional biases due to RoPE, sometimes over-weighting tokens at sequence boundaries rather than semantically relevant ones, a known issue \cite{vispruner-prune-viz} in VLMs with a language model backbone using RoPE embeddings.
    \item \textbf{Consensus benefit:} Our convex fusion ($\alpha=0.7$) successfully combines both signals, retaining tokens that are \emph{both} visually salient \emph{and} semantically aligned.
    \item \textbf{Robustness across K:} The consensus mechanism remains effective across different retention ratios, from moderate (K=288) to aggressive (K=128) compression.
\end{itemize}

